\documentclass[twoside]{article}
\usepackage[accepted]{aistats2026}
\usepackage{hyperref}       % hyperlinks
\usepackage{url}            % simple URL typesetting
\usepackage{booktabs}       % professional-quality tables
\usepackage{amsfonts}       % blackboard math symbols
\usepackage{graphicx}       % for including graphics
\usepackage{amsmath}
\usepackage{hyperref}
\usepackage{cleveref}
\usepackage{amsthm}         % for theorem environments
\usepackage{nicefrac}       % compact symbols for 1/2, etc.
\usepackage{subcaption}
\usepackage{mathtools}
\usepackage{float}
\usepackage{textcase}

\usepackage{microtype}      % microtypography
\usepackage{xcolor}
\newtheorem{definition}{Definition}
\newtheorem{assumption}{Assumption}
\newtheorem{proposition}{Proposition}
\newtheorem{theorem}{Theorem}
\newtheorem{remark}{Remark}
\newtheorem{lemma}{Lemma}       
\Crefname{assumption}{Assumption}{Assumptions}
\usepackage[round]{natbib}

\begin{document}

\runningauthor{Roubing Tang, Sabina J. Sloman,  Samuel~Kaski}
\runningtitle{Requirements for Robust Bayesian Active Learning under Model Misspecification}
\twocolumn[
%\aistatstitle{Generalization Analysis for Bayesian Optimal Experimental Design under Model Misspecification}

\aistatstitle{Representative, Informative, and De-Amplifying: Requirements for Robust Bayesian Active Learning Under Model Misspecification}

\aistatsauthor{ Roubing Tang$^{1}$ \And Sabina J. Sloman$^{1}$ \And  Samuel~Kaski$^{1,2,3}$ }
\aistatsaddress{
$^{1}$ Department of Computer Science, University of Manchester, Manchester, UK \\
$^{2}$ ELLIS Institute Finland, Helsinki, Finland \\
$^{3}$ Department of Computer Science, Aalto University, Helsinki, Finland \\
\{roubing.tang, sabina.sloman, samuel.kaski\}@manchester.ac.uk 
}]

\newcommand{\trainx}{\mathbf{\xi}^{train}}
\newcommand{\testx}{\mathbf{\xi}^{test}}

\newcommand{\starf}{f^{\star}}
\newcommand{\barf}{\bar{f}}
\newcommand{\dtest}{d_{\text{test}}}
\newcommand{\dtrain}{d_{\text{train}}}
\newcommand{\hatfn}{\hat{f}^{(n)}}

\newcommand{\acqfri}{R-I}
\newcommand{\acqfria}{R-IDeA}

\begin{abstract}

    In many science and industry settings, a central challenge is designing experiments under time and budget constraints.
    \emph{Bayesian Optimal Experimental Design (BOED)} is a paradigm to pick maximally informative designs that has been widely applied to such problems. During training, BOED selects inputs according to a pre-determined acquisition criterion to target \textit{informativeness}. During testing, the model learned during training encounters a naturally occurring distribution of test samples. This leads to an instance of covariate shift, where the train and test samples are drawn from different distributions (the training samples are not \textit{representative} of the test distribution). 
    Prior work has shown that in the presence of model misspecification, covariate shift amplifies generalization error.
    Our first contribution is to provide a mathematical analysis of generalization error in the presence of model misspecification, revealing that, beyond covariate shift, generalization error is also driven by a previously unidentified phenomenon we term \emph{error (de-)amplification}.
    We then develop a new acquisition function that mitigates the effects of model misspecification by including terms for representativeness, informativeness, and de-amplification (R-IDeA).
    Our experimental results demonstrate that the proposed method performs better than methods that target only informativeness, only representativeness, or both.

\end{abstract}

\section{INTRODUCTION} \label{sec:introduction}
    Bayesian modeling is a principled approach to making inferences when data is scarce or costly.
     Most Bayesian machine learning methods are developed under the assumption that the true data-generating process (DGP) is included in the chosen model family \citep{bernardo2009bayesian}. However, in complex real-world environments, this assumption rarely holds: The true DGP often lies outside of the assumed model family \citep{uppal2003model}.
     The inevitability of the phenomenon of \textit{model misspecification} \citep{walker2013bayesian} is captured by the saying that ``all models are wrong'' \citep{George1976Science,box1980sampling}. 
     Common causes of model misspecification include omitted variables \citep{wooldridge2010econometric}, mistaken beliefs about the structure of the error term (e.g., a failure to account for heteroskedasticity or autocorrelation; \citealt{greene2003econometric,grunwald2017inconsistency}), or the choice of a misinformed or underexpressive model class \citep{wooldridge2010econometric,dubova2025ockhams}. 
     The consequences of model misspecification range from biased inferences \citep{greene2003econometric,muller2013risk,caprio2023credal,bonhomme2022minimizing}, unreliable approximations (e.g., in simulation-based inference methods; \citealt{frazier2020model,lintusaari2017fundamentals,huang2023learning}), to suboptimal decision-making \citep{sutton1998reinforcement,rainforth2024modern}. 

    There is a substantial literature on the effects of model misspecification on Bayesian inference when data is independently and identically distributed (i.i.d.), or ``passively'' collected from the distribution to which the learner wants their inferences to generalize
    \citep{kleijn2006misspecification,kleijn2012bernsteinvonmises,knoblauch2022optimizationcentric,walker2013bayesian,nott2023bayesian,kelly2025simulation}.
    However, in part because of the widespread availability of large datasets, \textit{active} learning methods have become much more prevalent \citep{settles2009active}.
    These methods select the training data to tailor it to a specified learning objective \citep{silberman1996active,farquhar2021statistical}.
    Active learning methods rely on the specified model twice: once to make inferences to fit training data, and again to select the data \citep{konyushkova2017learning}.
    Model misspecification thus has a double impact on these methods, introducing potential bias in both the acquisition function and the resulting inferences. In particular, in the context of active learning, model misspecification can produce poor quality datasets \citep{sugiyama2005active,bach2006active,ali2014active,vincent2017darc,farquhar2021statistical}. 
    Understanding the consequences of model misspecification is of paramount importance to developing robust active learning methods.

     In a Bayesian setting, Bayesian Optimal Experimental Design (BOED) is a natural and frequently used active learning method \citep{rainforth2024modern,huan2024optimal}.
     BOED selects the optimal design by maximizing an acquisition function known as \textit{the expected information gain} \citep{rainforth2024modern,chaloner1995bayesian}, enabling budget and time efficiency in many applications, such as drug discovery \citep{park2013bayesian}, clinical trial design \citep{chaloner1995bayesian}, chemistry \citep{walker2019bayesian,hickman2022bayesian}, biology \citep{kreutz2009systems,thompson2023integrating}, and psychology \citep{cavagnaro2010adaptive,myung2013tutorial}.
     While the limitations of BOED in the presence of model misspecification have been acknowledged in the literature, characterizing and proposing methods to overcome this limitation is an area of on-going research \citep{overstall2022bayesian,sloman2022characterizing,catanach2023metrics,schmitt2023detecting,ivanova2024step,barlas2025robust,forster2025improving,overstall2025gibbs}.

    We provide a novel theoretical analysis of generalization error in the presence of model misspecification.
    Our analysis reveals that training datasets that lead to robustness to model misspecification have two properties: They are \textit{representative} of the target DGP, and they are \textit{de-amplifying}.
    The expected information gain does not include a term for either of these, and standard BOED can lead to training datasets that have neither of these characteristics.
    In this sense, standard BOED is not robust to model misspecification.

    \textbf{Unrepresentative Training Data.}
        BOED selects samples to achieve a particular objective, and these samples likely do not reflect the distribution to which the learner would like to generalize.
        In other words, BOED induces a form of distribution shift, whereby the distribution used for (active) learning is different than the distribution used for evaluation.
        Recent work on the interaction between model misspecification and distribution shift has introduced the concept of \textit{misspecification amplification} \citep{amortila2024mitigating}, whereby the generalization error attributable to misspecification is ``amplified'' by the density ratio between the test and training input distributions. 
        A similar phenomenon has been observed in the context of BOED: In the presence of model misspecification, the generalization error in some settings has been shown to depend on the degree of model misspecification and the extent of distribution shift \citep{sloman2022characterizing}.

    \textbf{De-amplifying Training Data.}
        As our novel decomposition of generalization error shows, generalization performance depends on not only the representativeness of the training data, but also on the way it interacts with model (mis)specification (which will be defined in \Cref{sec:theory}): Generalization performance is enhanced when training data is in regions where the direction in which the model will tend to adjust on the basis of these data opposes the direction in which the model is misspecified.
        We refer to this property as error ``de-amplification'' to stress that the effect is to counteract, rather than amplify, the effect of the misspecification.

\textbf{Contributions.}
In this work, we explore the problem of BOED under model misspecification and make the following contributions:

\begin{itemize}
    \item \textbf{Theoretical Decomposition of Generalization Error.} 
    Prior work has primarily explored the effects of misspecification and distribution shift, overlooking the role of de-amplifying designs.
    We formally decompose generalization error into three components: (1) misspecification bias, (2) estimation bias, and (3) a novel term we introduce, \emph{error (de-)amplification}.
    We also derive an upper bound on generalization error that characterizes its dependence on the representativeness of the training data, the degree to which these data are de-amplifying, and model misspecification.

    \item \textbf{Novel Acquisition Function Incorporating Representativeness and De-amplification.}
    We propose a novel acquisition function designed to mitigate the effects of model misspecification by identifying designs that not only are informative, but are additionally representative and de-amplifying. Our empirical results show that the new acquisition outperforms traditional BOED in the presence of misspecification.

\end{itemize}

\section{PRELIMINARIES}
\label{sec:Preliminaries}
\subsection{Problem Setting}
    A modeler aims to predict an observed variable  $y \in \mathbb{R}  $ which depends on a fully observable input (design) $ \xi \in \Xi \subseteq \mathbb{R}^d $.
    The relationship between the observed variable $y$ and the input $\xi$ is governed by a conditional distribution $y \vert \xi \sim P^{\star} $, referred to as the \textit{true data-generating process (DGP)}, which depends on the output of a true (possibly unknown) regression function $\starf(\xi)$ and observation noise. 
    Let $\{(\xi_i, y_i)\}_{i=1}^n$ be a dataset of $n$ i.i.d.~samples drawn from the true DGP $P^{\star}$.
    To approximate the true DGP,
    the modeler proposes a hypothetical model  $f(\xi, \theta) ~ : ~ \Xi \mapsto \mathbb{R}$, where  $\theta \in \Theta $ represents the parameters within the fixed parameter space $\Theta $.
     The model class is denoted $\mathcal{F}(\xi, \Theta) = \{f(\xi, \theta) : \theta \in \Theta\} $. 
    \textit{Model misspecification} arises when the assumed model class $\mathcal{F}(\xi, \Theta)$ fails to capture the true DGP \citep{walker2013bayesian,kleijn2012bernsteinvonmises}.
    Let $\hatfn(\xi)$ be a learned predictor, depending on training designs $\{ \xi_1, \ldots{}, \xi_n \}$. Let $\barf \in \mathcal{F}$ be the predictor that best approximates the true data-generating function $\starf$, i.e., $\barf = \arg\min_{f \in \mathcal{F}} R_{\mathrm{test}}(f)$.
        
    \begin{definition}[Model misspecification]
    Model misspecification occurs when the assumed model class $ \mathcal{F}(\xi, \Theta) = \{f(\xi, \theta) : \theta \in \Theta\} $ cannot mimic the true output $ \starf(\xi) $ for any parameter $\theta \in \Theta $. That is, the model is misspecified if
    \begin{equation}
        \starf(\xi) \notin \mathcal{F}(\xi, \Theta).
    \end{equation}
    \end{definition}

    In Bayesian inference, the modeler additionally specifies a prior distribution over the model parameters.
    According to this prior, the probability that the data the learner will encounter is generated by a value $\theta$ is $p(\theta)$.
    Model fitting is carried out by updating the prior distribution using Bayes' rule.
    The result is a posterior distribution which assigns to a $\theta$ a probability $p(\theta \mid y, \xi) \propto p(\theta) p(y \mid \theta, \xi)$. This process depends on both the assumed prior and the specified likelihood model. When the model is misspecified, i.e., the likelihood does not reflect the true DGP, the updated posterior becomes unreliable or biased \citep{frazier2023reliable, oberauer2025variance}.

\subsection{Bayesian Optimal Experimental Design}
\label{subsec:design_selection}
    Bayesian Optimal Experimental Design (BOED) is a model-based framework to select the optimal design $\xi$ by maximizing the expected information gained about the parameter $\theta$, enabling budget and time efficiency \citep{rainforth2024modern, chaloner1995bayesian}. 
    The expected information gain (EIG) is \citep{dong2024variational,lindley1956measure}:
    \begin{equation}
    \begin{aligned}\label{eq:eig}
        \operatorname{EIG}(\xi) &= \mathbb{E}_{p(y \mid \xi)}[\operatorname{IG}_{\theta}(\xi, y))]\\
        &=\mathbb{E}_{p(\theta,y \mid \xi)} [\log p(y \mid\theta,\xi)-\log p(y \mid \xi)]
    \end{aligned}
    \end{equation}
    
    The optimal design $\xi^{\star}$ is the design in the set of candidate designs $\Xi$ that maximizes the EIG: 

    \begin{equation}\label{equ:select_design}
        \xi^{\star} = {\operatorname{argmax}_{\xi \in \Xi}} \operatorname{EIG}(\xi).
    \end{equation}

    \emph{Traditional BOED methods} \citep{foster2019variational,sebastiani1997bayesian}, also called Bayesian Adaptive Design (BAD), iterate between making design decisions by evaluating \Cref{equ:select_design}, and updating the underlying model through Bayesian inference to condition on data obtained so far. 
    Traditional BOED is computationally expensive, due to the substantial costs required to both estimate and optimize $\operatorname{EIG}(\xi)$ and update the model at each step.

\subsection{Distribution Shift}
    Distribution shift is a well-known challenge in machine learning.
    It refers to the setting where the data distribution differs between the training and test phases.
    In BOED, training designs are selected via an acquisition criterion, while the model's performance at test-time is evaluated on a given test distribution of interest. This mismatch induces a specific form of distribution shift known as \emph{covariate shift}, where the distribution of inputs shifts (i.e., $p_{\text{train}}(\xi) \neq p_{\text{test}}(\xi)$) while the conditional output distribution remains unchanged (i.e., $ p_{\text{train}}(y \mid \xi) = p_{\text{test}}(y \mid \xi)$).
    Prior work has studied the covariate shift induced by BOED \citep{sugiyama2005active,ali2014active,sloman2022characterizing}. 
    To address the potential resultant biases, density ratio estimation --- whereby the training data are reweighted according to the estimated ratio between test and training input distributions --- has proven effective \citep{ge2023maximum}.

\section{THEORETICAL RESULTS} \label{sec:theory}
    \newcommand{\misspecification}{\mathbb{B}}
    \newcommand{\covshift}{\mathbb{C}}
    \newcommand{\amplification}{\mathbb{A}}
    \newcommand{\outcome}{y}
    \newcommand{\maxval}[1]{#1_{\infty}}

    \subsection{Decomposition of Generalization Error}
    Recent work has demonstrated that generalization error depends on an interaction between the degree of covariate shift (the degree to which the training data are unrepresentative of the test distribution) and of model misspecification \citep{amortila2024mitigating,ge2023maximum,wen2014robust}. 
    In this section, we show that generalization error additionally depends on the degree of presence of a phenomenon we term \textit{error (de-)amplification}. We show that generalization error can be decomposed into three terms, reflecting separate contributions of the degree of misspecification bias, of estimation bias, and of error (de-)amplification. 

    Prior work \citep{hastie2009elements,sugiyama2005active} has provided decompositions of generalization error in the context of linear regression---where the error (de-)amplification term vanishes under the orthogonality assumption between model bias and estimation error.
    We extend such analyses to general (nonlinear) models under misspecification, where this orthogonality no longer holds.

    \begin{definition}[Generalization error ($R_{\mathrm{test}})$]\label{def:Gerror}
    Let $d_{\text{test}}$ be the test data distribution.
        The generalization error is defined as
        \begin{equation} \label{equ:Gerror}
            R_{\mathrm{test}}(\hatfn) := \mathbb{E}_{\xi \sim d_{\text{test}}}\left[(\hatfn(\xi) - \starf(\xi))^2\right].
        \end{equation}
    \end{definition}

        \begin{proposition}[Generalization Error Decomposition] \label{prop:gen_de}
            \Cref{equ:Gerror} can be decomposed into the following
            \begin{equation} \label{equ:Gerror_de}
                \begin{aligned}
                    R_{\mathrm{test}}(\hatfn)
                    &= \underbrace{\mathbb{E}_{\xi \sim \dtest}\left[(\barf(\xi) - \starf(\xi))^2\right]}_{\text{Misspecification Bias ($\misspecification $)}} \\
                    &+ \underbrace{\mathbb{E}_{\xi \sim \dtest}\left[(\hatfn(\xi) - \barf(\xi))^2\right]}_{\text{Estimation Bias ($\covshift$)}} \\
                    & + \underbrace{2 \, \mathbb{E}_{\xi \sim \dtest}\left[(\barf(\xi) - \starf(\xi))(\hatfn(\xi) - \barf(\xi))\right]}_{\text{Error (de-)amplification ($\amplification$)}}.
                \end{aligned}
            \end{equation}
        \end{proposition}

        In the \emph{well-specified} case where the true function lies within the model class, $\barf(\xi) = \starf(\xi)$. In this case, both the bias and interaction terms vanish, and the generalization error reduces to:
        \begin{equation}
            R_{\mathrm{test}}(\hatfn) = \mathbb{E}_{\xi \sim \dtest}\left[(\hatfn(\xi) - \barf(\xi))^2\right],
        \end{equation}
        where the only quantity that depends on the training sample is $\hatfn(\xi)$, since the test data distribution $\dtest$ and the best predictor $\barf$ are fixed.

        In the \textit{misspecified} case where the true function lies outside the model class, $\barf(\xi) \neq \starf(\xi)$. In this case, all three terms in \Cref{equ:Gerror_de} contribute to generalization error.
        The terms have the following interpretations:
        \begin{itemize}
            \item \textbf{Misspecification Bias ($\misspecification$)} captures the discrepancy between the best predictor and the true data-generating function, and reflects the degree of model misspecification. This term is fixed and unaffected by the training data.
            \item \textbf{Estimation Bias ($\covshift$)} captures the discrepancy between the best predictor $\barf \in \mathcal{F}$ and the predictor arrived at on the basis of finite training data. In the BOED setting, this training data depends on the modeler's sequential evaluations of the EIG.
            \item \textbf{Error (De-)amplification ($\amplification$)}
             measures the correlation between the extent and direction of the model's bias and the extent and direction of the estimation error over the \textit{test distribution}. This term can either amplify or mitigate the overall generalization error:
            \begin{itemize}
                \item A positive correlation indicates that the directions of misspecification and estimation bias tend to coincide, which \emph{amplifies} (increases) the generalization error.
                \item A negative correlation indicates that the directions of misspecification and estimation bias tend to counteract each other, which \emph{de-amplifies} (decreases) the generalization error. 
            \end{itemize}
        \end{itemize}

        \noindent

    \subsection{An Upper Bound on Generalization Error with Error (De-)amplification}\label{sec:ub}
    While \Cref{prop:gen_de} provides valuable insights into the various contributors to generalization error,  computing these terms requires evaluating the outputs of $\starf$ and $\barf$ in expectation over the test samples.    
    Of course, this is infeasible in practice.
    
    To understand and control generalization error during training, the learner requires a formulation that explicitly relates it to quantities available during training.
    \Cref{prop:gen_error_covshift} provides such a formulation.
    \Cref{prop:gen_error_covshift} shows that the learner can control generalization error by selecting designs that (i) are \textit{representative} of the test distribution (reduce the degree of covariate shift), and (ii) are \textit{de-amplifying} (have the potential to counteract the misspecification bias).
    We apply these insights in our development of a novel acquisition function in \Cref{sec:new-acqf}.
    
    \Cref{prop:gen_error_covshift} builds on a result from  \cite{amortila2024mitigating}.
    In particular, we tighten the upper bound introduced by \cite{amortila2024mitigating} to depend on the degree of (de-)amplification of the training samples.

    We use $\dtrain$ (resp., $\dtest$) to refer to the training (resp., test) data distribution.
    Throughout, we assume that $\dtrain(\xi) >0$ for all $\xi \in \Xi$, i.e., that each candidate design has some positive probability of being encountered during training.
    
    We make use of the following definitions introduced by \cite{amortila2024mitigating}:
    \begin{definition}
        [The degree of covariate shift] 
        \label{assump:density_ratio}
            The density ratio between the test and training input distributions is \citep{amortila2024mitigating, sugiyama2007direct}:
            \begin{equation}
                \maxval{\covshift} := \sup_{\xi \in { \Xi}} \left| \frac{\dtest (\xi)}{\dtrain(\xi)} \right|.
            \end{equation}
    \end{definition}
    The degree of covariate shift $\maxval{\covshift}$ measures the worst-case distance between the selected training samples and the test samples. A more representative design (i.e., one that reduces covariate shift) helps control estimation bias.
   
        \begin{definition}[The degree of misspecification] \label{assump:miss_degree}
            The discrepancy between the predictive distribution induced by $\barf$ and that of the true data-generating function $\starf$ is \citep{amortila2024mitigating}:
            \begin{equation}
                \maxval{\misspecification} :=\|\barf - \starf\|_{\infty}=  \sup_{\xi \in \Xi} \mid \barf(\xi)- \starf(\xi)  \mid 
            \end{equation}
            where $\maxval{\misspecification} \geq 0$.
        \end{definition}
        The degree of misspecification $\maxval{\misspecification}$ measures the worst-case discrepancy between the data-generating function and the best predictor in the model class. 
        In other words, it is an upper bound on the misspecification bias $\misspecification$.
        Notice that if $\maxval{\misspecification} = 0$, the model is well-specified (i.e., $\starf \in \mathcal{F}$).
        On the other hand, if $\maxval{\misspecification} > 0$, the model is misspecified (i.e., $\starf \notin \mathcal{F}$) and $\maxval{\misspecification}$ quantifies the degree of misspecification.

        We also require the following assumption:
        \begin{assumption}[Boundedness of model class and outcomes \citep{amortila2024mitigating}] \label{assump:boundedness}
            For all $\xi \in \Xi$, $$\sup _{f \in \mathcal{F}} \|f\|_{\infty} \leq \maxval{\outcome} \text{, } \|\starf \|_{\infty} \leq \maxval{\outcome} \text{,  and } |y| \leq \maxval{\outcome} $$ for some $0<\maxval{\outcome} < \infty $
            and where $\|f\|_{\infty} = \sup_{\xi \in \Xi} \mid f(\xi)\mid $.
        \end{assumption}
        The finite setting ensures that the bound in Theorem 1 is non-vacuous.
        
    \paragraph{Our Result.}
    \Cref{prop:gen_error_covshift} extends the result of \cite{amortila2024mitigating} by explicitly characterizing the behavior of generalization error in a way that accounts for error (de-)amplification.
        While \Cref{prop:gen_error_covshift} characterizes generalization performance given a data set (i.e., in the data-fitting phase), it can also inform data selection by revealing properties of those data that facilitate generalization.
        The insights from \Cref{prop:gen_error_covshift} motivate us to incorporate error (de-)amplification and representativeness into the decision-making (design selection) phase, thereby reducing the generalization error in the data-fitting phase, particularly in settings with a limited number of training samples.
         
        \begin{theorem}[Generalization Error Bound under Covariate Shift with Amplification]
        \label{prop:gen_error_covshift}
            Let $\mathcal{F}$ be a finite model class, and let $\starf$ denote the true regression function. Let $\hatfn$ be the empirical risk minimizer over training data drawn from the distribution $d_{\mathrm{train}}$. 
            Then, with probability at least $1 - \delta$, the generalization error under covariate shift satisfies:
            \begin{equation}\label{equ:bound}
                R_{\text{test}}(\hatfn) \leq \maxval{\covshift} \cdot \left\{
                    \begin{aligned}
                    & \maxval{\misspecification}^2 
                         + \frac{224 \maxval{\outcome}^2 \log (|\mathcal{F}| / \delta)}{3n}
                         - 2\widehat{\amplification}(\hatfn) , \\
                         &\quad\quad\quad\quad\quad\quad\quad\quad \text{if } \widehat{\amplification}(\hatfn) < 0 \\
                    & 
                         \maxval{\misspecification}^2 + \frac{128\maxval{\outcome}^2\log (|\mathcal{F}| / \delta)}{3n} - \sqrt{3}\widehat{\amplification}(\hatfn),\\
                         &\quad\quad\quad\quad\quad\quad\quad\quad \text{if } 0\leq\widehat{\amplification}(\hatfn)
                    \end{aligned}
                    \right.
            \end{equation}
            where $\widehat{\amplification}(\hatfn) \coloneqq \mathbb{E}_{\xi \sim d_\text{train}}\left[ (\hatfn(\xi)- \barf(\xi)) ( \barf(\xi) -\starf(\xi)) \right]$. The proof can be found in  \Cref{appendix:proof_thm2}.

        \end{theorem}
        \begin{remark}[Connection to \Cref{prop:gen_de}]
            This bound is consistent with the full generalization error decomposition structure, including the cross-term $\widehat{\amplification}(f)$, which captures the interaction between properties of the model (misspecification bias) and of the sampling strategy (estimation bias). 
            \Cref{prop:gen_de} does not explicitly reveal how the representativeness and (de-)amplifying properties of the training data interact with model bias.
            In contrast, \Cref{prop:gen_error_covshift} makes this interaction explicit, providing a more interpretable perspective for understanding and controlling the generalization error during the training phase.
        \end{remark}
        
        \begin{remark}
        \label{remark:uncomputable_A}
            \Cref{prop:gen_de} defines the generalization error as a function of the learned function $\hatfn$ and the true data-generating function $\starf$, in expectation over the test distribution. However, since the true data-generating function $\starf$ is unknown and outside the model class, it cannot easily offer guidance for decision-making. In contrast, although the $\widehat{\amplification}$ term that appears in \Cref{prop:gen_error_covshift} depends on the unknown $\barf$, $\barf$ is within the model class $\mathcal{F}$, which suggests that it can be better approximated in practice.
            We leverage this in our construction of a novel acquisition function in \Cref{sec:r-idea}.
        \end{remark}
        \begin{remark}
            \Cref{prop:gen_error_covshift} requires the assumption that $\hatfn$ is the member of $\mathcal{F}$ that minimizes risk in the training data.
            Although our experiments adopt the Bayesian learning framework, in which the learner predicts on the basis of a distribution over members of $\mathcal{F}$, \Cref{prop:gen_error_covshift} still provides valuable insights about the role of representative and (de-)amplifying training samples.
        \end{remark}

        \Cref{prop:gen_error_covshift} reveals that the following factors contribute to generalization error:
        \begin{itemize}
            \item \textbf{Representativeness of the Training Data ($\maxval{\covshift}$).}
                The multiplicative factor of $\maxval{\covshift}$ implies that, by reducing $\maxval{\covshift}$, choosing more representative training data can reduce generalization error.
                
            \item \textbf{Misspecification Bias ($\maxval{\misspecification}$).}
                Misspecification bias cannot be reduced by the training data. However, because of the multiplicative effect of $\maxval{\covshift}$, the effect of model misspecification on generalization error can be amplified by unrepresentative training samples \citep{amortila2024mitigating}.

            \item \textbf{Error (De-)amplification ($\widehat{\amplification}$).}
                This term captures a key component of generalization error: the interaction between the learner's misspecification and estimation errors.
                The term $\widehat{\amplification}(\hatfn)$ can be interpreted as an average, across the training samples, of the (signed) estimation errors weighted by the (signed) misspecification errors.
                Where $\widehat{\amplification}(\hatfn)$ is large (error de-amplification),
                the learner's misspecification and estimation biases tend to agree, and so sampling at the given design (reducing the estimation error) has a \textit{de-amplifying} effect.

                For any $\xi \in \Xi$, the misspecification error
                $\barf(\xi) - \starf(\xi)$ is fixed and cannot be removed by additional data. In selecting the de-amplifying design, the objective is to induce a negative correlation between the estimation error $\hat f(\xi) - \bar f(\xi)$ and the misspecification error in regions where the misspecification error is large.
                In other words, when the misspecification error is large, the estimation error should not exacerbate it; ideally, the estimation error offsets (de-amplifies) the misspecification.
                \Cref{fig:deamplifying_region} illustrates the (de-)amplifying regions in a simple example.
                \begin{figure}
                    \centering
                    \includegraphics[width=1.0\linewidth]{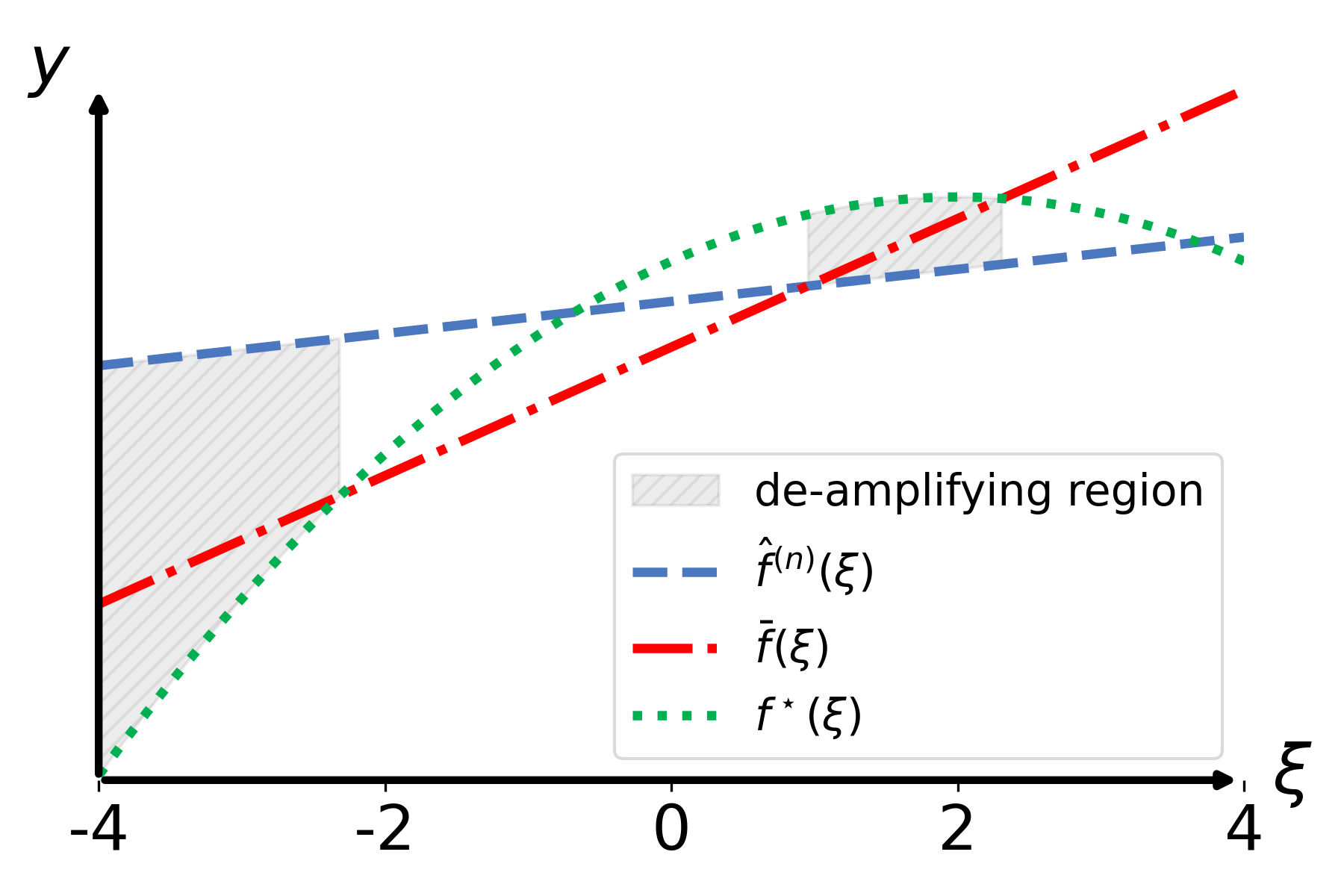}
                    \caption{
                    Illustration of error amplification and de-amplification.
                    The green curve denotes the true data-generating function $\starf(\xi)$,
                    the red line represents the best-in-class approximation $\barf(\xi)$,
                    and the blue line shows the learned predictor $\hatfn(\xi)$.
                    The grey dashed shading highlights  \emph{de-amplifying} regions, where
                    $\barf(\xi)$ lies between $\starf(\xi)$ and $\hatfn(\xi)$,
                    so that estimation error and misspecification error partially offset each other.
                    In contrast, the remaining regions indicate \emph{amplifying} regions,
                    where estimation error reinforces misspecification, leading to larger overall prediction error.
                    }
                    \label{fig:deamplifying_region}
                \end{figure}

        \end{itemize}

\section{TWO NOVEL ACQUISITION FUNCTIONS}\label{sec:new-acqf}
    Leveraging the insights from \Cref{prop:gen_error_covshift}, we design two novel acquisition functions.
    \acqfri{} identifies designs that are both \textit{representative} and \textit{informative} about the parameter of interest.
    \acqfria{} identifies designs that are \textit{representative}, \textit{informative},  and \textit{de-amplifying}.

    \subsection{\acqfri{}: A Representative and Informative Acquisition Function} \label{sec:1-idea}      
        To account for covariate shift, we modify the standard EIG acquisition function by introducing a maximum mean discrepancy (MMD)-based correction term. The idea is to encourage the selection of design points that not only have high information gain but also help reduce the difference between the distributions of training and test points. Specifically, we use the following form:
            \begin{equation}
                \label{equ:acqfri}
                \text{\acqfri}(\xi_t) 
                = \text{EIG}(\xi_t) \cdot 
                \underbrace{\left(1 - \lambda \frac{\text{MMD}(h_{t-1} \cup \{\xi_t\}, d_\text{test})}{\text{MMD}(h_{t-1}, d_\text{test})} \right))}_{\text{Robust Ratio}}
            \end{equation}
             where $h_{t-1}$ is the history of selected designs before time step $t$.
             The motivation and expression for MMD can be found in \Cref{appendix:appendix_ex_detail}.
             
             This robust acquisition function penalizes designs that are only representative or only informative; in other words, the designs selected by \acqfri{} are both representative and informative. 
             The hyperparameter $\lambda$ controls the tradeoff between informativeness and representativeness. When $\lambda$ tends to zero, the selected designs are informative, and \acqfri{} selects similar designs to traditional BOED.

    \subsection{\acqfria{}: A Representative, Informative, and De-amplifying Acquisition Function}\label{sec:r-idea}
        \newcommand{\DeAregion}{\Xi_{\mathbb{A}+}}
        \newcommand{\approxDeAregion}{\widehat{\Xi}_{\mathbb{A}+}}
        \newcommand{\proxyDeAregion}{\widehat{\Xi}_{\mathbb{A}+}^g}
        \Cref{prop:gen_error_covshift} shows that larger $\widehat{\amplification}(\hat{f}^{(n)})$ implies that designs on which $\hat{f}^{(n)}$ was trained
        \emph{de-amplify} generalization error.  
        We refer to a design $\xi$ as de-amplifying whenever $(\hatfn(\xi)- \barf(\xi)) ( \barf(\xi) -\starf(\xi)) \geq \tau_0$ for a given threshold $\tau_0$. Ideally, we would design an acquisition function that selects only designs in the de-amplifying region $\DeAregion(\tau_0) \coloneqq \Big\{\xi \in \Xi:(\hatfn(\xi)- \barf(\xi)) ( \barf(\xi) -\starf(\xi)) \geq \tau_0\Big\}$.
        However, as discussed in \Cref{remark:uncomputable_A}, this cannot be determined exactly since $\starf$ and $\barf$ are unknown.
        We instead construct a subset of the de-amplifying region, the \textit{approximate de-amplifying region}:
        \begin{theorem}[Approximate de-amplifying region]
        \label{theor:worstcase}
            Let $\barf$ be the predictor that best approximates the true data-generating function $\starf$. Then,
            \begin{equation}\label{equ:worstcase}
                \approxDeAregion(\tau_1) \subseteq \DeAregion(\tau_0) \nonumber
            \end{equation}
            where the \textnormal{approximate de-amplifying region}
            $\approxDeAregion(\tau_1) \coloneqq \Big\{\xi \in \Xi:|\hatfn(\xi)-\barf(\xi)|  
            \geq \tau_1 \Big\}$, $\tau_1 = \tau_0/\maxval{\misspecification} + c\maxval{\misspecification}$, $c \geq 2$, and $\tau_0 \geq 0$.
            A detailed derivation can be found in Appendix \ref{appendix:proof_ridea_all}.
        \end{theorem}
        $\tau_0$ and $c$ are constants that determine the minimum threshold required for a design to be considered part of the approximate de-amplifying region.

        Unlike $\DeAregion$, $\approxDeAregion$ does not depend on $f^{\star}$.
        However, it does depend on $\barf$.
        As discussed in \Cref{remark:uncomputable_A}, $\barf$ is within the model class $\mathcal{F}$.
        Below, we leverage this in construction of the \textit{proxy approximate de-amplification region}, which depends on a trainable proxy $g$:  
      
        \begin{lemma}[Proxy approximate de-amplification region $\approxDeAregion(\tau)$]\label{lem:proxy_fidelity}

        Given a \textnormal{proxy function} $g ~ : ~ \Xi \mapsto \mathbb{R}$ such that $\sup_{\xi\in\Xi}|g(\xi)-\bar f(\xi)| \le \tau_2$ for some $\tau_2 \geq 0$,
        \[
          \proxyDeAregion(\tau) \coloneqq \big\{\xi \in \Xi:\,|\hatfn(\xi)-g(\xi)| \ge \tau \big\}\;\subseteq\;\approxDeAregion(\tau_1).
        \]
        where $\tau = \tau_1+\tau_2$.
        The proof can be found in \Cref{appendix:proof_ridea_all}.
        \end{lemma}

        Heuristically, a good proxy function $g$ (i) is aligned with $\barf$, and (ii) maintains disagreement with $\hatfn$ 
        thereby ensuring the proxy approximate de-amplification region is sufficiently large (notice that in the extreme case where $g = \hatfn$, $\proxyDeAregion$ is empty).
        At time step $t$, $g$ is trained to (i) fit the observations collected up until the previous time step $\{ y_i : i \in \{ 1\ldots{}t-1 \} \}$, and (ii) maintain disagreement with $\hatfn$.
        This leads to the following objective:
        \begin{equation} \label{equ:Lg}
        \begin{aligned}
             g = \arg\min_g \mathcal{L}(g) &\coloneqq \frac{1}{n}\sum_{i=1}^n (g(\xi_i)-y_i)^2
             \\&\hspace{-2mm}+  \frac{1}{n}\sum_{i=1}^n \max\!\big(0, \;\tau - |\hatfn(\xi_i)-g(\xi_i)|\big).
        \end{aligned}           
        \end{equation}

        Leveraging the \textit{proxy approximate de-amplifying region} and learned $g$, we introduce an acquisition function for the decision-making phase that balances de-amplification with informativeness and representativeness.  The novel acquisition function is given by
        \begin{equation} \label{equ:eig_de_amplify}
            \begin{aligned}
                &\text{\acqfria{}}(\xi_t)  = \text{\acqfri{}}(\xi_t) \, \mathrm{DeA}(\xi_t), \\ 
                & \text{where} \quad \mathrm{DeA}(\xi_t) =
                    \text{Sigmoid }\!\left(\dfrac{|\hatfn(\xi_t)-g(\xi_t)|-\tau}{\kappa}\right),
            \end{aligned}
        \end{equation}
        where $\tau$ is as defined in \Cref{theor:worstcase} and 
        further details of $\mathrm{DeA}$ derivation are provided in \Cref{appendix:proof_ridea_all}. $\text{\acqfria{}}$ requires two hyperparameters: $\tau$ corresponds to the de-amplifying threshold, i.e., to the minimum separation level required for a design to be considered in the proxy de-amplifying region, and $\kappa$ corresponds to the optional smoothing parameter\footnote{We fixed $\kappa =1$ in our experiments.}.
        The hyperparameter $\tau$ controls the tradeoff between de-amplification and the requirements of informativeness and representativeness: When $\tau$ is large, the proxy approximate de-amplifying region is conservative in the sense that it includes only designs with very high $\widehat{\amplification}$; DeA's downweighting of most designs reflects their effective exclusion from this region.

        More generally, \Cref{equ:eig_de_amplify} can be framed as a framework, or family, of acquisition functions.
        While we specified R-I using the EIG and robust ratio shown in \Cref{equ:acqfri} to measure the degrees of informativeness and representativeness, respectively, of a design, one could easily substitute these terms with problem- or application-specific measures.

\section{EXPERIMENTS}\label{sec:experiments}
    
    This section contains comparative experiments and analysis to explore which algorithm performs best in the presence of model misspecification in three experimental paradigms: a polynomial regression experiment, a source location paradigm, and a pharmacokinetic setting.
    We also empirically validate the theoretical results of Section \ref{sec:theory}.
    The code to reproduce our experiments is available at \href{https://github.com/TrbingWY/robust-boed.git}{https://github.com/TrbingWY/robust-boed.git}

    We compare the following methods: A \textbf{Random} strategy selects designs from the test distribution at random.
    \textbf{Bayesian adaptive design (BAD)} \citep{foster2019variational} selects designs according to the traditional BOED strategy, i.e., according to the EIG (\Cref{eq:eig}).
    \textbf{Representative and informative BAD (\acqfri{})} selects designs according to our novel representative acquisition function (\Cref{sec:1-idea}).
    \textbf{Representative, informative and de-amplifying BAD (\acqfria{})} selects designs according to our novel de-amplifying acquisition function (\Cref{sec:r-idea}).
    To explore how the hyperparameters affect our proposed acquisition function, we conduct experiments with different values of $\lambda$ in \acqfri{}  and $\tau$ in \acqfria{}.
    These results are given in \Cref{appendix:appendix_ex_results}.

    The generalization performance of each method is evaluated using the Mean Squared Error (MSE), while the degree of covariate shift is measured by the Maximum Mean Discrepancy (MMD). Further details are provided in \Cref{appendix:appendix_ex_detail}.

\subsection{Polynomial Regression Experiments} \label{subsec:toy}
    In the \textbf{polynomial regression setting}, the DGP is a degree-two polynomial regression model, $y=1+2 x-0.5x^2+\epsilon$ where $\epsilon \sim \mathcal{N}(0,0.1)$. 
    In the \textit{misspecified case}, we use a linear model to fit the data this model generates.
    In the \textit{well-specified case}, we use a quadratic model to fit the data this model generates.
    More implementational details are given in \Cref{appendix:appendix_ex_detail_toy}, and more experimental results (one of which is in the well-specified case) can be found in \Cref{appendix:ex_toy}.
    In particular, we also explore the performance of our proposed acquisition functions under different misspecification degrees in \Cref{appendix:poly_high_mis,appendix:poly_differ_mis}.

\begin{figure}[h]
            \centering
            \begin{subfigure}[b]{0.65\linewidth}
                \centering
                \includegraphics[width=\linewidth]{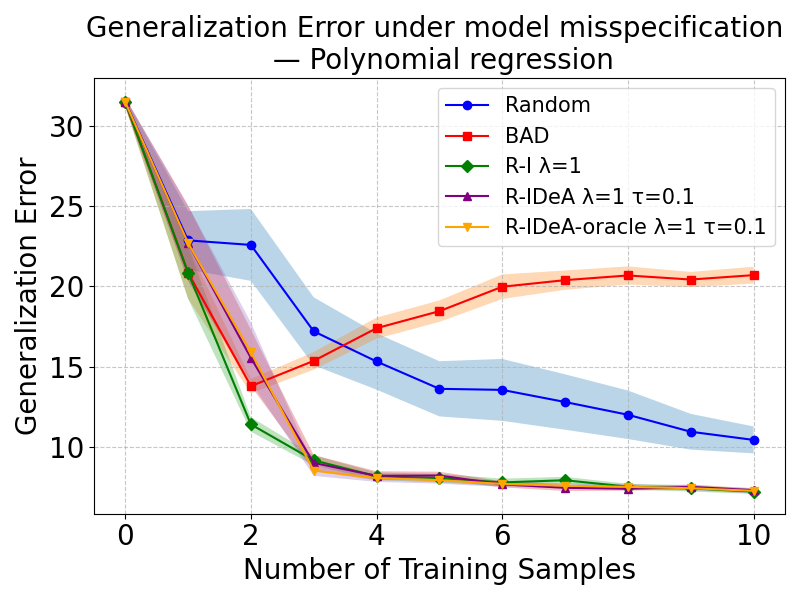}
                \caption{}
                \label{subfig:toy_base_misresults_Gerror}
            \end{subfigure}
            \begin{subfigure}[b]{0.6\linewidth}
                \centering
                \includegraphics[width=\linewidth]{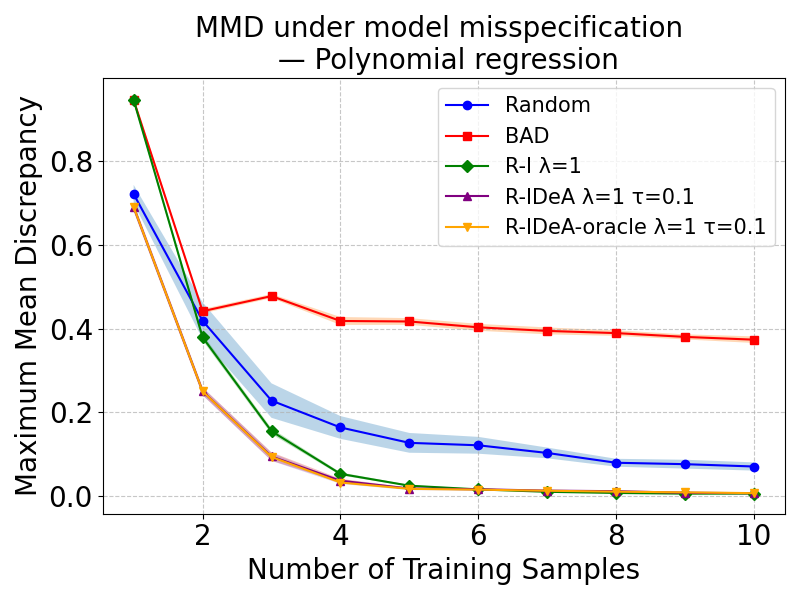}
                \caption{}
                \label{subfig:toy_base_misresults_mmd}
            \end{subfigure}
        \caption{\textbf{Polynomial regression experiments (misspecified case).}
        Comparison of different design strategies (Random, BAD, proposed \acqfri{}, proposed \acqfria{}, and \acqfria{}-oracle, which uses the true $\barf$ instead of the proxy $g$) under misspecified models in the polynomial regression experiments.
        \textit{Left}: Generalization error across methods.
        \textit{Right}: MMD distance across methods; higher values indicate a greater degree of covariate shift.}
        \label{fig:toy_base_misresults_mis}
    \end{figure}

    \autoref{fig:toy_base_misresults_mis} shows that under model misspecification,
    \acqfri{} outperforms both BAD and Random, indicating that incorporating informativeness and representativeness leads to more effective design selection. 
    \acqfria{} further improves generalization performance and achieves the best results overall, demonstrating that jointly accounting for informativeness, representativeness, and de-amplification is most effective. 
    Finally, we compare the performance of \acqfria{} and \acqfria{}-oracle, which uses the true $\barf$ instead of the proxy $g$.
    \acqfria{} performs comparably to \acqfria{}-oracle, implying that \acqfria{} is effective even while relying on the proxy $g$ to approximate $\barf$.

\subsection{Source Localization Experiments}
    The \textbf{acoustic energy attenuation model} 
        simulates the total intensity at location $\xi$ of a signal emitted from multiple sources at locations $\theta = \{ \theta_k \}_{k=1}^{K}$.
        The objective of the design problem is to strategically select points at which to observe the total signal to infer the locations of the source effectively. More implementational details can be found in \Cref{appendix:appendix_ex_detail_location}, and more experimental results are in \Cref{appendix:ex_source}.

\begin{figure}[h]  
                    \centering 
                    \begin{subfigure}[b]{0.65\linewidth}
                        \centering
                        \includegraphics[width=\linewidth]{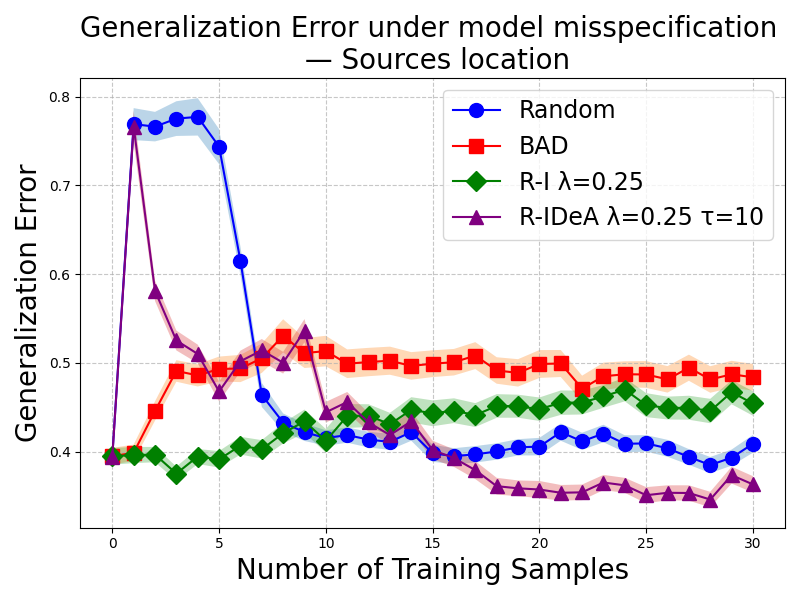}
                        \caption{}
                        \label{subfig:source_overall_misresults_gerror}
                    \end{subfigure}
                    \begin{subfigure}[b]{0.6\linewidth}
                        \centering
                        \includegraphics[width=\linewidth]{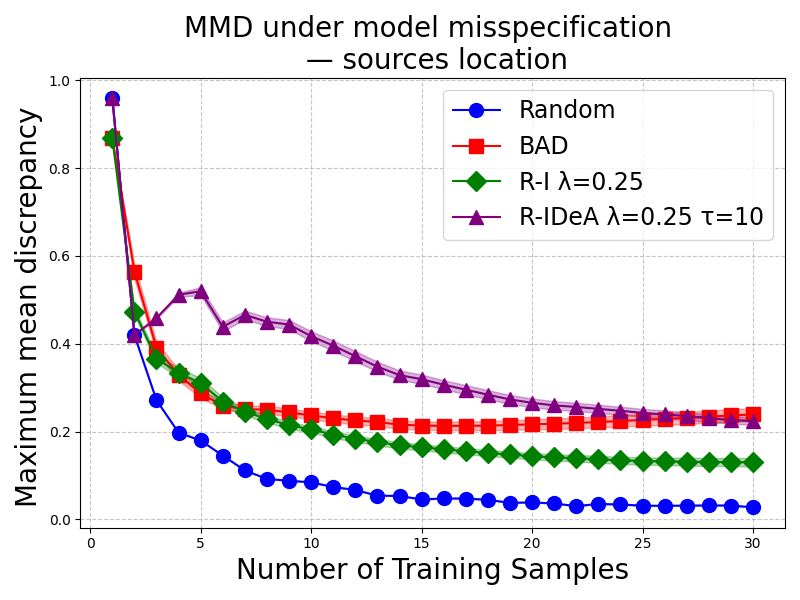}
                        \caption{}
                        \label{subfig:source_overall_misresult_mmd}
                    \end{subfigure}
                    \caption{\textbf{Source localization experiments (misspecified case).} Comparison of baseline methods (Random, BAD) and our proposed \acqfri{} and \acqfria{} in the source localization experiments.
                    \textit{Top}: Generalization error across methods.
                    \textit{Bottom}: MMD distance across methods; higher values indicate a greater degree of covariate shift.}
                    \label{fig:source_overall_misresults}
            \end{figure}

        Interestingly, under model misspecification, we observe that the generalization error of  BAD and \acqfri{} slightly increases (\Cref{subfig:source_overall_misresults_gerror}) while the degree of covariate shift these methods induce decreases (\Cref{subfig:source_overall_misresult_mmd}).
        We speculate that this discrepancy is due to the amplifying properties of the designs selected by these methods.
        In the presence of model misspecification,
        \acqfri{} both outperforms BAD and induces less covariate shift, highlighting the effectiveness of representative designs.    
        \acqfria{} performs better than other methods and induces the highest covariate shift.
        This implies that covariate shift is not the only factor influencing generalization error, and that \acqfria{} may perform better than other methods due to its selection of de-amplifying training data.
    
    \subsection{Pharmacokinetic Experiments}
         According to the \textbf{pharmacokinetic model}, the distribution of an administered drug in the body is determined by three key parameters: the absorption rate $k_\alpha$, the elimination rate $k_e$, and the volume $V$. These define the parameter vector of interest, $\theta = (k_\alpha, k_e, V)$. The design task is to adaptively select blood sampling times $0 \leq \xi_t \leq 24$ hours for each patient, measured from the moment of drug administration (with patient 2 receiving the drug only after collecting a sample from patient 1, and so on).
         More implementational details can be found in \Cref{appendix:appendix_ex_detail_pk}, and more experimental results can be found in \Cref{appendix:ex_pk}. 

         \Cref{subfig:pk_overall_misresults_gerror} shows that, under model misspecification, \acqfria{} exhibits the best performance, suggesting that the selection of de-amplifying and representative designs can help reduce generalization error, consistent with the theoretical result established in \Cref{prop:gen_error_covshift}.
         \Cref{subfig:pk_overall_misresult_mmd} illustrates that in addition to exhibiting the best generalization performance, \acqfria{} induces the largest degree of covariate shift, further demonstrating that designs that are de-amplifying in addition to representative contribute to robustness under model misspecification.
        
         \begin{figure}[h]  
                    \centering 
                    \begin{subfigure}[b]{0.65\linewidth}
                        \centering
                        \includegraphics[width=\linewidth]{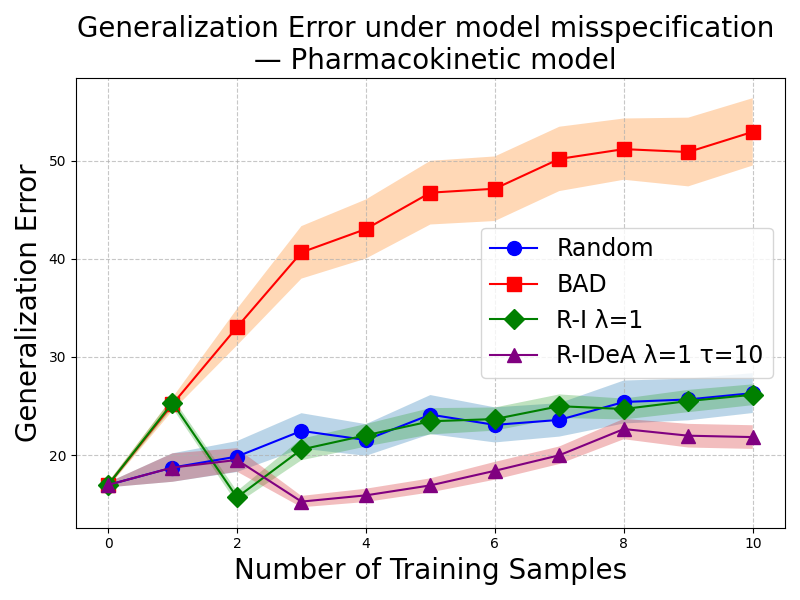}
                        \caption{}
                        \label{subfig:pk_overall_misresults_gerror}
                    \end{subfigure}
                    \begin{subfigure}[b]{0.6\linewidth}
                        \centering
                        \includegraphics[width=\linewidth]{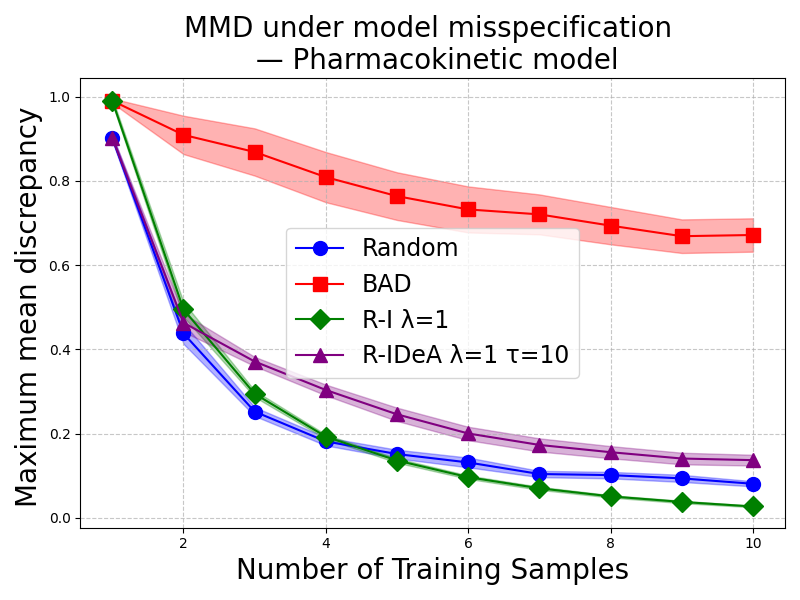}
                        \caption{}
                        \label{subfig:pk_overall_misresult_mmd}
                    \end{subfigure}
                    \caption{\textbf{Pharmacokinetic model experiments (misspecified case).} Comparison of baseline methods (Random, BAD) and our proposed \acqfri{} and \acqfria{} in the Pharmacokinetic model experiments
                    \textit{Top}: Generalization error across methods.
                    \textit{Bottom}: MMD distance across methods; higher values indicate a greater degree of covariate shift.}
                    \label{fig:pk_overall_misresults}
            \end{figure}
     
\section{CONCLUSION}\label{sec:conclusion}
    When models are correctly specified, powerful experimental designs are \textnormal{informative} about a parameter of interest.
    When models are misspecified, effective experimental designs are informative and robust to the misspecification.
    Our detailed analysis unpacks what is required for robustness.
    Our analysis reveals that robustness is a function of designs' \textnormal{representativeness} of the test distribution and \textnormal{de-amplification} of misspecification errors.
    We leverage these insights to propose a novel method for BOED in the presence of potential model misspecification. Our empirical results demonstrate the effectiveness of the proposed method.

\textbf{Limitations and Future Work}
    Our proposed method is informed by insights from \Cref{prop:gen_error_covshift}, which provides an upper bound on generalization performance.
    The degree to which \Cref{prop:gen_error_covshift} reflects actual generalization performance depends on the tightness of this bound.
    Assessing the tightness of this bound is therefore an important direction for future work.
    Moreover, additional empirical results in \Cref{appendix:appendix_ex_results} suggest that the properties of de-amplification, representativeness, and informativeness are not independent. Consequently, tuning a hyperparameter to control one property will implicitly affect the others. This highlights the importance of selecting hyperparameters automatically and appropriately, rather than simply increasing or decreasing their values in a heuristic manner. Developing principled methods for automatic hyperparameter selection is an important direction for future work.

\section*{Acknowledgements}
The authors thank Zhang Wan and Xiaomei Mi for their helpful discussion.
This work was supported by EU grant (101120237 and ERC ODD-ML 101201120) and the Research Council of Finland Flagship programme: Finnish Center for Artificial Intelligence FCAI and decisions 359207, 359567, 358958. SJS and SK were supported by the UKRI Turing AI World-Leading Researcher Fellowship, [EP/W002973/1].

\newpage
\bibliographystyle{plainnat}
\bibliography{reference}

@article{chaloner1995bayesian,
  title={Bayesian experimental design: A review},
  author={Chaloner, Kathryn and Verdinelli, Isabella},
  journal={Statistical science},
  pages={273--304},
  year={1995},
  publisher={JSTOR}
}

@misc{hastie2009elements,
  title={The elements of statistical learning: data mining, inference, and prediction},
  author={Hastie, Trevor},
  year={2009},
  publisher={Springer}
}

@article{huan2024optimal,
  title={Optimal experimental design: Formulations and computations},
  author={Huan, Xun and Jagalur, Jayanth and Marzouk, Youssef},
  journal={Acta Numerica},
  volume={33},
  pages={715--840},
  year={2024},
  publisher={Cambridge University Press}
}

@inproceedings{wen2014robust,
  title={Robust learning under uncertain test distributions: Relating covariate shift to model misspecification},
  author={Wen, Junfeng and Yu, Chun-Nam and Greiner, Russell},
  booktitle={International Conference on Machine Learning},
  pages={631--639},
  year={2014},
  organization={PMLR}
}

@article{sugiyama2007direct,
  title={Direct importance estimation with model selection and its application to covariate shift adaptation},
  author={Sugiyama, Masashi and Nakajima, Shinichi and Kashima, Hisashi and Buenau, Paul and Kawanabe, Motoaki},
  journal={Advances in neural information processing systems},
  volume={20},
  year={2007}
}

@inproceedings{amortila2024mitigating,
  title={Mitigating covariate shift in misspecified regression with applications to reinforcement learning},
  author={Amortila, Philip and Cao, Tongyi and Krishnamurthy, Akshay},
  booktitle={The Thirty Seventh Annual Conference on Learning Theory},
  pages={130--160},
  year={2024},
  organization={PMLR}
}

@article{sloman2022characterizing,
  title={Characterizing the robustness of Bayesian adaptive experimental designs to active learning bias},
  author={Sloman, Sabina J and Oppenheimer, Daniel M and Broomell, Stephen B and Shalizi, Cosma Rohilla},
  journal={arXiv preprint arXiv:2205.13698},
  year={2022}
}

@article{caprio2023credal,
  title={Credal Bayesian Deep Learning},
  author={Caprio, Michele and Dutta, Souradeep and Jang, Kuk Jin and Lin, Vivian and Ivanov, Radoslav and Sokolsky, Oleg and Lee, Insup},
  journal={arXiv e-prints},
  pages={arXiv--2302},
  year={2023}
}

@article{grunwald2017inconsistency,
  title={Inconsistency of Bayesian inference for misspecified linear models, and a proposal for repairing it},
  author={Gr{\"u}nwald, Peter and Van Ommen, Thijs},
  year={2017}
}

@inproceedings{catanach2023metrics,
  title={Metrics for bayesian optimal experiment design under model misspecification},
  author={Catanach, Tommie A and Das, Niladri},
  booktitle={2023 62nd IEEE Conference on Decision and Control (CDC)},
  pages={7707--7714},
  year={2023},
  organization={IEEE}
}

@article{nott2023bayesian,
  title={Bayesian inference for misspecified generative models},
  author={Nott, David J and Drovandi, Christopher and Frazier, David T},
  journal={Annual Review of Statistics and Its Application},
  volume={11},
  year={2023},
  publisher={Annual Reviews}
}

@inproceedings{schmitt2023detecting,
  title={Detecting model misspecification in amortized Bayesian inference with neural networks},
  author={Schmitt, Marvin and B{\"u}rkner, Paul-Christian and K{\"o}the, Ullrich and Radev, Stefan T},
  booktitle={DAGM German Conference on Pattern Recognition},
  pages={541--557},
  year={2023},
  organization={Springer}
}

@article{vincent2017darc,
  title={The DARC Toolbox: automated, flexible, and efficient delayed and risky choice experiments using Bayesian adaptive design},
  author={Vincent, Benjamin T and Rainforth, Tom},
  journal={PsyArXiv. October},
  volume={20},
  year={2017}
}

@article{huang2023learning,
  title={Learning robust statistics for simulation-based inference under model misspecification},
  author={Huang, Daolang and Bharti, Ayush and Souza, Amauri and Acerbi, Luigi and Kaski, Samuel},
  journal={Advances in Neural Information Processing Systems},
  volume={36},
  pages={7289--7310},
  year={2023}
}

@article{overstall2022bayesian,
  title={Bayesian decision-theoretic design of experiments under an alternative model},
  author={Overstall, Antony and McGree, James},
  journal={Bayesian Analysis},
  volume={17},
  number={4},
  pages={1021--1041},
  year={2022},
  publisher={International Society for Bayesian Analysis}
}

@article{overstall2025gibbs,
  title={Gibbs optimal design of experiments},
  author={Overstall, Antony M and Holloway-Brown, Jacinta and McGree, James M},
  journal={arXiv preprint arXiv:2310.17440},
  year={2025}
}

@InProceedings{forster2025improving,
    title={Improving Robustness to Model Misspecification in Bayesian Experimental Design},
    author={Forster, Alexander J and Ivanova, Desi R and Rainforth, Tom},
    booktitle={Workshop at the 7th Symposium on Advances in Approximate Bayesian Inference},
    year={2025}
}

@book{boyd2004convex,
  title={Convex optimization},
  author={Boyd, Stephen P and Vandenberghe, Lieven},
  year={2004},
  publisher={Cambridge university press}
}

@article{oberauer2025variance,
  title={Variance, bias, and computational cost of estimating the Bayes factor using bridge sampling and the Savage-Dickey density ratio},
  author={Oberauer, Klaus and Aust, Frederik and Musfeld, Philipp},
  year={2025}
}

@article{frazier2023reliable,
  title={Reliable Bayesian inference in misspecified models},
  author={Frazier, David T and Kohn, Robert and Drovandi, Christopher and Gunawan, David},
  journal={arXiv preprint arXiv:2302.06031},
  year={2023}
}

@article{paszke2019pytorch,
  title={Pytorch: An imperative style, high-performance deep learning library},
  author={Paszke, A},
  journal={arXiv preprint arXiv:1912.01703},
  year={2019}
}

@article{bingham2019pyro,
  title={Pyro: Deep universal probabilistic programming},
  author={Bingham, Eli and Chen, Jonathan P and Jankowiak, Martin and Obermeyer, Fritz and Pradhan, Neeraj and Karaletsos, Theofanis and Singh, Rohit and Szerlip, Paul and Horsfall, Paul and Goodman, Noah D},
  journal={Journal of machine learning research},
  volume={20},
  number={28},
  pages={1--6},
  year={2019}
}

@article{gretton2012kernel,
  title={A kernel two-sample test},
  author={Gretton, Arthur and Borgwardt, Karsten M and Rasch, Malte J and Sch{\"o}lkopf, Bernhard and Smola, Alexander},
  journal={The Journal of Machine Learning Research},
  volume={13},
  number={1},
  pages={723--773},
  year={2012},
  publisher={JMLR. org}
}

@article{viehmann2021partial,
  title={Partial wasserstein and maximum mean discrepancy distances for bridging the gap between outlier detection and drift detection},
  author={Viehmann, Thomas},
  journal={arXiv preprint arXiv:2106.12893},
  year={2021}
}

@inproceedings{bharti2023optimally,
  title={Optimally-weighted estimators of the maximum mean discrepancy for likelihood-free inference},
  author={Bharti, Ayush and Naslidnyk, Masha and Key, Oscar and Kaski, Samuel and Briol, Fran{\c{c}}ois-Xavier},
  booktitle={International Conference on Machine Learning},
  pages={2289--2312},
  year={2023},
  organization={PMLR}
}

@article{farquhar2021statistical,
  title={On statistical bias in active learning: How and when to fix it},
  author={Farquhar, Sebastian and Gal, Yarin and Rainforth, Tom},
  journal={arXiv preprint arXiv:2101.11665},
  year={2021}
}

@article{konyushkova2017learning,
  title={Learning active learning from data},
  author={Konyushkova, Ksenia and Sznitman, Raphael and Fua, Pascal},
  journal={Advances in neural information processing systems},
  volume={30},
  year={2017}
}

@book{silberman1996active,
  title={Active learning: 101 strategies to teach any subject.},
  author={Silberman, Mel},
  year={1996},
  publisher={ERIC}
}

@article{settles2009active,
  title={Active learning literature survey},
  author={Settles, Burr},
  year={2009},
  journal= {University of Wisconsin-Madison Department of Computer Sciences}
}

@article{kelly2025simulation,
  title={Simulation-based Bayesian inference under model misspecification},
  author={Kelly, Ryan P and Warne, David J and Frazier, David T and Nott, David J and Gutmann, Michael U and Drovandi, Christopher},
  journal={arXiv preprint arXiv:2503.12315},
  year={2025}
}

@article{park2013bayesian,
  title={Bayesian active learning for drug combinations},
  author={Park, Mijung and Nassar, Marcel and Vikalo, Haris},
  journal={IEEE transactions on biomedical engineering},
  volume={60},
  number={11},
  pages={3248--3255},
  year={2013},
  publisher={IEEE}
}

@book{shalev2014understanding,
  title={Understanding machine learning: From theory to algorithms},
  author={Shalev-Shwartz, Shai and Ben-David, Shai},
  year={2014},
  publisher={Cambridge university press}
}

@article{bonhomme2022minimizing,
  title={Minimizing sensitivity to model misspecification},
  author={Bonhomme, St{\'e}phane and Weidner, Martin},
  journal={Quantitative Economics},
  volume={13},
  number={3},
  pages={907--954},
  year={2022},
  publisher={Wiley Online Library}
}

@article{thompson2023integrating,
  title={Integrating a tailored recurrent neural network with Bayesian experimental design to optimize microbial community functions},
  author={Thompson, Jaron C and Zavala, Victor M and Venturelli, Ophelia S},
  journal={PLOS Computational Biology},
  volume={19},
  number={9},
  pages={e1011436},
  year={2023},
  publisher={Public Library of Science San Francisco, CA USA}
}

@article{kreutz2009systems,
  title={Systems biology: experimental design},
  author={Kreutz, Clemens and Timmer, Jens},
  journal={The FEBS journal},
  volume={276},
  number={4},
  pages={923--942},
  year={2009},
  publisher={Wiley Online Library}
}

@article{hickman2022bayesian,
  title={Bayesian optimization with known experimental and design constraints for chemistry applications},
  author={Hickman, Riley J and Aldeghi, Matteo and H{\"a}se, Florian and Aspuru-Guzik, Al{\'a}n},
  journal={Digital Discovery},
  volume={1},
  number={5},
  pages={732--744},
  year={2022},
  publisher={Royal Society of Chemistry}
}

@article{walker2019bayesian,
  title={Bayesian Design of Experiments: Implementation, Validation and Application to Chemical Kinetics},
  author={Walker, Eric A and Ravisankar, Kishore},
  journal={arXiv preprint arXiv:1909.03861},
  year={2019}
}

@article{lintusaari2017fundamentals,
  title={Fundamentals and recent developments in approximate Bayesian computation},
  author={Lintusaari, Jarno and Gutmann, Michael U and Dutta, Ritabrata and Kaski, Samuel and Corander, Jukka},
  journal={Systematic biology},
  volume={66},
  number={1},
  pages={e66--e82},
  year={2017},
  publisher={Oxford University Press}
}

@article{frazier2020model,
  title={Model misspecification in approximate Bayesian computation: consequences and diagnostics},
  author={Frazier, David T and Robert, Christian P and Rousseau, Judith},
  journal={Journal of the Royal Statistical Society Series B: Statistical Methodology},
  volume={82},
  number={2},
  pages={421--444},
  year={2020},
  publisher={Oxford University Press}
}

@book{sutton1998reinforcement,
  title={Reinforcement learning: An introduction},
  author={Sutton, Richard S and Barto, Andrew G and others},
  volume={1},
  number={1},
  year={1998},
  publisher={MIT press Cambridge}
}

@book{greene2003econometric,
  title={Econometric analysis},
  author={Greene, William H},
  year={2003},
  publisher={Pearson Education India}
}

@book{wooldridge2010econometric,
  title={Econometric analysis of cross section and panel data},
  author={Wooldridge, Jeffrey M},
  year={2010},
  publisher={MIT press}
}

@article{muller2013risk,
  title={Risk of Bayesian inference in misspecified models, and the sandwich covariance matrix},
  author={M{\"u}ller, Ulrich K},
  journal={Econometrica},
  volume={81},
  number={5},
  pages={1805--1849},
  year={2013},
  publisher={Wiley Online Library}
}

@article{box1980sampling,
  title={Sampling and Bayes’ inference in scientific modelling and robustness},
  author={Box, George EP},
  journal={Journal of the Royal Statistical Society Series A: Statistics in Society},
  volume={143},
  number={4},
  pages={383--404},
  year={1980},
  publisher={Oxford University Press}
}

@book{bernardo2009bayesian,
  title={Bayesian theory},
  author={Bernardo, Jos{\'e} M and Smith, Adrian FM},
  volume={405},
  year={2009},
  publisher={John Wiley \& Sons}
}

@article{walker2013bayesian,
  title={Bayesian inference with misspecified models},
  author={Walker, Stephen G},
  journal={Journal of statistical planning and inference},
  volume={143},
  number={10},
  pages={1621--1633},
  year={2013},
  publisher={Elsevier}
}

@article{George1976Science,
 ISSN = {01621459, 1537274X},
 URL = {http://www.jstor.org/stable/2286841},
 journal = {Journal of the American Statistical Association},
 number = {356},
 pages = {791--799},
 publisher = {[American Statistical Association, Taylor & Francis, Ltd.]},
 title = {Science and Statistics},
 author= {George E. P. Box.},
 urldate = {2025-04-09},
 volume = {71},
 year = {1976}
}

@article{uppal2003model,
  title={Model misspecification and underdiversification},
  author={Uppal, Raman and Wang, Tan},
  journal={The Journal of Finance},
  volume={58},
  number={6},
  pages={2465--2486},
  year={2003},
  publisher={Wiley Online Library}
}

@article{sugiyama2005active,
  title={Active learning for misspecified models},
  author={Sugiyama, Masashi},
  journal={Advances in neural information processing systems},
  volume={18},
  year={2005}
}

@article{lindley1956measure,
  title={On a measure of the information provided by an experiment},
  author={Lindley, Dennis V},
  journal={The Annals of Mathematical Statistics},
  volume={27},
  number={4},
  pages={986--1005},
  year={1956},
  publisher={Institute of Mathematical Statistics}
}

@article{myung2013tutorial,
  title={A tutorial on adaptive design optimization},
  author={Myung, Jay I and Cavagnaro, Daniel R and Pitt, Mark A},
  journal={Journal of mathematical psychology},
  volume={57},
  number={3-4},
  pages={53--67},
  year={2013},
  publisher={Elsevier}
}

@article{cavagnaro2010adaptive,
  title={Adaptive design optimization: A mutual information-based approach to model discrimination in cognitive science},
  author={Cavagnaro, Daniel R and Myung, Jay I and Pitt, Mark A and Kujala, Janne V},
  journal={Neural computation},
  volume={22},
  number={4},
  pages={887--905},
  year={2010},
  publisher={MIT Press}
}

@article{dong2024variational,
  title={Variational Bayesian Optimal Experimental Design with Normalizing Flows},
  author={Dong, Jiayuan and Jacobsen, Christian and Khalloufi, Mehdi and Akram, Maryam and Liu, Wanjiao and Duraisamy, Karthik and Huan, Xun},
  journal={arXiv preprint arXiv:2404.13056},
  year={2024}
}

@inproceedings{sebastiani1997bayesian,
  title={Bayesian experimental design and Shannon information},
  author={Sebastiani, Paola and Wynn, Henry P},
  booktitle={Proceedings of the Section on Bayesian Statistical Science},
  volume={44},
  pages={176--181},
  year={1997},
  organization={The Association}
}

@article{foster2019variational,
  title={Variational Bayesian optimal experimental design},
  author={Foster, Adam and Jankowiak, Martin and Bingham, Elias and Horsfall, Paul and Teh, Yee Whye and Rainforth, Thomas and Goodman, Noah},
  journal={Advances in Neural Information Processing Systems},
  volume={32},
  year={2019}
}

@article{rainforth2024modern,
  title={Modern Bayesian experimental design},
  author={Rainforth, Tom and Foster, Adam and Ivanova, Desi R and Bickford Smith, Freddie},
  journal={Statistical Science},
  volume={39},
  number={1},
  pages={100--114},
  year={2024},
  publisher={Institute of Mathematical Statistics}
}

@inproceedings{foster2021deep,
  title={Deep adaptive design: Amortizing sequential bayesian experimental design},
  author={Foster, Adam and Ivanova, Desi R and Malik, Ilyas and Rainforth, Tom},
  booktitle={International conference on machine learning},
  pages={3384--3395},
  year={2021},
  organization={PMLR}
}

@article{ge2023maximum,
  title={Maximum likelihood estimation is all you need for well-specified covariate shift},
  author={Ge, Jiawei and Tang, Shange and Fan, Jianqing and Ma, Cong and Jin, Chi},
  journal={arXiv preprint arXiv:2311.15961},
  year={2023}
}

@article{ivanova2024step,
  title={Step-DAD: Semi-Amortized Policy-Based Bayesian Experimental Design},
  author={Ivanova, Desi R and Hedman, Marcel and Guan, Cong and Rainforth, Tom},
  year={2024}
}

@inproceedings{ali2014active,
    title={Active Learning with Model Selection},
    author={Alnur Ali and Rich Caruana and Ashish Kapoor},
    booktitle={Proceedings of the Twenty-Eighth AAAI Conference on Artificial Intelligence},
    year={2014}
}

@article{barlas2025robust,
  title={Robust Experimental Design via Generalised Bayesian Inference},
  author={Barlas, Yasir Zubayr and Sloman, Sabina J and Kaski, Samuel},
  journal={arXiv preprint arXiv:2511.07671},
  year={2025}
}

@techreport{bach2006active,
    title={Active learning for misspecified generalized linear models},
    author={Francis Bach},
    institution={Ecole des mines de Paris},
    year={2006},
    number={N15/06/MM}
}

@article{dubova2025ockhams,
    title={Is Ockham's razor losing its edge? New perspectives on the principle of model parsimony},
    author={Marina Dubova and Suyog Chandramouli and Gerd Gigerenzer and Peter Gr\"{u}nwald and William Holmes and Tania Lombrozo and Marco Marelli and Sebastian Musslick and Bruno Nicenboim and Lauren N. Ross and Richard Shiffrin and Martha White and Eric-Jan Wagenmakers and Paul-Christian B\"{u}rkner and Sabina J. Sloman},
    journal={PNAS},
    year={2025}
}

@article{kleijn2006misspecification,
    title={Misspecification in Infinite-dimensional Bayesian Statistics},
    author={B. J. K. Kleijn and A. W. {van der Vaart}},
    journal={The Annals of Statistics},
    volume={34},
    number={2},
    year={2006}
}

@article{kleijn2012bernsteinvonmises,
    title={The Bernstein-Von-Mises theorem under misspecification},
    author={B.J.K. Kleijn and A.W. {van der Vaart}},
    journal={Electronic Journal of Statistics},
    volume={6},
    year={2012}
}

@article{knoblauch2022optimizationcentric,
    title={An Optimization-centric View on Bayes' Rule: Reviewing and Generalizing Variational Inference},
    author={Jeremias Knoblauch and Jack Jewson and Theodoros Damoulas},
    journal={Journal of Machine Learning Research},
    volume={23},
    year={2022}
}

\newpage
\clearpage
\section*{Checklist}

\begin{enumerate}

  \item For all models and algorithms presented, check if you include:
  \begin{enumerate}
    \item A clear description of the mathematical setting, assumptions, algorithm, and/or model. [Yes]
    \item An analysis of the properties and complexity (time, space, sample size) of any algorithm. [Yes]
    \item (Optional) Anonymized source code, with specification of all dependencies, including external libraries. [Yes/No/Not Applicable]
  \end{enumerate}

  \item For any theoretical claim, check if you include:
  \begin{enumerate}
    \item Statements of the full set of assumptions of all theoretical results. [Yes]
    \item Complete proofs of all theoretical results. [Yes]
    \item Clear explanations of any assumptions. [Yes]     
  \end{enumerate}

  \item For all figures and tables that present empirical results, check if you include:
  \begin{enumerate}
    \item The code, data, and instructions needed to reproduce the main experimental results (either in the supplemental material or as a URL). [Yes]
    \item All the training details (e.g., data splits, hyperparameters, how they were chosen). [Yes]
    \item A clear definition of the specific measure or statistics and error bars (e.g., with respect to the random seed after running experiments multiple times). [Yes]
    \item A description of the computing infrastructure used. (e.g., type of GPUs, internal cluster, or cloud provider). [Yes]
  \end{enumerate}

  \item If you are using existing assets (e.g., code, data, models) or curating/releasing new assets, check if you include:
  \begin{enumerate}
    \item Citations of the creator If your work uses existing assets. [Yes]
    \item The license information of the assets, if applicable. [Not Applicable]
    \item New assets either in the supplemental material or as a URL, if applicable. [Not Applicable]
    \item Information about consent from data providers/curators. [Not Applicable]
    \item Discussion of sensible content if applicable, e.g., personally identifiable information or offensive content. [Not Applicable]
  \end{enumerate}

  \item If you used crowdsourcing or conducted research with human subjects, check if you include:
  \begin{enumerate}
    \item The full text of instructions given to participants and screenshots. [Not Applicable]
    \item Descriptions of potential participant risks, with links to Institutional Review Board (IRB) approvals if applicable. [Not Applicable]
    \item The estimated hourly wage paid to participants and the total amount spent on participant compensation. [Not Applicable]
  \end{enumerate}

\end{enumerate}

\newpage

\clearpage
\appendix
\thispagestyle{empty}
\onecolumn
\aistatstitle{Supplementary Materials}
\section{PROOF of Proposition \ref{prop:gen_de}} \label{appendix:proof_thm1}

    \begin{equation}
        \begin{aligned}
            R_{\mathrm{test}}(f) &:= \mathbb{E}_{\xi \sim d_{\text{test}}}\left[(f(\xi) - \starf(\xi))^2\right] \\
            & = \mathbb{E}_{\xi \sim d_{\text{test}}}\left( \left( \left[ f(\xi) - \barf(\xi) \right] + \left [\barf(\xi) - \starf(\xi) \right] \right)^2\right) \\
            &= \underbrace{\mathbb{E}_{\xi \sim \dtest}\left[(\barf(\xi) - \starf(\xi))^2\right]}_{\text{Misspecification Bias}} 
                    + \underbrace{\mathbb{E}_{\xi \sim \dtest}\left[(f(\xi) - \barf(\xi))^2\right]}_{\text{Estimation Bias}} \\
            &\quad + \underbrace{2 \, \mathbb{E}_{\xi \sim \dtest}\left[(\barf(\xi) - \starf(\xi))(f(\xi) - \barf(\xi))\right]}_{\text{Error (De-)amplification}}.
        \end{aligned}
    \end{equation}
    
\section{PROOF of \Cref{prop:gen_error_covshift}} \label{appendix:proof_thm2}
    Theorem: with probability at least $1 - \delta$, the generalization error under covariate shift satisfies:
    \begin{equation}
        R_{\text{test}}(\hatfn) \leq \maxval{\covshift} \cdot \left\{
            \begin{aligned}
            & \misspecification^2 
                 + \frac{224 \maxval{\outcome}^2 \log (|\mathcal{F}| / \delta)}{3n}
                 - 2\widehat{\amplification}(\hatfn) , & \text{if } \widehat{\amplification} < 0 \\
            & 
                 \misspecification^2 + \frac{128\maxval{\outcome}^2\log (|\mathcal{F}| / \delta)}{3n} - \sqrt{3}\widehat{\amplification}(\hatfn)  , & \text{if } \widehat{\amplification}\geq 0
            \end{aligned}
            \right.
    \end{equation}
                
    \paragraph{Proof of Theorem}
        We provide the full derivation of the generalization error bound that preserves the error (de-)amplification term. Our analysis, which handles misspecification, is adapted from the proof of Proposition 2.1 in \citep{amortila2024mitigating}. In the proof, we first analyse the empirical error considering the training data distribution $d_{\text{train}}$, and then use the bounded density ratio to analyse the generalization error in the testing data distribution $d_{
        \text{test}}$.
    
    The goal is to bound the generalization risk
    \[
    R(f) := \mathbb{E}_{\xi}[(f(\xi) - \starf(\xi))^2],
    \]
    for any $f \in \mathcal{F}$. And the empirical risk is defined as:
    \[
     \quad \widehat{R}(f):=\frac{1}{n} \sum_{i=1}^n\left(f\left(\xi_i\right)-y_i\right)^2,
    \]
    
    Observe that conditional on any $\xi$ we have:
    \begin{equation}
    \begin{aligned}
        & \widehat{R}(f) -\widehat{R}(\barf) \\
        & = \mathbb{E}[(f\left.(\xi)-y)^2-(\barf(\xi)-y)^2 \mid \xi\right] \\
        &=\mathbb{E}\left[(f(\xi)-y)^2-\left(\barf(\xi)-\starf(\xi)+\starf(\xi)-y\right)^2 \mid \xi \right] \\
        & =\mathbb{E}\left[(f(\xi)-y)^2-\left(\barf(\xi)-\starf(\xi)\right)^2-2\left(\barf(\xi)-\starf(\xi)\right)\left(\starf(\xi)-y\right)-\left(\starf(\xi)-y\right)^2 \mid \xi \right] \\
        & =\mathbb{E}\left[(f(\xi)-y)^2-\left(\barf(\xi)-\starf(\xi)\right)^2-\left(\starf(\xi)-y\right)^2 \mid \xi\right] \\
        & =f(\xi)^2-\starf(\xi)^2-2 \mathbb{E}_{\text {train }}[y \mid \xi ]\left(f(\xi)-\starf(\xi)\right)-\left(\barf(\xi)-\starf(\xi)\right)^2 \\
        &=\left(f(\xi)-\starf(\xi)\right)^2-\left(\barf(\xi)-\starf(\xi)\right)^2 \quad (\text{since}\quad \mathbb{E}_{\text {train }}[y \mid \xi ] = \starf(\xi))
    \end{aligned}
    \end{equation}

    thus 
    $$
     \mathbb{E} (\widehat{R}(f) -\widehat{R}(\barf) ) = \mathbb{E}(\left(f(\xi)-\starf(\xi)\right)^2-\left(\barf(\xi)-\starf(\xi)\right)^2 ) = R(f) -R(\barf)
    $$
    
    And
    \begin{equation} \label{equ:var}
    \begin{aligned}
        &\operatorname{Var} \left[(f(\xi)-y)^2-(\barf(\xi)-y)^2\right]  \\
        & = \mathbb{E}\left[\left((f(\xi)-y)^2-(\barf(\xi)-y)^2\right)^2\right] - \left(\mathbb{E}\left[\left((f(\xi)-y)^2-(\barf(\xi)-y)^2\right)\right] \right)^2 \quad (\text{since} \quad \operatorname{Var}(X) = \mathbb{E}(X^2) - (\mathbb{E}X)^2 )\\
        &\leq \mathbb{E}\left[\left((f(\xi)-y)^2-(\barf(\xi)-y)^2\right)^2\right] \\
        & = \mathbb{E}\left[\left( f^2(\xi) -2 f(\xi)y -\barf^2(\xi) + 2y \barf(\xi)   \right )^2\right] \\
        & =\mathbb{E}\left[(f(\xi)-\barf(\xi))^2(f(\xi)+\barf(\xi)-2 y)^2\right]  \\ 
        & \leq 16 \maxval{\outcome}^2 \mathbb{E}\left[(f(\xi)-\barf(\xi))^2\right]   \quad \text{(since } |f(\xi)|, |\bar{f}(\xi)|, |\outcome| \leq \maxval{\outcome} \text{)} \\  
        & = 16 \maxval{\outcome}^2\mathbb{E}\left[(f(\xi)- \starf(\xi) -(\barf(\xi)-\starf(\xi) ))^2\right] \\
        & = 16 \maxval{\outcome}^2\mathbb{E}\left[\left(f(\xi)-\starf(\xi)\right)^2+\left(\barf(\xi)-\starf(\xi)\right)^2\right] - 32 \maxval{\outcome}^2\mathbb{E}\left[ (f(\xi) - \starf(\xi)) ( \barf(\xi) -\starf(\xi)) \right]\\
        & =16 \maxval{\outcome}^2\mathbb{E}\left[\left(f(\xi)-\starf(\xi)\right)^2-\left(\barf(\xi)-\starf(\xi)\right)^2 + 2\left(\barf(\xi)-\starf(\xi)\right)^2 \right]  - 32 \maxval{\outcome}^2\mathbb{E}\left[ (f(\xi) - \starf(\xi)) ( \barf(\xi) -\starf(\xi)) \right] \\
        & =16 \maxval{\outcome}^2\mathbb{E}\left[\left(f(\xi)-\starf(\xi)\right)^2-\left(\barf(\xi)-\starf(\xi)\right)^2+2\left(\barf(\xi)-\starf(\xi) \right)^2 \right]  \\
        & \quad- 32 \maxval{\outcome}^2\mathbb{E}\left[ (f(\xi) - \barf(\xi) + \barf(\xi) -\starf(\xi)) ( \barf(\xi) -\starf(\xi)) \right] \\
        & =16 \maxval{\outcome}^2\mathbb{E}\left[\left(f(\xi)-\starf(\xi)\right)^2-\left(\barf(\xi)-\starf(\xi)\right)^2 \right] 
        + 32 \maxval{\outcome}^2 \mathbb{E} \left(\barf(\xi)-\starf(\xi)^2 \right)  \\
        & \quad - 32 \maxval{\outcome}^2\mathbb{E}\left[ (f(\xi) - \barf(\xi)) ( \barf(\xi) -\starf(\xi)) \right]  
        - 32 \maxval{\outcome}^2 \mathbb{E} [(\barf(\xi) -\starf(\xi))^2]\\
        & \leq 16 \maxval{\outcome}^2\mathbb{E}\left[\left(f(\xi)-\starf(\xi)\right)^2-\left(\barf(\xi)-\starf(\xi)\right)^2\right]
        - 32 \maxval{\outcome}^2 \underbrace{\mathbb{E}\left[ (f(\xi) - \barf(\xi)) ( \barf(\xi) -\starf(\xi)) \right]}_{ \widehat{\amplification}(f)} \\
        & = 16 \maxval{\outcome}^2\mathbb{E}\left[\left(f(\xi)-\starf(\xi)\right)^2-\left(\barf(\xi)-\starf(\xi)\right)^2\right] 
        - 32 \maxval{\outcome}^2 \widehat{\amplification}(f) \\
        & =16 \maxval{\outcome}^2 ( R(f) -R(\barf) )  
        - 32 \maxval{\outcome}^2 \widehat{\amplification}(f)
    \end{aligned}
    \end{equation}

    \paragraph{ Application of Bernstein’s Inequality and Union Bound }
    Based on Bernstein’s inequality \citep{shalev2014understanding} and union bound (see Lemma 2.2 in \citep{shalev2014understanding}),
    Bernstein’s inequality becomes:

    $$
        \mathbf{P}\left( \mathbf{E} Z^{(f)} -\frac{1}{n} \sum_{i=1}^n Z^{(f)}_i < \frac{c}{3 n} \log (|\mathcal{F}| / \delta)+\sqrt{\frac{2(\operatorname{Var} Z^{(f)}) \log (|\mathcal{F}| / \delta)}{n}}\right) \geq 1-\delta
    $$

    Setting $Z^{(f)}_{i} = (f(\xi) - y_i)^2 - (\barf(\xi) - y_i)^2$,
    thus $\frac{1}{n} \sum_{i=1}^n Z_i^{(f)} = \widehat{R}(f) - \widehat{R}(\barf); \quad \mathbb{E} Z^{(f)} = R(f) - R(\barf)$ ,
    $| Z^{(f)}-EZ^{(f)}| \leq 2 \sup |Z^{(f)}| \leq 2 \sup (f(\xi) - y_i)^2 - (\barf(\xi) - y_i)^2 =2 \sup (f(\xi) - y_i + \barf(\xi) - y_i)(f(\xi) - \barf(\xi)) = 2 \cdot (4\maxval{\outcome}) (2 \maxval{\outcome}) = c$,
    thus $c = 16 \maxval{\outcome}^2$ 
    
    Now, by using Bernstein's inequality and a union bound over $f \in \mathcal{F}$ that with probability at least $1-\delta$
    \begin{equation}
        \forall f \in \mathcal{F}: R(f)-R(\bar{f})-(\widehat{R}(f)-\widehat{R}(\bar{f})) \leq \sqrt{\frac{\left(16 (R(f)-R(\bar{f}))
         -  32 \widehat{\amplification}(f)\right) 2\maxval{\outcome}^2 \log (|\mathcal{F}| / \delta)}{n}}
        +\frac{16 \maxval{\outcome}^2 \log (|\mathcal{F}| / \delta)}{3 n} .
    \end{equation}
    
    Supppose the 
    \begin{equation} \label{equ:assumption_ERM}
        \widehat{R}\left(\hatfn\right) - \widehat{R}(\barf) \leq 0
    \end{equation}
    we have
    \begin{equation}
    \begin{aligned}
        R\left(\hatfn\right)-R(\bar{f}) 
        & \leq \sqrt{\frac{\left(16  \left(R\left(\hatfn\right)-R(\bar{f})\right)
         -  32 \widehat{\amplification}(\hatfn)\right) 2\maxval{\outcome}^2 \log (|\mathcal{F}| / \delta)}{n}}
        +\frac{16\maxval{\outcome}^2 \log (|\mathcal{F}| / \delta)}{3 n}  
    \end{aligned}
    \end{equation}

    To simplify, make 
    \begin{equation}
    \begin{aligned}
    & A:=  R(\hatfn)-R(\bar{f}) \\
    & C:= \maxval{\outcome}^2\frac{\log (|\mathcal{F}| / \delta)}{n} >0 \\
    \end{aligned}
    \end{equation}

    \paragraph{Why $(16 A -  32 \widehat{\amplification}) \geq 0$}
        {
        \begin{equation}
        \begin{aligned}
            A &:=  R(\hatfn)-R(\bar{f}) =  \mathbb{E}_{\xi}[(\hatfn(\xi) - \starf(\xi))^2]  - \mathbb{E}_{\xi}[(\barf(\xi) - \starf(\xi))^2] \\
            & =  \mathbb{E}_{\xi}[(\hatfn(\xi) - \starf(\xi) + \barf(\xi) - \starf(\xi)) (\hatfn(\xi) - \barf(\xi) )]
        \end{aligned}
        \end{equation}
        Then
        \begin{equation}
        \begin{aligned}
            1/2A -  \widehat{\amplification}
            & =  1/2\mathbb{E}_{\xi}[(\hatfn(\xi) - \starf(\xi) + \barf(\xi) - \starf(\xi)) (\hatfn(\xi) - \barf(\xi) )] -  \mathbb{E}_{\xi}\left[ (\hatfn- \barf(\xi)) ( \barf(\xi) -\starf(\xi)) \right] \\
            & = \mathbb{E}_{\xi}[( 1/2\hatfn(\xi) - 1/2\starf(\xi) + 1/2\barf(\xi) - 1/2\starf(\xi) - \barf(\xi) + \starf(\xi)) (\hatfn(\xi) - \barf(\xi) )] \\
            &= \mathbb{E}_{\xi}[( 1/2\hatfn(\xi) - 1/2\barf(\xi) ) (\hatfn(\xi) - \barf(\xi) )] \geq 0
        \end{aligned}
        \end{equation}
        }

    \subsection { Suppose that $\widehat{\amplification} < 0$ }
         So the above inequality function equals:
        \begin{equation}
        \begin{aligned}
            A & \leq\sqrt{(16 A - 32 \widehat{\amplification}) 2C} + \frac{16C}{3} \\
            & \leq \sqrt{32 A C}  + \sqrt{ - 64 \widehat{\amplification} C}  + \frac{16C}{3}  \quad (\text{since} \sqrt{a+b} < \sqrt{a} + \sqrt{b})\\ 
            & = \sqrt{ A \cdot 32C }  +  \sqrt{-2 \widehat{\amplification} \cdot32 C}  + \frac{16C}{3}\\
            &  \leq \frac{1}{2} A + 16C  -  \widehat{\amplification}  + 16C + \frac{16C}{3}  \quad ( \sqrt{ab} \leq a/2 +b/2 )\\
            & = \frac{1}{2} A  - \widehat{\amplification} + \frac{112C}{3}
        \end{aligned}
        \end{equation}
        so that
        \begin{equation}
            A \leq  -2  \widehat{\amplification} + \frac{224C}{3}
        \end{equation}
        So
        \begin{equation}
             R\left(\hatfn\right)-R(\bar{f})  \leq  - 2\widehat{\amplification}(\hatfn) + \frac{224 \maxval{\outcome}^2 \log (|\mathcal{F}| / \delta)}{3n}
        \end{equation}
        Re-arranging and using that ${R}_{\text{train}}(\barf) \leq \misspecification^2$, thus we have
        \begin{equation}
             R_{\text{train}}\left(\hatfn\right) \leq \misspecification^2 
             - 2\widehat{\amplification}(\hatfn) 
             + \frac{224 \maxval{\outcome}^2 \log (|\mathcal{F}| / \delta)}{3n}
        \end{equation}

        Finally,  let $\maxval{\covshift} := \sup_{\xi \in \Xi} \left| \frac{d_{\text{test}}(\xi)}{d_{\text{train}}(\xi)} \right|$, the upper bound of generalization error can be expressed via the density ratio:
        
        \begin{equation}
            R_{\text{test}}\left( \hatfn \right)
            = \mathbb{E}_{\text{train}} \left[
            \frac{d_{\text{test}}(\xi)}{d_{\text{train}}(\xi)} \left( \hatfn(\xi) - f^*(\xi) \right)^2
            \right]
            \leq \maxval{\covshift} 
            \cdot \left( \misspecification^2 
            -2\widehat{\amplification}(\hatfn)
            + \frac{224 \maxval{\outcome}^2 \log(|\mathcal{F}|/\delta)}{3n} \right)
        \end{equation}
        
        where $ \widehat{\amplification}(\hatfn) = \mathbb{E}_{\text{train}}\left[ (\hatfn- \barf(\xi)) ( \barf(\xi) -\starf(\xi)) \right]$
    
    \subsection { Suppose that $\widehat{\amplification} \geq 0$ }
    Although the second solution, based on solving a quadratic equation, yields a tighter upper bound, it involves more complex derivations. In contrast, the first solution offers a simpler and more interpretable decomposition, while preserving the same theoretical insights into the behavior of the error (de-)amplification term $\widehat{\amplification}(\hatfn)$. Therefore, we adopt the first formulation in the main text and provide the alternative quadratic-based bound in the appendix for completeness.
    
    \subsubsection{Solution 1: First Order Based Upper Bound Derivation}
     Suppose that $\widehat{\amplification} \geq 0$ and $A >0$.
        We use the first-order upper bound for concave functions \citep{boyd2004convex}:
        $$
        \sqrt{a + b} \leq \sqrt{a} + \frac{b}{2\sqrt{a}} \quad \text{for } a > 0,\ a + b \geq 0.
        $$

        Since $(16 A -  32 \widehat{\amplification}) \geq 0$, thus $0 \leq \widehat{\amplification} \leq  1/2 A$

        \begin{equation}
        \begin{aligned}
            A & \leq\sqrt{(16 A  - 32 \widehat{\amplification}) 2C} + \frac{16C}{3} \\
            & \leq \sqrt{32 A C}
            - \frac{64 \widehat{\amplification} C }{2\sqrt{32 A C }} + \frac{16C}{3}  \\ 
            & = \sqrt{ A \cdot 32C }   -  \frac{4\sqrt{2} \widehat{\amplification} \sqrt{C} }{\sqrt{ A }}   + \frac{16C}{3}\\
            &  \leq \frac{1}{2} A + 16C  -   \frac{4 \sqrt{2} \widehat{\amplification} \sqrt{C} }{\sqrt{ A }}  + \frac{16C}{3}  \quad (\text{since} \sqrt{ab} \leq a/2 +b/2 )\\
            & = \frac{1}{2} A -  \frac{4\sqrt{2}  \widehat{\amplification} \sqrt{C} }{\sqrt{ A }}  + \frac{64C}{3} \\
            & \leq \frac{1}{2} A  -  \frac{4\sqrt{2} \widehat{\amplification} \sqrt{C} }{\sqrt{ A_{max} }}  + \frac{64C}{3}
        \end{aligned}
        \end{equation}
    
        so that
        \begin{equation}
        \begin{aligned}
                A &\leq -\frac{8\sqrt{2} \widehat{\amplification} \sqrt{C} }{\sqrt{ A_{max} }} + \frac{128C}{3} \\
                & = -\frac{8\sqrt{2}  \widehat{\amplification} \sqrt{C} }{\sqrt{ \frac{128C}{3} }} + \frac{128C}{3} \quad (\text{make} \widehat{\amplification} =0 , \text{thus} A_{max} < \frac{128C}{3}  )\\
                & =  \frac{128C}{3} - \sqrt{3}\widehat{\amplification} 
        \end{aligned}
        \end{equation}
        thus:
        \begin{equation}
        \begin{aligned}
            R\left(\hatfn\right)-R(\bar{f}) 
            & \leq   \frac{128\maxval{\outcome}^2\log (|\mathcal{F}| / \delta)}{3n} - \sqrt{3}\widehat{\amplification} 
        \end{aligned}
        \end{equation}
    
         Re-arranging and using that ${R}_{\text{train}}(\barf) \leq \misspecification^2$, thus we have
        \begin{equation}
        \begin{aligned}
            R_{train}\left(\hatfn\right)
            & \leq \misspecification^2 + \frac{128\maxval{\outcome}^2\log (|\mathcal{F}| / \delta)}{3n} - \sqrt{3}\widehat{\amplification} 
        \end{aligned}
        \end{equation}
    
        Finally,  let $\maxval{\covshift} := \sup_{\xi \in \Xi} \left| \frac{d_{\text{test}}(\xi)}{d_{\text{train}}(\xi)} \right|$, the upper bound of generalization error can be expressed via the density ratio:

        \begin{equation}
        \begin{aligned}
            R_{test}\left(\hatfn\right)
            & \leq \maxval{\covshift} 
            \cdot \left(\misspecification^2 + \frac{128\maxval{\outcome}^2\log (|\mathcal{F}| / \delta)}{3n} - \sqrt{3}\widehat{\amplification}  )\right)
        \end{aligned}
        \end{equation}

        where $ \widehat{\amplification}(\hatfn) = \mathbb{E}_{\text{train}}\left[ (\hatfn- \barf(\xi)) ( \barf(\xi) -\starf(\xi)) \right]$

    \subsubsection{Solution 2: Quadratic-based Upper Bound Derivation }
     Or solving the quadratic inequality $A  \leq\sqrt{(16 A  - 32 \widehat{\amplification}) 2C} + \frac{16C}{3}$, thus
     \begin{equation}
         \begin{aligned}
            & (A - \frac{16C}{3})^2 \leq 32 AC  - 64 \widehat{\amplification} C\\
            & \Rightarrow A^2 - \frac{32AC}{3} + (\frac{16C}{3})^2 \leq 32 AC  - 64 \widehat{\amplification} C\\
            & \Rightarrow A^2 - \frac{128AC}{3} +(\frac{16C}{3})^2 + 64 \widehat{\amplification} C \leq 0
         \end{aligned}
     \end{equation}
    thus,
    \begin{equation}
        A \leq \frac{64C}{3} + \sqrt{\frac{1280C^2}{3} - 64\widehat{\amplification} C}
    \end{equation}

    here exists a solution if and only if$ \frac{1280C^2}{3} - 64\widehat{\amplification} C \geq 0 $, which equals $\widehat{\amplification} \leq \frac{20C}{3}$

    thus:
        \begin{equation}
        \begin{aligned}
            R\left(\hatfn\right)-R(\bar{f}) 
            & \leq   \frac{64\maxval{\outcome}^2\log (|\mathcal{F}| / \delta)}{3n} + \sqrt{\frac{1280 \maxval{\outcome}^4 [\log (|\mathcal{F}| / \delta)]^2}{3n^2} - \frac{64\widehat{\amplification} \maxval{\outcome}^2\log (|\mathcal{F}| / \delta)}{n}}
        \end{aligned}
        \end{equation}
    
         Re-arranging and using that ${R}_{\text{train}}(\barf) \leq \misspecification^2$, thus we have
        \begin{equation}
        \begin{aligned}
            R_{train}\left(\hatfn\right)
            & \leq \misspecification^2 + \frac{64\maxval{\outcome}^2\log (|\mathcal{F}| / \delta)}{3n} + \sqrt{\frac{1280 \maxval{\outcome}^4 [\log (|\mathcal{F}| / \delta)]^2}{3n^2} - \frac{64\widehat{\amplification} \maxval{\outcome}^2\log (|\mathcal{F}| / \delta)}{n}}
        \end{aligned}
        \end{equation}
    
        Finally, let $\maxval{\covshift} := \sup_{\xi \in \Xi} \left| \frac{d_{\text{test}}(\xi)}{d_{\text{train}}(\xi)} \right|$, the upper bound of generalization error can be expressed via the density ratio:

        \begin{equation}
        \begin{aligned}
            R_{test}\left(\hatfn\right)
            & \leq \maxval{\covshift} 
            \cdot \left(\misspecification^2 + \frac{64\maxval{\outcome}^2\log (|\mathcal{F}| / \delta)}{3n} + \sqrt{\frac{1280 \maxval{\outcome}^4 [\log (|\mathcal{F}| / \delta)]^2}{3n^2} - \frac{64\widehat{\amplification} \maxval{\outcome}^2\log (|\mathcal{F}| / \delta)}{n}}  )\right)
        \end{aligned}
        \end{equation}

        where $ \widehat{\amplification}(\hatfn) = \mathbb{E}_{\text{train}}\left[ (\hatfn- \barf(\xi)) ( \barf(\xi) -\starf(\xi)) \right]$

    \subsection { Loosen the $\widehat{\amplification}$}
    We reproduce this derivation for completeness, following the approach in \citep{amortila2024mitigating}, although it is not directly used in our main results.
    
        In the inequality \ref{equ:var}, we keep the interaction term $\widehat{\amplification}$. and if we loose the $\widehat{\amplification}$ by using the AM-GM inequality $\left((a+b)^2 \leq 2 a^2+2 b^2\right)$, the above equation becomes :
    
        \begin{equation}
        \begin{aligned} \label{equ:loosA}
            &\operatorname{Var} \left[(f(\xi)-y)^2-(\barf(\xi)-y)^2\right]  \\
            & = \mathbb{E}\left[\left((f(\xi)-y)^2-(\barf(\xi)-y)^2\right)^2\right] - \left(\mathbb{E}\left[\left((f(\xi)-y)^2-(\barf(\xi)-y)^2\right)\right] \right)^2\\
            &\leq \mathbb{E}\left[\left((f(\xi)-y)^2-(\barf(\xi)-y)^2\right)^2\right] \\
            & = \mathbb{E}\left[\left( f^2(\xi) -2 f(\xi)y -\barf^2(\xi) + 2y \barf(\xi)   \right )^2\right] \\
            & =\mathbb{E}\left[(f(\xi)-\barf(\xi))^2(f(\xi)+\barf(\xi)-2 y)^2\right]  \\ 
            & \leq 16 \maxval{\outcome}^2 \mathbb{E}\left[(f(\xi)-\barf(\xi))^2\right]   \quad \text{(since } |f(\xi)|, |\bar{f}(\xi)|, |y| \leq \maxval{\outcome} \text{)} \\  
            & = 16 \maxval{\outcome}^2 \mathbb{E}\left[(f(\xi)- \starf(\xi)+\starf(\xi)-\barf(\xi) )^2 \right] \\
            & \leq 32 \maxval{\outcome}^2 \mathbb{E}\left[( (f(\xi)- \starf(\xi))^2 + (\barf(\xi) -\starf(\xi))^2 )\right] \quad  \text{(since }  (a +b)^2 \leq 2 a^2 + 2b^2 ) \\
            & = 32 \maxval{\outcome}^2\mathbb{E}\left[\left(f(\xi)-\starf(\xi)\right)^2 - \left(\barf(\xi)-\starf(\xi)\right)^2 + 2 \left(\barf(\xi)-\starf(\xi)\right)^2 \right] \\
            & \leq 32 \maxval{\outcome}^2\mathbb{E}\left[\left(f(\xi)-\starf(\xi)\right)^2-\left(\barf(\xi)-\starf(\xi)\right)^2\right] 
            +64 \maxval{\outcome}^2 \misspecification^2 \\
            & = 32 \maxval{\outcome}^2 ( R(f) -R(\barf) )  +64 \maxval{\outcome}^2 \misspecification^2
        \end{aligned}
        \end{equation}
        
        Using the same logic in the above derivation, based on the \Cref{equ:loosA}, we have:
    
        \begin{equation}
        \begin{aligned}
            A & \leq\sqrt{(32 A+64 B )2C} + \frac{16C}{3} \\
            & \leq \sqrt{64 A C} + \sqrt{128 B C}   + \frac{16C}{3}  \quad (\text{since} \sqrt{a+b} < \sqrt{a} + \sqrt{b})\\ 
            & = \sqrt{ A \cdot 64C } + \sqrt{ 2B \cdot 64C}   + \frac{16C}{3}\\
            &  \leq \frac{1}{2} A + 32C +B +32C + \frac{16C}{3}  \quad (\text{since} \sqrt{ab} \leq a/2 +b/2 )\\
            & = \frac{1}{2} A + B + \frac{208C}{3}
        \end{aligned}
        \end{equation}
    
            so that
        \begin{equation}
            A \leq 2B + \frac{416C}{3}
        \end{equation}
        So
        \begin{equation}
             R\left(\hatfn\right)-R(\bar{f})  \leq 2 \misspecification^2  + \frac{416 \maxval{\outcome}^2 \log (|\mathcal{F}| / \delta)}{3n}
        \end{equation}
        Re-arranging and using that ${R}_{\text{train}}(\barf) \leq \misspecification^2$, thus we have
         \begin{equation}
             R_{\text{train}}\left(\hatfn\right) \leq 3 \misspecification^2  
             + \frac{416 \maxval{\outcome}^2\log (|\mathcal{F}| / \delta)}{3n}
        \end{equation}
        
        Let $\maxval{\covshift} := \sup_{\xi \in \Xi} \left| \frac{d_{\text{test}}(\xi)}{d_{\text{train}}(\xi)} \right|$, the upper bound of generalization error can be expressed via the density ratio:
        
        \begin{equation}
            R_{\text{test}}\left( \hatfn \right)
            = \mathbb{E}_{\text{train}} \left[
            \frac{d_{\text{test}}(\xi)}{d_{\text{train}}(\xi)} \left( \hatfn(\xi) - f^*(\xi) \right)^2
            \right]
            \leq \maxval{\covshift} 
            \cdot \left( 3\misspecification^2 
            + \frac{416 \maxval{\outcome}^2 \log(|\mathcal{F}|/\delta)}{3n} \right)
        \end{equation}

\section{CONSTRUCTING THE APPROXIMATE/PROXY DE-AMPLIFYING REGION}\label{appendix:proof_ridea_all}
    \subsection{PROOF of \Cref{theor:worstcase}} \label{appendix:proof_worstcase}
    \textbf{Theorem 2} [Approximate de-amplifying region]
            Let $\barf$ be the predictor that best approximates the true data-generating function $\starf$ in expectation with respect to the test distribution. Then,
            \begin{equation}\label{equ:worstcase}
                \approxDeAregion(\tau_1) \subseteq \DeAregion(\tau_0) \nonumber
            \end{equation}
            where the \textnormal{approximate de-amplifying region}
            $\approxDeAregion(\tau_1) \coloneqq \Big\{\xi:|\hatfn(\xi)-\barf(\xi)|  
            \geq \tau_1 \Big\}$, $\tau_1 = \tau_0/\maxval{\misspecification} + c\maxval{\misspecification}$, $c \geq 2$, and $\tau_0 \geq 0$. 
            
\paragraph{Proof Sketch:}
    To compute a subset of $\Big\{\xi:(\hatfn(\xi)-\barf(\xi)) (\barf(\xi)- \starf(\xi)) \geq \tau_0\Big\}$ 
    and remove the dependence on $\starf$, we instead construct a subset of the de-amplifying region, the \textit{approximate de-amplifying region}:
    \begin{equation} \label{eq:appendix_de_region}
        \begin{aligned}
            \Big\{\xi:|\hatfn(\xi)-\barf(\xi)| \geq c \maxval{\misspecification} + \tau_0/\maxval{\misspecification}, \; c \geq 2, \tau_0 \geq 0  \Big\} 
         \subseteq 
         \Big\{\xi:(\hatfn(\xi)-\barf(\xi)) (\barf(\xi)- \starf(\xi)) \geq \tau_0\Big\}.
        \end{aligned}
    \end{equation}
    
    Below, we detail the derivation of each of the following subset relations:
    \begin{equation}\label{eq:approx-DeA} 
            \begin{aligned}
            \approxDeAregion(\tau_1) &=
             \Big\{\xi:|\hatfn(\xi)-\barf(\xi)| \geq c \maxval{\misspecification} + \tau_0/\maxval{\misspecification}, \; c \geq 2 , \tau_0 \geq 0\Big\} \\
              &\subseteq^{(i)}
             \Big\{\xi:|\hatfn(\xi)-\starf(\xi)| \geq \tau_0/\maxval{\misspecification} + \maxval{\misspecification}, \tau_0 \geq 0\Big\} 
             \\
              & \subseteq^{(ii)}
             \Big\{\xi:|\hatfn(\xi)-\starf(\xi)| - |\barf(\xi)- \starf(\xi)| \geq \tau_0/\maxval{\misspecification}, \tau_0 \geq 0\Big\} \\
             & \subseteq^{(iii)}\Big\{\xi:(\hatfn(\xi)-\barf(\xi)) (\barf(\xi)- \starf(\xi)) \geq \tau_0\Big\} \\
             &= \DeAregion(\tau_0) \nonumber \\
            \end{aligned}
        \end{equation}

    \paragraph{Proof of subset relation (iii)}
         We would design an acquisition function that selects only designs in the de-amplifying region $\DeAregion(\tau_0)$, where $\DeAregion(\tau_0) \coloneqq \Big\{\xi:(\hatfn(\xi)- \barf(\xi)) ( \barf(\xi) -\starf(\xi)) \geq \tau_0\Big\}$. And the left side of the inequality can be expressed as below:
        \begin{equation}
            \begin{aligned}
            &(\hatfn(\xi)- \barf(\xi)) ( \barf(\xi) -\starf(\xi))  \\
            &=   (\hatfn(\xi)- \starf(\xi) + \starf(\xi) -\barf(\xi)) ( \barf(\xi) -\starf(\xi))  \\
            & =  (\hatfn(\xi)- \starf(\xi)) ( \barf(\xi) -\starf(\xi)) -  ( \barf(\xi) -\starf(\xi))^2  
            \end{aligned}
        \end{equation}
        
        Thus, we refer to a design $\xi$ as de-amplifying whenever $(\hatfn(\xi)- \barf(\xi)) ( \barf(\xi) -\starf(\xi)) = (\hatfn(\xi)- \starf(\xi)) ( \barf(\xi) -\starf(\xi)) -  ( \barf(\xi) -\starf(\xi))^2\geq \tau_0$ for a given threshold $\tau_0$.

        Make an assumption that $\tau_0 \geq 0$, and based on $ |\barf(\xi) -\starf(\xi)|  < \maxval{\misspecification}$, thus we have
        $$|\hatfn(\xi)- \starf(\xi)| - |\barf(\xi) -\starf(\xi)| \geq \tau_0/|\barf(\xi) -\starf(\xi)| \geq \tau_0/\maxval{\misspecification}, $$ the subset (iii) follows.

    \paragraph{Proof of subset relation (ii)}
        From $\Big\{\xi:|\hatfn(\xi)-\starf(\xi)| - |\barf(\xi)- \starf(\xi)| \geq \tau_0/\maxval{\misspecification}\Big\}$  
            we obtain
            \[
              \Big\{\xi:|\hatfn(\xi)-\starf(\xi)| \geq \tau_0/\maxval{\misspecification} +|\barf(\xi)- \starf(\xi)|\Big\}.
            \]
             
            Since $|\barf(\xi)- \starf(\xi)| \leq \maxval{\misspecification}$, it follows that
            \[
              \Big\{\xi:|\hatfn(\xi)-\starf(\xi)| \geq \tau_0/\maxval{\misspecification} + \maxval{\misspecification}\Big\} 
              \subseteq  
              \Big\{\xi:|\hatfn(\xi)-\starf(\xi)| - |\barf(\xi)- \starf(\xi)| \geq \tau_0/\maxval{\misspecification}\Big\}.
            \]
    
    \paragraph{Proof of subset relation (i)}
        Making a assumption that $\{\xi:|\hatfn(\xi)-\barf(\xi)| \geq c \maxval{\misspecification} + \tau_0/\maxval{\misspecification}\}$ holds, then
        \[
        |\hatfn(\xi)-\starf(\xi)| = |\hatfn(\xi)-\barf(\xi)+ \barf(\xi)-\starf(\xi)| 
        \geq  |\hatfn(\xi)-\barf(\xi)|- |\starf(\xi)- \barf(\xi)| 
        \geq (c-1)\maxval{\misspecification} + \tau_0/\maxval{\misspecification}.
        \]
    
        Combining this with  
        $\Big\{\xi:|\hatfn(\xi)-\starf(\xi)| \geq \tau_0/\maxval{\misspecification} + \maxval{\misspecification}\Big\}$,  
        we see that for $c \geq 2$, subset (i) follows.
       
    \subsection{PROOF of \Cref{lem:proxy_fidelity}}
    \textbf{Lemma 1} [Proxy region approximates $\approxDeAregion(\tau)$]
        Given a \textnormal{proxy function} $g ~ : ~ \Xi \mapsto \mathbb{R}$ such that $\sup_{\xi\in\Xi}|g(\xi)-\bar f(\xi)| \le \tau_2$ for some $\tau_2 \geq 0$,
        \[
          \proxyDeAregion(\tau) \coloneqq \big\{\xi \in \Xi:\,|\hatfn(\xi)-g(\xi)| \ge \tau \big\}\;\subseteq\;\approxDeAregion(\tau_1).
        \]
        where $\tau = \tau_1+\tau_2$.
        
      \textbf{Proof.} 
      By the triangle inequality,
        $$|\hatfn(\xi)-\bar f(\xi)| \ge |\hatfn(\xi)-g(\xi)| - |g(\xi)-\bar f(\xi)| \ge |\hatfn(\xi)-g(\xi)|-\tau_2$$
        Thus when $$|\hatfn(\xi)-g(\xi)|\ge \tau_1+\tau_2,$$ we have
        $$
        |\hatfn(\xi)-\bar f(\xi)| \ge |\hatfn(\xi)-g(\xi)|-\tau_2 \ge \tau_1
        \Rightarrow |\hatfn(\xi)-\bar f(\xi)|\ge \tau_1
        $$
        So   
        \[
          \big\{\xi:\,|\hatfn(\xi)-g(\xi)| \ge \tau_1+\tau_2\big\}\;\subseteq\;\approxDeAregion(\tau_1).
        \]

\section{ADDITIONAL DETAILS OF THE NUMERICAL EXPERIMENTS} \label{appendix:appendix_ex_detail}
    Our experiments were implemented using PyTorch \citep{paszke2019pytorch} and Pyro \citep{bingham2019pyro}.
    All experiments were conducted on a shared computing cluster using NVIDIA A100 GPUs. Each job was allocated one A100 GPU, 8 CPU cores, and 32GB of RAM.

    \paragraph{Measuring the degree of covariate shift}
        To measure the distance between the distributions or datasets, we use the \textit{ Maximum mean discrepancy} (MMD). 
        A number of advantages of this distance are put forward in the literature: 1) it is more robust to outliers than other discrepancy measurements (like KL divergence) \citep{gretton2012kernel}; 2) it can be approximated on the basis of differently-sized samples from the distributions being compared \citep{gretton2012kernel}; 3) the measurement is robust to repeated samples (unlike the Wasserstein distance) \citep{viehmann2021partial}; 4) the measurement can be computed efficiently using samples \citep{bharti2023optimally,huang2023learning}. 
        
        We compute the squared Maximum Mean Discrepancy (MMD) \citep{gretton2012kernel} between two empirical distributions 
        $\mathcal{P} = \{p_i\}_{i=1}^m$ and $\mathcal{Q} = \{q_j\}_{j=1}^n$ using the Gaussian (RBF) kernel:
        \[
        \mathrm{MMD}^2(\mathcal{P}, \mathcal{Q})
        = \frac{1}{m^2} \sum_{i=1}^m \sum_{i'=1}^m k(p_i, p_{i'}) 
        + \frac{1}{n^2} \sum_{j=1}^n \sum_{j'=1}^n k(q_j, q_{j'}) 
        - \frac{2}{mn} \sum_{i=1}^m \sum_{j=1}^n k(p_i, q_j),
        \]
        where $k(p,q) = \exp\!\left(-\tfrac{\|p-q\|^2}{2\sigma^2}\right)$ is the RBF kernel with fixed bandwidth $\sigma=1$.

    \paragraph{Measuring the generalization performance}
        To evaluate the generalization performance, we compute the \textit{Mean Squared Error (MSE)} between the model predictions $\hatfn$ and the corresponding true observations from the data-generating process (DGP) $y$, given the test samples.
        $$\mathrm{MSE}=\frac{1}{D} \sum_{d=1}^D \frac{1}{N} \sum_{i=1}^N(\hatfn_{d,i}-y_{d,i})^2$$
        where D is the number of test samples, and N is the sampling number. 
    
    \subsection {Polynomial Regression Experiments}\label{appendix:appendix_ex_detail_toy}

    Both well-specified and misspecified models were run across 20 runs. For each run, the design is adaptively selected, and the total number of designs is $T=10$.
    \paragraph{Well-specified case}
        \begin{itemize}
            \item \textbf{DGP}: The DGP is a degree-two polynomial regression model, $y=1+2 x-0.5x^2+\epsilon$ where $\epsilon \sim \mathcal{N}(0,0.1)$
            \item \textbf{ well-specified model}: $f(x)=\beta_0+\beta_1 x + \beta_2 x^2$ So the feature functions are $\phi(x) = [1, x, x^2]^{\top}$
            \item \textbf{Test distribution} (arbitrary $\testx$): $\testx \sim \mathcal{U}(-4,4)$
        \end{itemize}

        \paragraph{Misspecified case}
        \begin{itemize}
            \item \textbf{DGP}: The DGP is a degree-two polynomial regression model, $y=1+2 x-0.5x^2+\epsilon$ where $\epsilon \sim \mathcal{N}(0,0.1)$
            \item  \textbf{misspecified model}: $f(x)=\beta_0+\beta_1 x$ So the feature functions are $\phi(x) = [1, x]^{\top}$
            \item  \textbf{Test distribution} (arbitrary $\testx$): $\testx \sim \mathcal{U}(-4,4)$
        \end{itemize}

    \subsection{Source Localization Experiments} \label{appendix:appendix_ex_detail_location}

        We take $T = 30$ iteration steps for selecting training samples $\trainx$ and 200 test samples for the observed $\starf(\testx)$ and predicted output $f(\testx)$. We set $K = 2$ sources and limit the range of design to $\trainx \in [-4, 4]$. We use 200 samples drawn from the uniform distribution over $\testx \in [-4, 4]$ as the test distribution to estimate the MMD.
    
        In this experiment, we have $K$ sources with unknown location parameter is $\theta = \{\theta_k\}_{k=1}^{K}$. We assume the number of sources $K$ is known.

        \paragraph{Acoustic energy attenuation model}
        The total intensity, a superposition of the individual ones, at location $\xi$ from $K$ sources is considered in the following equation \citep{foster2021deep, ivanova2024step}:
            \begin{equation}
                \mu(\theta, \xi)=b+\sum_{k=1}^K \frac{\alpha_k}{m+\left\|\theta_k-\xi\right\|^2}
            \end{equation}
        The prior distribution of each location parameter $\theta_k$ is a normal distribution, and the observation noise is Gaussian noise. Therefore, the prior and likelihood are in the following:
        $$
            \theta_k \stackrel{\text { i.i.d. }}{\sim} N\left(0_d, I_d\right), \log y \mid \theta, \xi \sim N(\log \mu(\theta, \xi), \sigma)
        $$

        \paragraph{Well-specified case}
        The hyperparameters of the true DGP used in our well-specification experiments can be found in the table below:
        $$
            \begin{array}{lr}
                \text { Parameter } & \text { Value } \\
                \hline \text { Number of sources, } K & 2 \\
                \text { Base signal, } b & 10^{-1} \\
                \text { Max signal, } m & 10^{-4} \\
                \alpha_1, \alpha_2 & 1 \\
                \text { Signal noise, } \sigma & 0.1
            \end{array}
        $$

        The model hyperparameters used in our experiments are the same in the above table.
        
        \paragraph{Misspecfied case}
        The hyperparameters of the true DGP used in our mis-specification experiments can be found in the table below:
        $$
            \begin{array}{lr}
                \text { Parameter } & \text { Value } \\
                \hline \text { Number of sources, } K & 2 \\
                \text { Base signal, } b & 4*10^{-1} \\
                \text { Max signal, } m &  4 *10^{-4} \\
                \alpha_1, \alpha_2 & 0.4 \\
                \text { Signal noise, } \sigma & 0.1
            \end{array}
        $$
        The model hyperparameters used in our experiments are the same as the table in the well-specified case.

        \subsection{Pharmacokinetic Model} \label{appendix:appendix_ex_detail_pk}
            We take $T = 10$ iteration steps for selecting training samples $\trainx$ and 200 test samples for the observed $\starf(\testx)$ and predicted output $f(\testx)$. We limit the range of design to $\trainx \in [0, 24]$. We use 200 samples drawn from the uniform distribution over $\testx \in [0, 24]$ as the test distribution to estimate the MMD.
            In the experiment, we set the true theta as $\theta_{\text{real}} = [1.5, 0.15, 15.0]$

           \paragraph{Well-specified case.}
               The drug concentration $z$, measured $\xi$ hours after administration, and the corresponding noisy observation $y$, are modeled as
                \begin{equation}
                    z(\xi; \theta) = \frac{D_V}{V} \cdot \frac{k_\alpha}{k_\alpha - k_e}
                    \left[ e^{-k_e \xi} - e^{-k_\alpha \xi} \right], 
                    \qquad
                    y(\xi; \theta) = z(\xi; \theta)(1 + \epsilon) + \eta,
                    \label{eq:drug_model}
                \end{equation}
                where $\theta = (k_\alpha, k_e, V)$, $D_V = 400$ is a constant, 
                $\epsilon \sim \mathcal{N}(0, 0.01)$ is multiplicative noise (to capture heteroscedasticity), 
                and $\eta \sim \mathcal{N}(0, 0.1)$ is additive observation noise.  
    
                The prior distribution for the parameters $\theta$ is specified as
                \begin{equation}
                    \log \theta \sim \mathcal{N}\!\left(
                    \begin{bmatrix}
                        \log 1 \\
                        \log 0.1 \\
                        \log 20
                    \end{bmatrix},
                    \begin{bmatrix}
                        0.05 & 0 & 0 \\
                        0 & 0.05 & 0 \\
                        0 & 0 & 0.05
                    \end{bmatrix}
                    \right).
                    \label{eq:prior}
                \end{equation}
               Since both noise sources are Gaussian, the observation likelihood and DGP are the same and are also Gaussian:
                \begin{equation} \label{equ:pk_well}
                    y(\xi;\theta)\sim\mathcal{N}\!\big(z(\xi;\theta),\;0.01\,z(\xi;\theta)^2+0.1\big),
                \end{equation}

            \paragraph{Misspecified case.}
                To introduce model misspecification, DGP comes from \Cref{equ:pk_well}.
                And a dual–absorption model with two parallel absorption rates is generated to make a prediction.  
                Let $k_{a1}=k_\alpha$ and $k_{a2}=\rho\,k_{a1}$ with $\rho\in(0,1)$, and let $f\in(0,1)$ denote the fraction of the fast pathway.  
                The mean concentration is
                \begin{equation}
                  z_{\text{pre}}(\xi;\theta,\rho,f)
                  = \tfrac{D_V}{V}\!\left[
                      f\cdot\tfrac{k_{a1}}{k_{a1}-k_e}\big(e^{-k_e\xi}+e^{-k_{a1}\xi}\big)
                      +(1-f)\cdot\tfrac{k_{a2}}{k_{a2}-k_e}\big(e^{-k_e\xi}+e^{-k_{a2}\xi}\big)
                    \right].
                \end{equation}
                We set $\rho=0.25$ and $f=0.6$.  
                The observation in the assumed model is,
                \[
                  y_{\text{pre}}(\xi;\theta)\sim\mathcal{N}\!\big(z_{\text{pre}}(\xi;\theta,\rho,f),\;0.02\,z_{\text{pre}}(\xi;\theta,\rho,f)^2+0.2\big).
                \]

\section{ADDITIONAL RESULTS OF THE NUMERICAL EXPERIMENTS}
\label{appendix:appendix_ex_results}
    \subsection{Polynomial Regression Experiments}\label{appendix:ex_toy}
        \subsubsection{Model Well-specification}  \label{appendix:ex_toy_linear}      
\begin{figure}[h]  
                \centering 
                \begin{subfigure}[b]{0.4\linewidth}
                    \centering
                    \includegraphics[width=\linewidth]{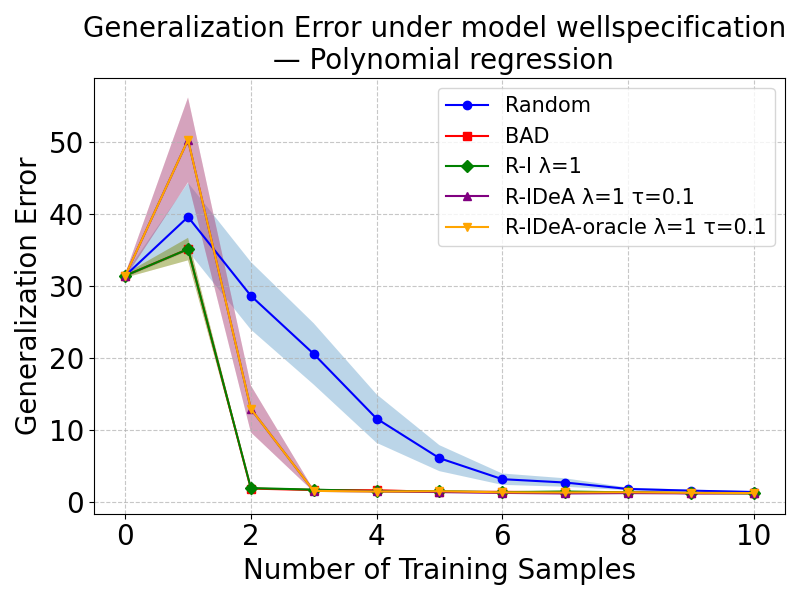}
                    \caption{}
                    \label{subfig:toy_base_results_gerror}
                \end{subfigure}
                \begin{subfigure}[b]{0.4\linewidth}
                    \centering
                    \includegraphics[width=\linewidth]{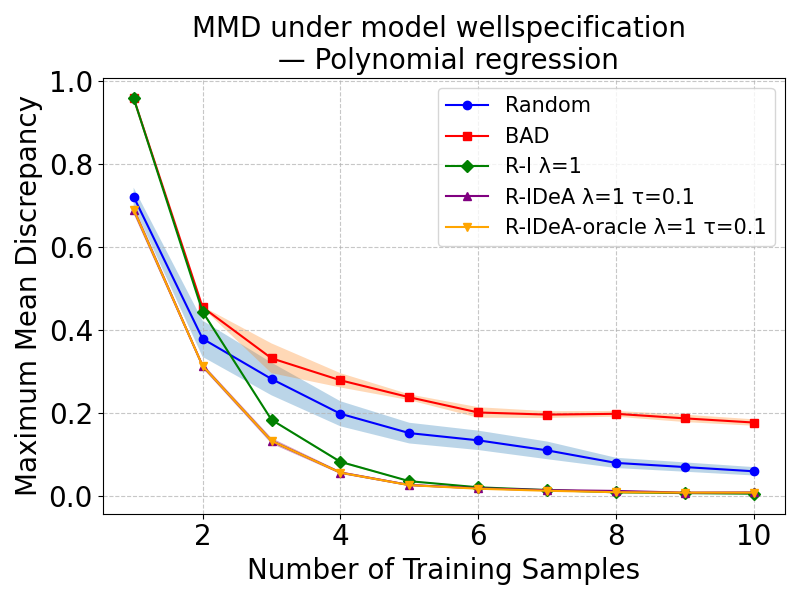}
                    \caption{}
                    \label{subfig:toy_base_results_mmd}
                \end{subfigure}
                \caption{\textbf{Polynomial regression experiments (well-specified case).} Comparison of different design strategies (Random, BAD, proposed \acqfri{}, proposed \acqfria{}, and \acqfria{}-oracle under well-specified models in polynomial regression.
                \textit{Left}: Generalization error across methods.
                \textit{Right}: MMD distance across methods; higher values indicate a greater degree of covariate shift.
                }
                \label{fig:toy_results_well}
            \end{figure}

            \Cref{fig:toy_results_well} shows that, in the well-specified case, all methods yield similar generalization error (\Cref{subfig:toy_base_results_gerror}), regardless of the degree of covariate shift (\Cref{subfig:toy_base_results_mmd}). This indicates that covariate shift does not significantly impact generalization performance when the model is well-specified.
            Moreover, the proposed \acqfri{} acquisition function, which combines informativeness and representativeness, converges more quickly than random selection. 
             \acqfria{} has comparable results to \acqfria{}-oracle, which uses the true value of $\barf$ instead of the proxy $g$, demonstrating the effectiveness of our method for selecting the proxy $g$.
            For clarity, we present results for $\lambda = 1$ and $\tau = 0.1$, as other settings exhibit similar behavior.
            
        \subsubsection{Model Misspecification} \label{subsec:ex_poly_hyper}

\begin{figure}[H]  
                    \centering 
                    \begin{subfigure}[b]{0.4\linewidth}
                        \centering
                        \includegraphics[width=\linewidth]{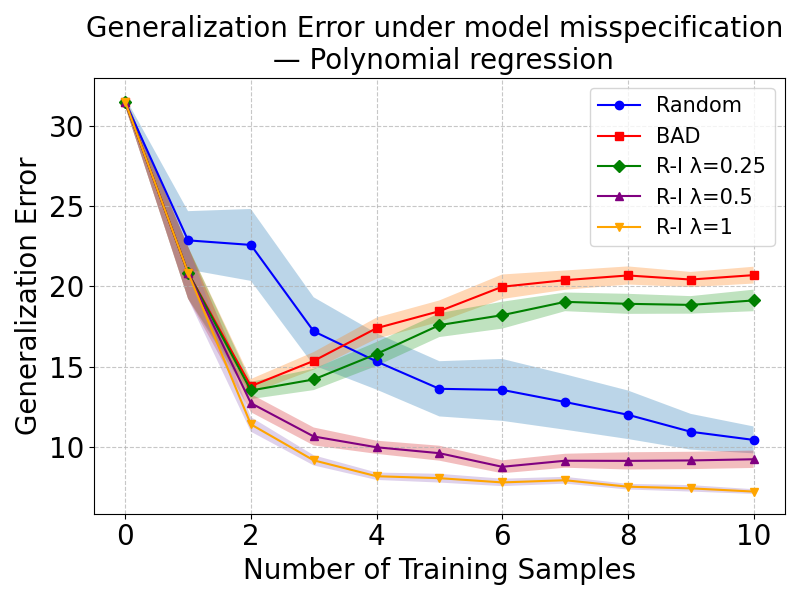}
                        \caption{}
                        \label{subfig:toy_lambda_misresults_gerror}
                    \end{subfigure}
                    \begin{subfigure}[b]{0.4\linewidth}
                        \centering
                        \includegraphics[width=\linewidth]{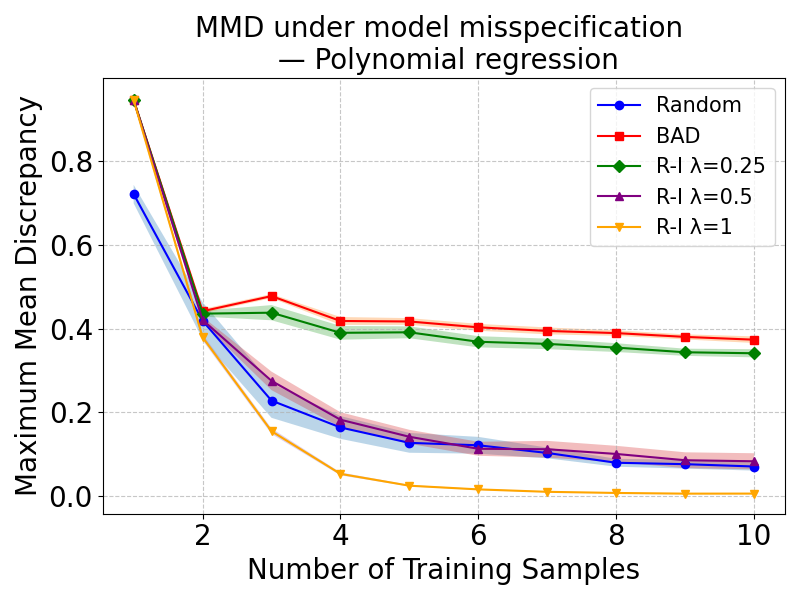}
                        \caption{}
                        \label{subfig:toy_lambda_misresult_mmd}
                    \end{subfigure}
                    \caption{\textbf{Polynomial regression experiments (effect of $\lambda$).} Comparison of baseline methods (Random, BAD) and our proposed \acqfri{} with varying $\lambda$ in the polynomial regression experiments.
                    \textit{Left}: Generalization error across methods.
                    \textit{Right}: MMD distance across methods; higher values indicate a greater degree of covariate shift.}
                    \label{fig:toy_lambda_misresults}
                \end{figure}

            \paragraph{\acqfri{} - varying $\lambda$}
            \Cref{subfig:toy_lambda_misresults_gerror,subfig:toy_lambda_misresult_mmd} show the performance of the proposed \acqfri{} acquisition function under various values of $\lambda$.
                For larger values of $\lambda$, we expect the representativeness term to dominate the acquisition function, resulting in a design distribution that resembles the test distribution.
                \Cref{subfig:toy_lambda_misresult_mmd} shows that when designs are more representative, generalization error is reduced (\Cref{subfig:toy_lambda_misresults_gerror}), consistent with the theoretical prediction introduced in \Cref{prop:gen_error_covshift}.
                These results demonstrate again that representative designs effectively reduce estimation bias and improve generalization performance.

\begin{figure}[H]  
                    \centering 
                    \begin{subfigure}[b]{0.4\linewidth}
                        \centering
                        \includegraphics[width=\linewidth]{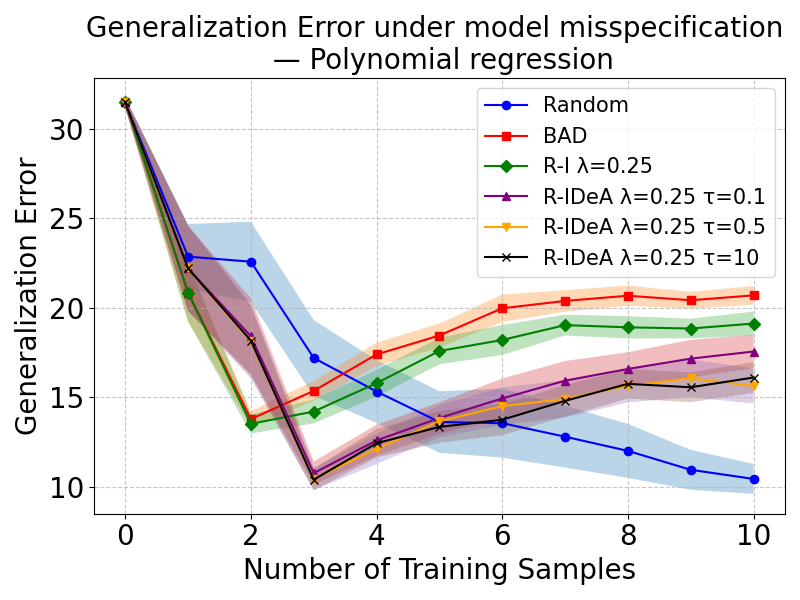}
                        \caption{}
                        \label{subfig:toy_lambda025_tau_misresults_gerror}
                    \end{subfigure}
                    \begin{subfigure}[b]{0.4\linewidth}
                        \centering
                        \includegraphics[width=\linewidth]{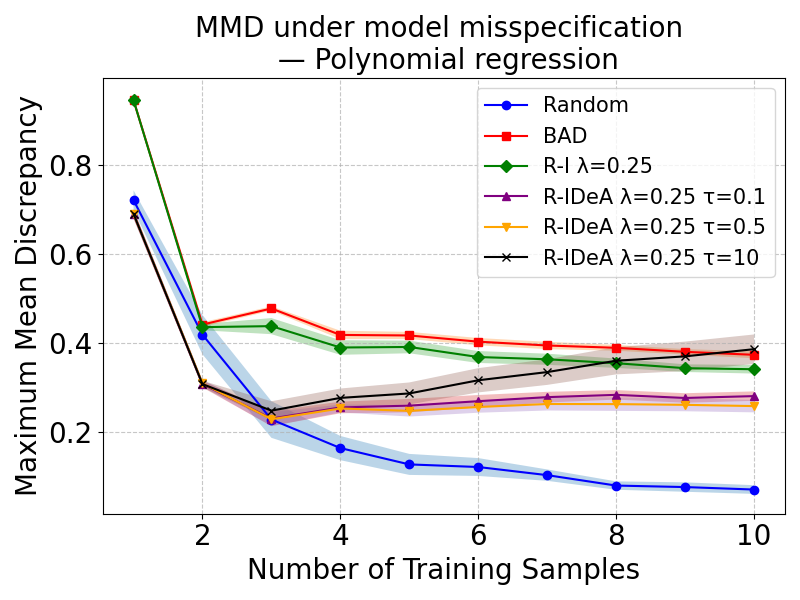}
                        \caption{}
                        \label{subfig:toy_lambda025_tau_misresult_mmd}
                    \end{subfigure}
                    
                    \begin{subfigure}[b]{0.4\linewidth}
                        \centering
                        \includegraphics[width=\linewidth]{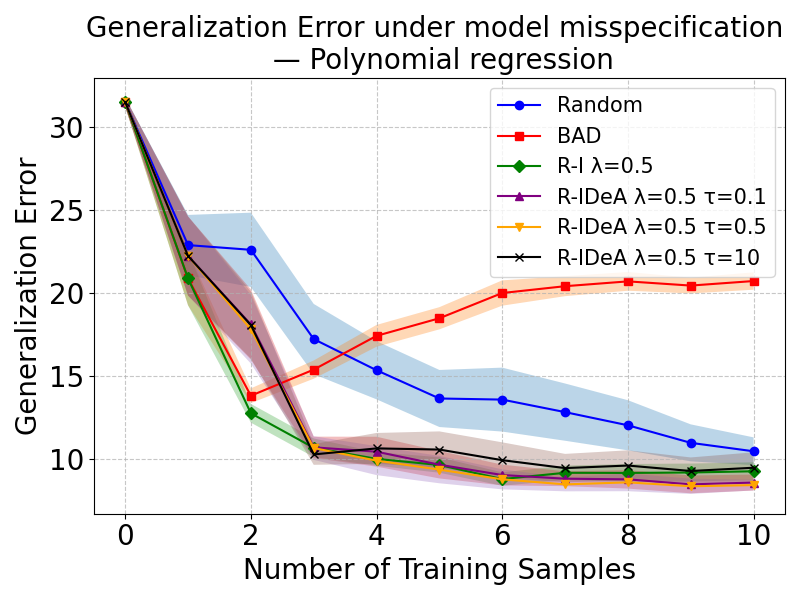}
                        \caption{}
                        \label{subfig:toy_lambda05_tau_misresults_gerror}
                    \end{subfigure}
                    \begin{subfigure}[b]{0.4\linewidth}
                        \centering
                        \includegraphics[width=\linewidth]{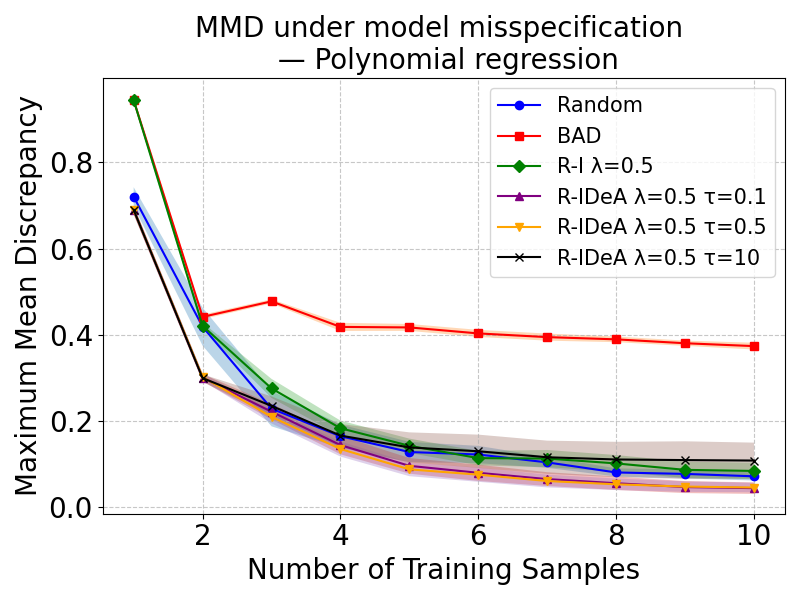}
                        \caption{}
                        \label{subfig:toy_lambda05_tau_misresult_mmd}
                    \end{subfigure}
                    
                    \begin{subfigure}[b]{0.4\linewidth}
                        \centering
                        \includegraphics[width=\linewidth]{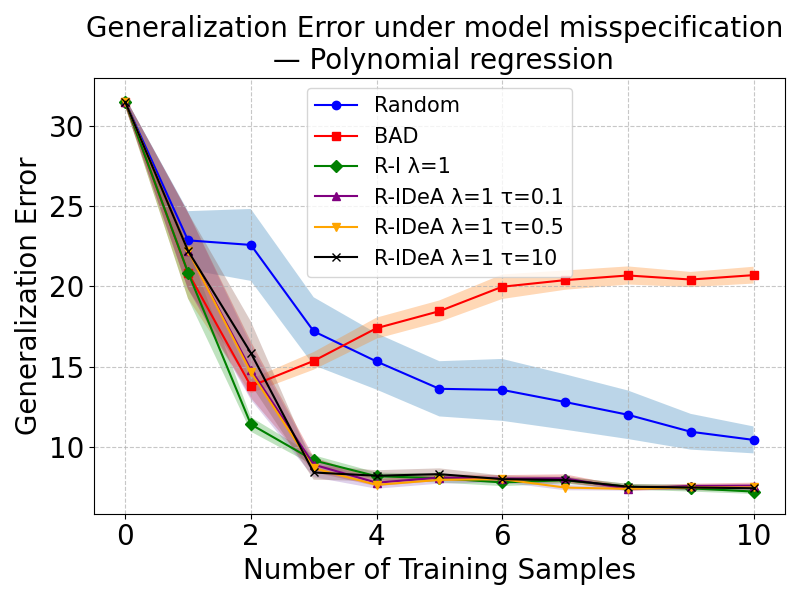}
                        \caption{}
                        \label{subfig:toy_lambda1_tau_misresults_gerror}
                    \end{subfigure}
                    \begin{subfigure}[b]{0.4\linewidth}
                        \centering
                        \includegraphics[width=\linewidth]{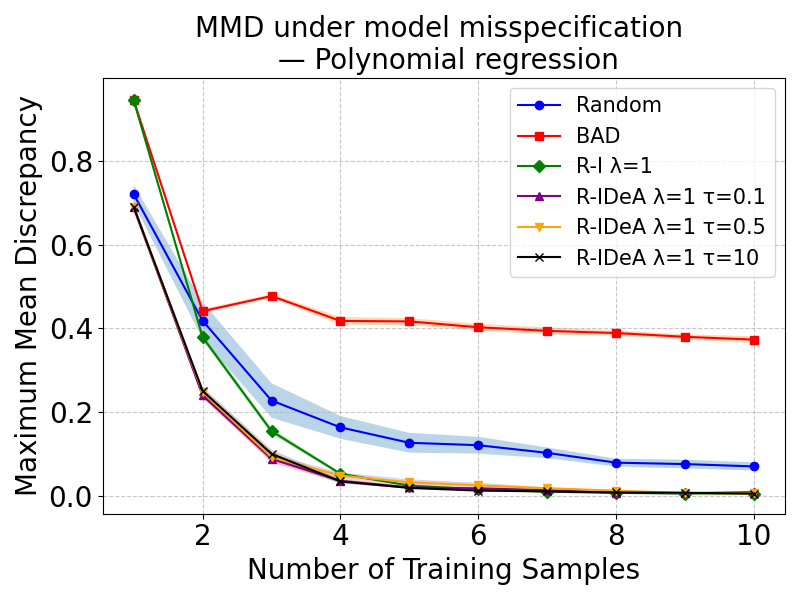}
                        \caption{}
                        \label{subfig:toy_lambda1_tau_misresult_mmd}
                    \end{subfigure}
                    \caption{\textbf{Polynomial regression experiments (effect of $\tau$).} Comparison of baseline methods (Random, BAD), our proposed \acqfri{} and our proposed \acqfria{} with varying $\tau$ in the polynomial regression experiments.
                    Rows correspond to variation in $\lambda$.
                    \textit{Left}: Generalization error across methods.
                    \textit{Right}: MMD distance across methods; higher values indicate a greater degree of covariate shift.}
                    \label{fig:toy_tau_misresults}
                \end{figure}

            \paragraph{\acqfria{} - varying $\tau$}
                \Cref{fig:toy_tau_misresults} shows the performance of our \acqfria{} under various learned adversarial proxies with different values of $\tau$.
                Over all values of $\tau$, \acqfria{} has a higher generalization performance than \acqfri{}, suggesting that de-amplifying designs indeed increase generalization performance.
                
                From \Cref{equ:Lg,equ:eig_de_amplify}, we see that larger values of $\tau$ result in design distributions with stricter de-amplifying constraints.
                In other words, when $\tau$ is large, we expect the de-amplifying term to dominate the acquisition function.
                For smaller values of $\tau$, the de-amplifying constraint is loosened, resulting in more candidate designs being effectively considered as part of the de-amplifying region.
                Therefore, $\tau$ should be chosen to balance de-amplification and other properties (informativeness and representativeness).
            
                \Cref{subfig:toy_lambda025_tau_misresults_gerror,subfig:toy_lambda05_tau_misresults_gerror,subfig:toy_lambda1_tau_misresults_gerror} show that when $\tau = 0.5$, \acqfria{} performs better than under other values of $\tau$ we tested.
                Also, \acqfria{} with $\tau = 0.5$ induces less covariate shift (lower MMD) than \acqfria{} with other values of $\tau$ (e.g., $\tau = 10$), showing that increasing the degree of de-amplification could reduce the degree of representativeness. These results show that the design properties of de-amplification and representativeness are interrelated; therefore, balancing de-amplification with other properties is important. 
                We leave the development of systematic methods for selecting the hyperparameters in \acqfri{} and \acqfria{} for future work.
                
        \subsubsection{High Degrees of Misspecification}
        \label{appendix:poly_high_mis}
        In the discussion before, we showed the results of experiments where the DGP was linear (\Cref{appendix:ex_toy_linear}) and quadratic (\Cref{subsec:ex_poly_hyper}), corresponding to no and mild misspecification, respectively.
        
\begin{figure}  
                    \centering 
                    \begin{subfigure}[b]{0.4\linewidth}
                        \centering
                        \includegraphics[width=\linewidth]{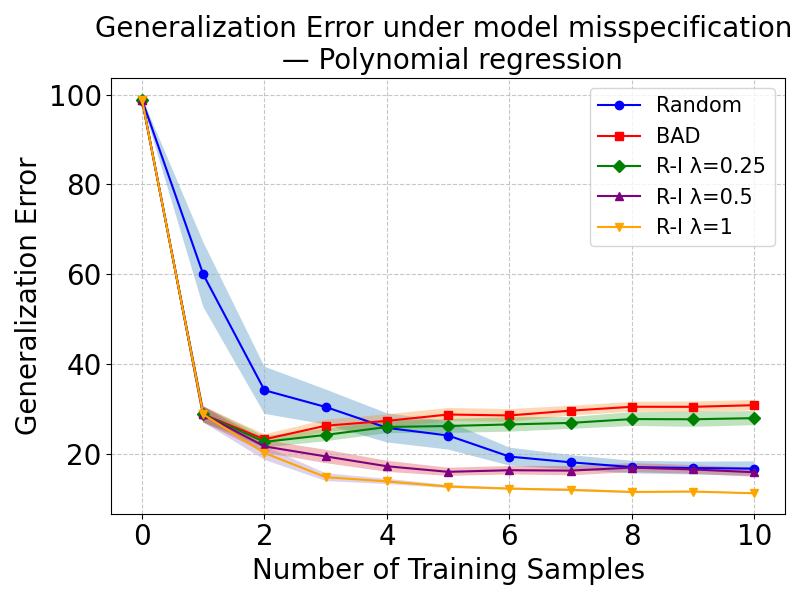}
                        \caption{}
                        \label{subfig:toy_lambda_mis3results_gerror}
                    \end{subfigure}
                    \begin{subfigure}[b]{0.4\linewidth}
                        \centering
                        \includegraphics[width=\linewidth]{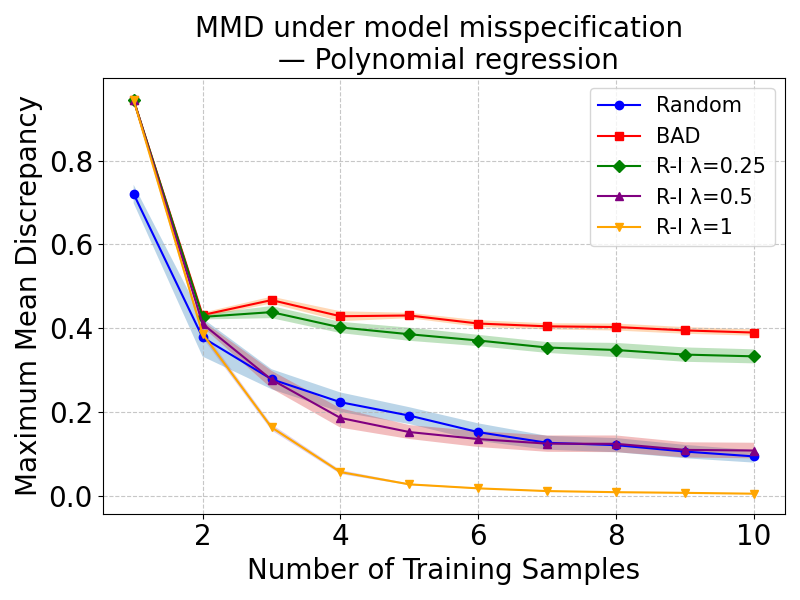}
                        \caption{}
                        \label{subfig:toy_lambda_mis3result_mmd}
                    \end{subfigure}
                    \caption{\textbf{Polynomial regression experiments (severely misspecified case).} Comparison of baseline methods (Random, BAD) and our proposed \acqfri{} with varying $\lambda$.
                    \textit{Left}: Generalization error across methods.
                    \textit{Right}: MMD distance across methods; higher values indicate a greater degree of covariate shift.}
                    \label{fig:toy_lambda_mis3results}
            \end{figure}

\begin{figure}  
                    \centering 
                    \begin{subfigure}[b]{0.4\linewidth}
                        \centering
                        \includegraphics[width=\linewidth]{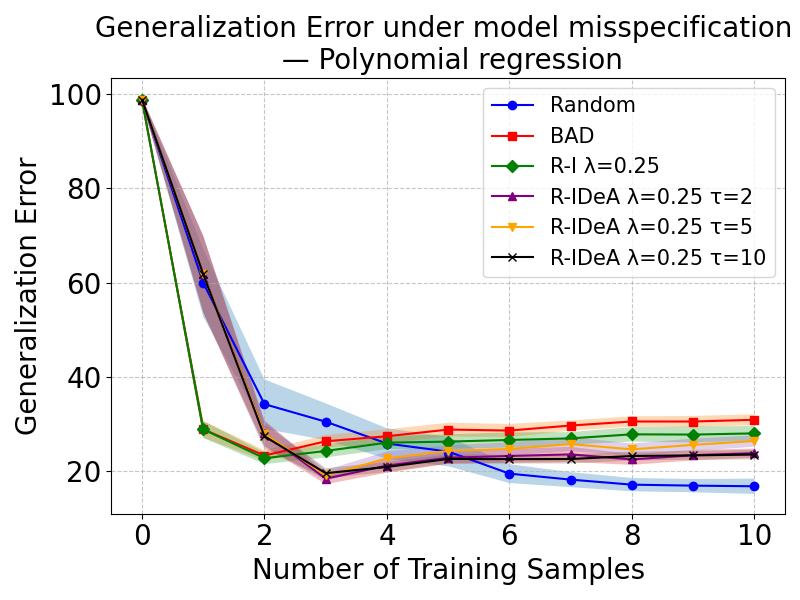}
                        \caption{}
                        \label{subfig:toy_lambda025_tau_mis3results_gerror}
                    \end{subfigure}
                    \begin{subfigure}[b]{0.4\linewidth}
                        \centering
                        \includegraphics[width=\linewidth]{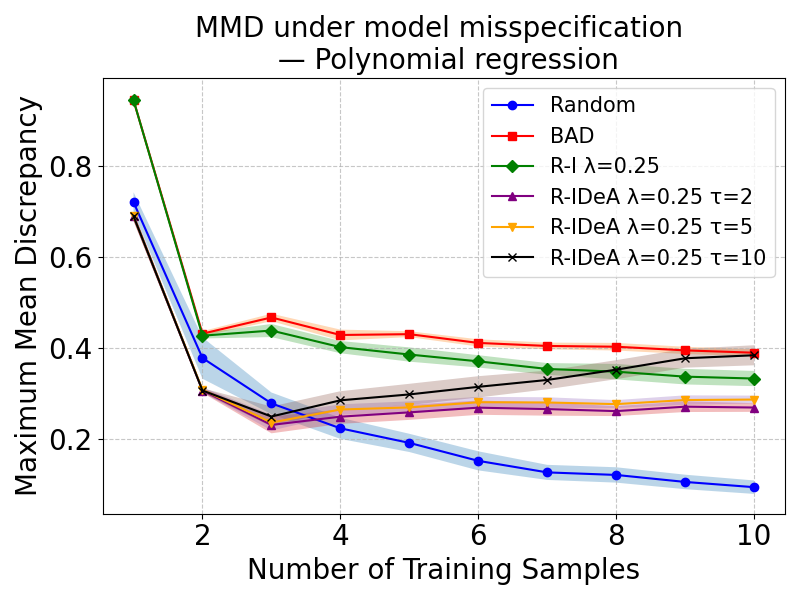}
                        \caption{}
                        \label{subfig:toy_lambda025_tau_mis3result_mmd}
                    \end{subfigure}
                    
                    \begin{subfigure}[b]{0.4\linewidth}
                        \centering
                        \includegraphics[width=\linewidth]{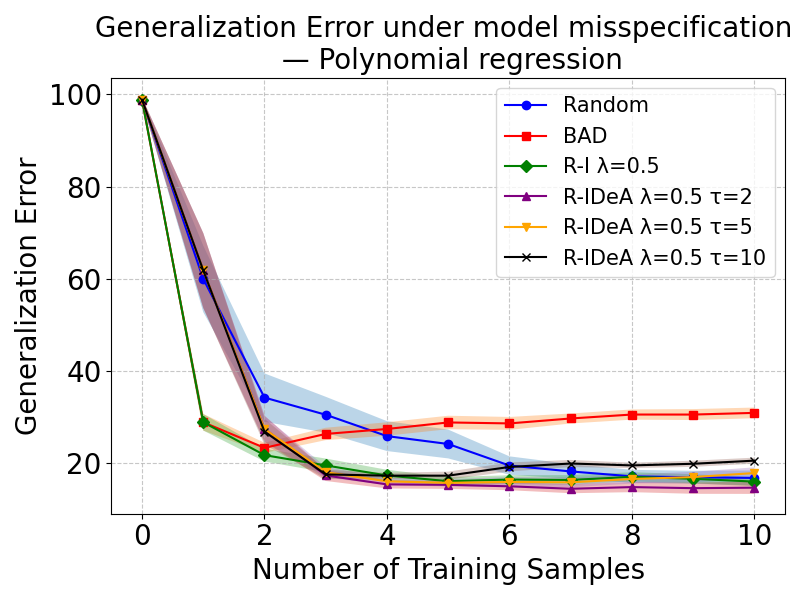}
                        \caption{}
                        \label{subfig:toy_lambda05_tau_mis3results_gerror}
                    \end{subfigure}
                    \begin{subfigure}[b]{0.4\linewidth}
                        \centering
                        \includegraphics[width=\linewidth]{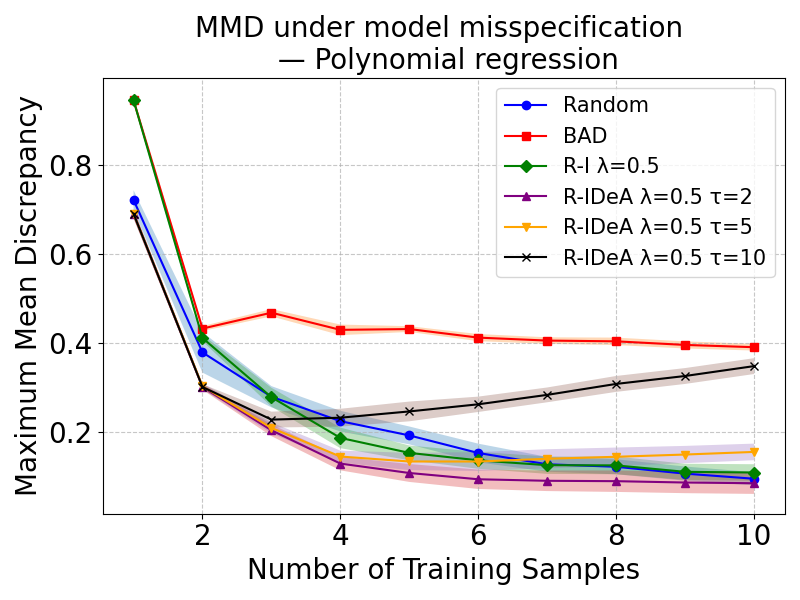}
                        \caption{}
                        \label{subfig:toy_lambda05_tau_mis3result_mmd}
                    \end{subfigure}
                    
                    \begin{subfigure}[b]{0.4\linewidth}
                        \centering
                        \includegraphics[width=\linewidth]{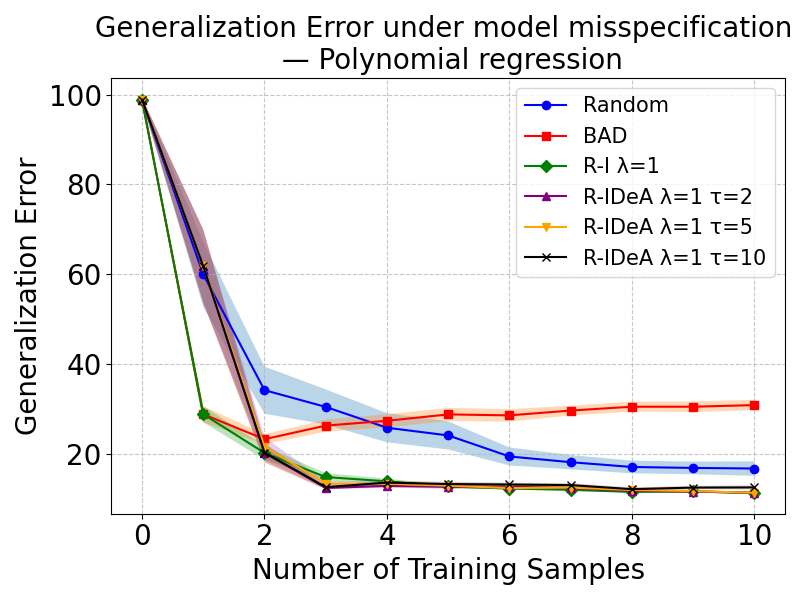}
                        \caption{}
                        \label{subfig:toy_lambda1_tau_mis3results_gerror}
                    \end{subfigure}
                    \begin{subfigure}[b]{0.4\linewidth}
                        \centering
                        \includegraphics[width=\linewidth]{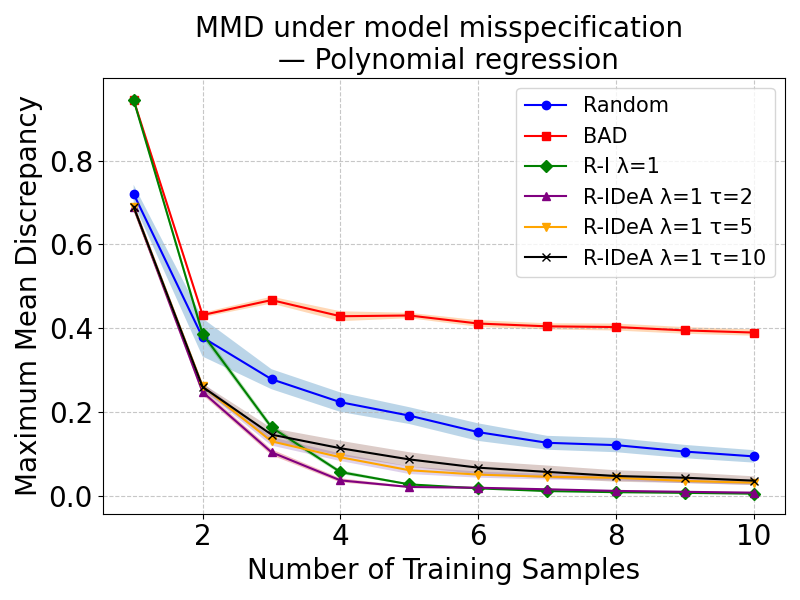}
                        \caption{}
                        \label{subfig:toy_lambda1_tau_mis3result_mmd}
                    \end{subfigure}
                    \caption{\textbf{Polynomial regression experiments (effects of $\lambda$ and $\tau$ in the severely misspecified case).} Comparison of baseline methods (Random, BAD), our proposed \acqfri{} and our proposed \acqfria{} with varying $\tau$.
                    Rows correspond to variation in $\lambda$.
                    \textit{Left}: Generalization error across methods.
                    \textit{Right}: MMD distance across methods; higher values indicate a greater degree of covariate shift.}
                    \label{fig:toy_tau_mis3results}
            \end{figure}

        To investigate the impact of the proposed acquisition function under different levels of misspecification, we ran the same experiments with a higher degree of misspecification by adding a cubic term to the DGP (the DGP is $y=1+2 x-0.5x^2+ 0.2 x^3+ \epsilon$ where $\epsilon \sim \mathcal{N}(0,0.1)$). The results, shown in \Cref{fig:toy_lambda_mis3results,fig:toy_tau_mis3results}, show a similar trend: \acqfri{} and \acqfria{} continue to provide stable and competitive performance even as misspecification increases. 
        We note that when $\lambda = 0.25$ (i.e., informativeness is weighted much more highly than representativeness), both \acqfri{} and \acqfria{} perform worse than Random (\Cref{subfig:toy_lambda025_tau_mis3results_gerror,subfig:toy_lambda025_tau_mis3result_mmd}), demonstrating the importance of representativeness again.

        \subsubsection{Comparison of Different Degrees of Misspecification}
        \label{appendix:poly_differ_mis}
\begin{figure} 
            \centering
            \includegraphics[width=0.5\linewidth]{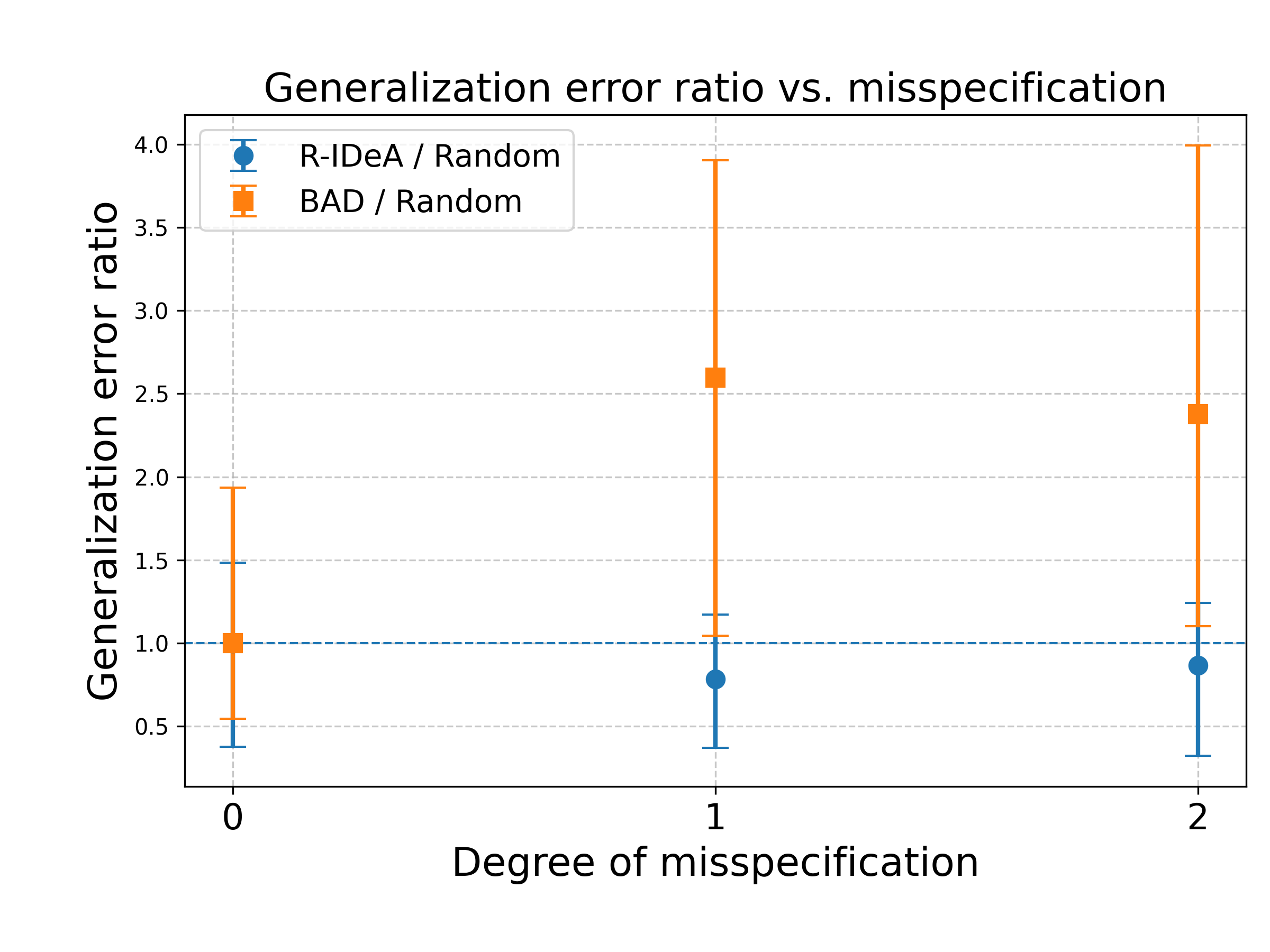}
            \caption{\textbf{Polynomial regression experiments (effect of degree of misspecification).} Comparison of the BAD method and our proposed R-IDeA across different degrees of misspecification. The x-axis represents the degree of misspecification.
            At degree 0, the model is well-specified. At degree 1,  the DGP is $y=1+2 x-0.5x^2+ \epsilon$ while the assumed model is linear. At degree 2, the DGP is $y=1+2 x-0.5x^2+ 0.2 x^3+ \epsilon$ while the assumed model is linear. 
            The y-axis shows the ratio between the generalization error resulting from each method and a random design selection strategy with the 10 selected designs.}
            \label{fig:ratio_vs_mis}
        \end{figure}

        We also compared the relative performance of R-IDeA under different degrees of misspecification.
        \Cref{fig:ratio_vs_mis} shows different degrees of misspecification ($x$-axis) and the ratio of generalization error resulting from using a given method and from a random design strategy ($y$-axis). This ratio is equal to 1 in the well-specified case, showing that each method yields similar generalization error under model well-specification.
        Under model misspecification, \Cref{fig:ratio_vs_mis} shows that the performance of the proposed \acqfria{} increases slightly as misspecification grows but that the ratio remains below 1. These results indicate that the relative advantage of \acqfria{} becomes smaller under severe misspecification, but remains positive. We speculate this slight reduction may be because of an inappropriate choice of $\tau$ (we keep the same $\tau$ regardless of the degree of misspecification). 
        For BAD, as the misspecification degree grows, the error ratio still remains above 1, suggesting that BAD is not robust to the degree of misspecification. The large variation in the relative performance of BAD and Random indicates the instability of the BAD method.

    \subsection{Source Localization Experiments}\label{appendix:ex_source}
        \subsubsection{Model Well-specification}
\begin{figure}[h]  
                \centering 
                \begin{subfigure}[b]{0.4\linewidth}
                    \centering
                    \includegraphics[width=\linewidth]{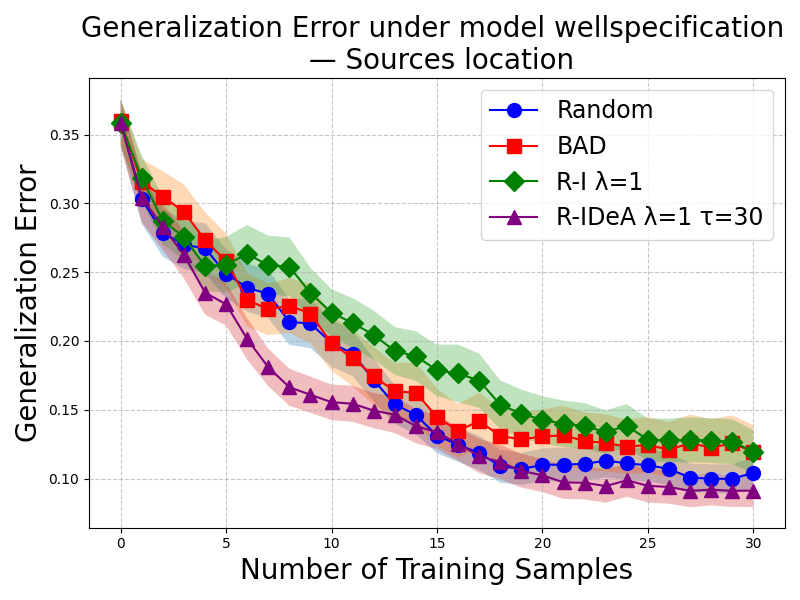}
                    \caption{}
                    \label{subfig:source_base_results_gerror}
                \end{subfigure}
                \begin{subfigure}[b]{0.4\linewidth}
                    \centering
                    \includegraphics[width=\linewidth]{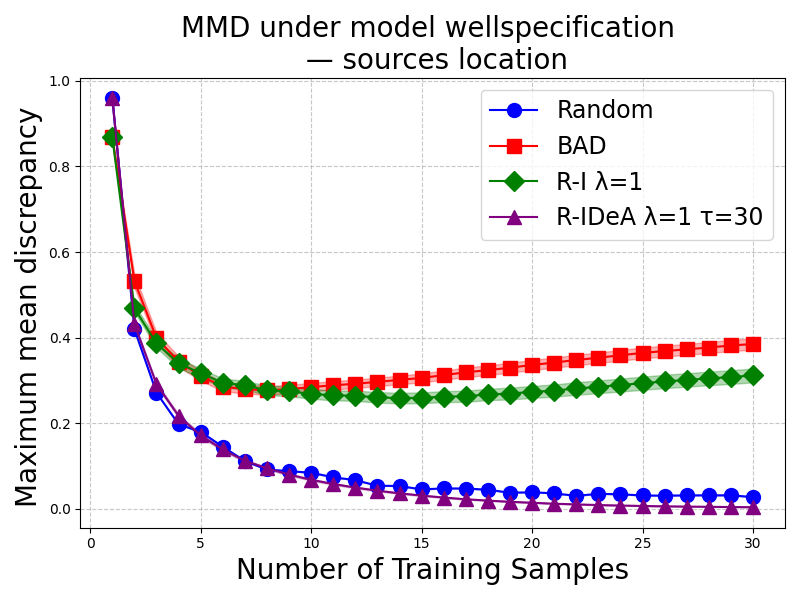}
                    \caption{}
                    \label{subfig:source_base_results_mmd}
                \end{subfigure}
                \caption{\textbf{Sources location experiments (well-specified case).} Comparison of different design strategies (Random, BAD, proposed \acqfri{}, proposed \acqfria{} under well-specified models in sources location.
                \textit{Left}: Generalization error across methods.
                \textit{Right}: MMD distance across methods; higher values indicate a greater degree of covariate shift.
                }
                \label{fig:source_results_well}
            \end{figure}

            Like in the toy example, \Cref{fig:source_results_well}
            illustrates that covariate shift does not impact generalization performance when the model is well-specified.
            In the well-specified model, Random and BAD result in similar generalization errors. This may be due to poor estimation of the expected information gain (EIG) in high-dimensional spaces.
            \acqfria{} has the best and fastest performance in the well-specified case compared to other methods. 

        \subsubsection{Model Misspecification} \label{sub:ex_source_hyper}
                 
\begin{figure}  
                    \centering 
                    \begin{subfigure}[b]{0.4\linewidth}
                        \centering
                        \includegraphics[width=\linewidth]{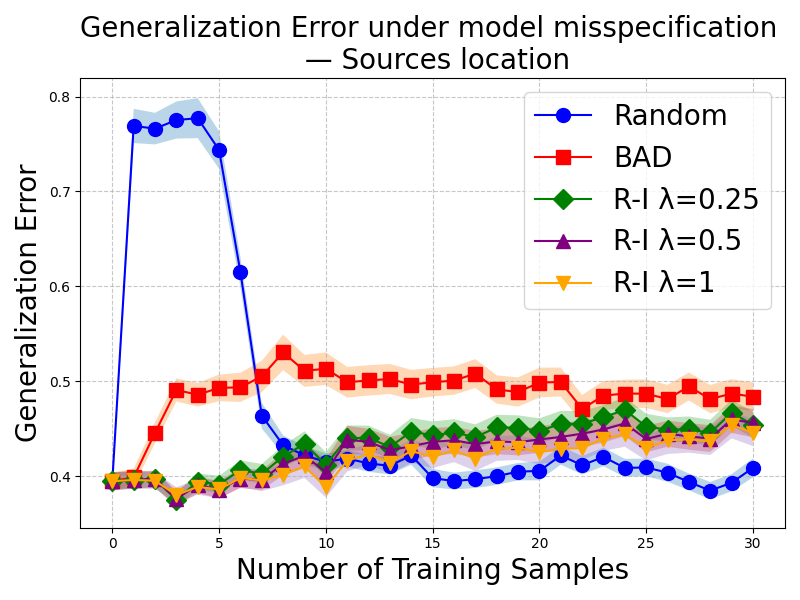}
                        \caption{}
                        \label{subfig:source_lambda_misresults_gerror}
                    \end{subfigure}
                    \begin{subfigure}[b]{0.4\linewidth}
                        \centering
                        \includegraphics[width=\linewidth]{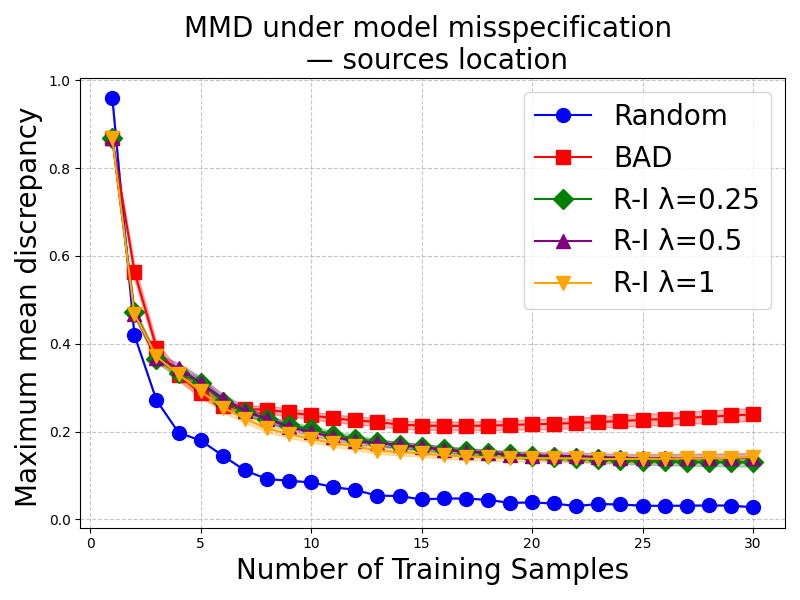}
                        \caption{}
                        \label{subfig:source_lambda_misresult_mmd}
                    \end{subfigure}
                    \caption{\textbf{Sources location experiments (effect of $\lambda$).} Comparison of baseline methods (Random, BAD) and our proposed \acqfri{} with varying $\lambda$ in the sources location experiments
                    \textit{Left}: Generalization error across methods.
                    \textit{Right}: MMD distance across methods; higher values indicate a greater degree of covariate shift.}
                    \label{fig:source_lambda_misresults}
            \end{figure}

        \paragraph{\acqfri{} - varying $\lambda$}
            \Cref{subfig:source_lambda_misresults_gerror} shows that our novel \acqfri{} acquisition function achieves a smaller generalization error than BAD. \Cref{subfig:source_lambda_misresult_mmd} further demonstrates that the designs selected by this robust acquisition function are more representative than those selected by BAD. These results show that representative designs effectively reduce estimation bias and improve generalization performance. 
            However, varying the value of $\lambda$ leads to similar generalization performance and covariate shift, suggesting that performance is not sensitive to $\lambda$ and the ability of \acqfri{} to improve representativeness in high-dimensional settings is limited.

\begin{figure}  
                    \centering 
                    \begin{subfigure}[b]{0.4\linewidth}
                        \centering
                        \includegraphics[width=\linewidth]{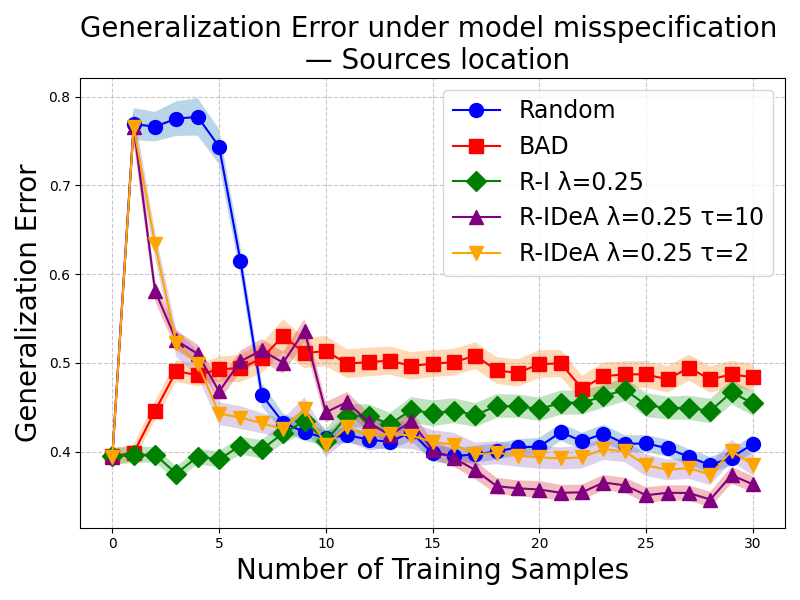}
                        \caption{}
                        \label{subfig:source_lambda025_tau_misresults_gerror}
                    \end{subfigure}
                    \begin{subfigure}[b]{0.4\linewidth}
                        \centering
                        \includegraphics[width=\linewidth]{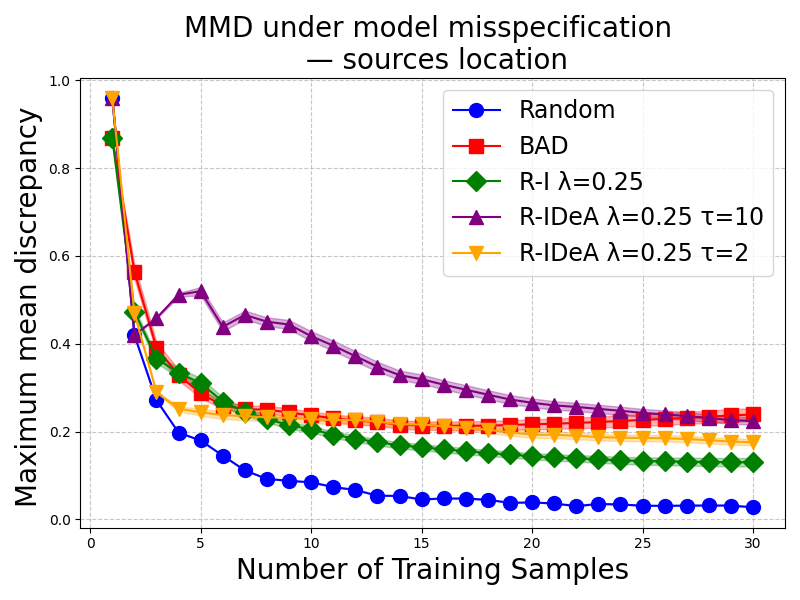}
                        \caption{}
                        \label{subfig:source_lambda025_tau_misresult_mmd}
                    \end{subfigure}
                    
                    \begin{subfigure}[b]{0.4\linewidth}
                        \centering
                        \includegraphics[width=\linewidth]{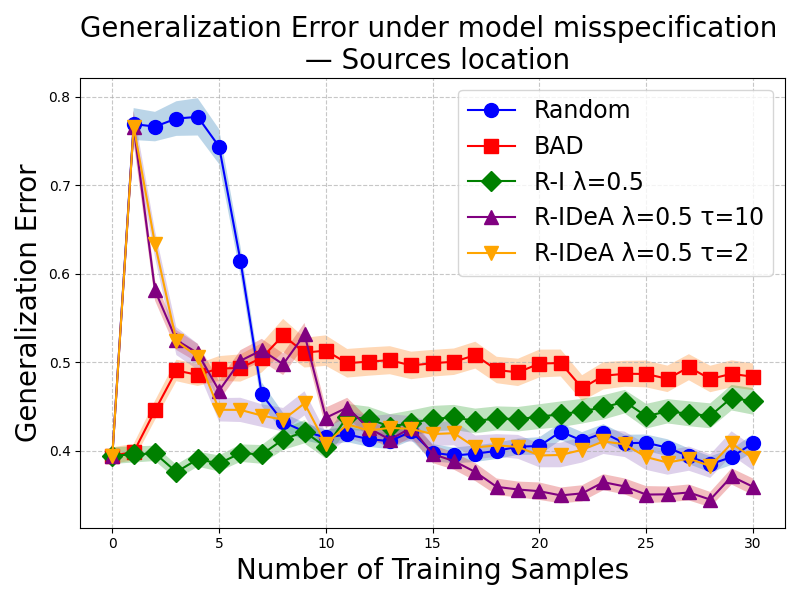}
                        \caption{}
                        \label{subfig:source_lambda05_tau_misresults_gerror}
                    \end{subfigure}
                    \begin{subfigure}[b]{0.4\linewidth}
                        \centering
                        \includegraphics[width=\linewidth]{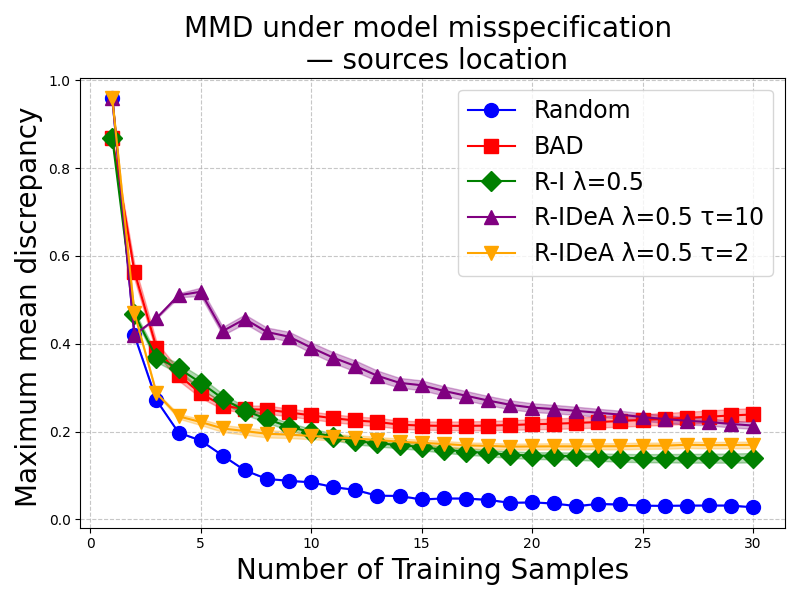}
                        \caption{}
                        \label{subfig:source_lambda05_tau_misresult_mmd}
                    \end{subfigure}
                    
                    \begin{subfigure}[b]{0.4\linewidth}
                        \centering
                        \includegraphics[width=\linewidth]{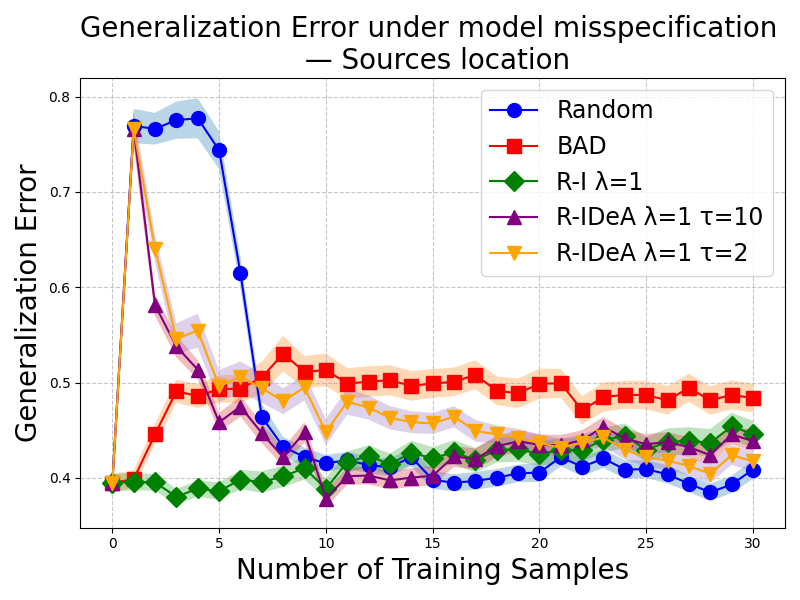}
                        \caption{}
                        \label{subfig:source_lambda1_tau_misresults_gerror}
                    \end{subfigure}
                    \begin{subfigure}[b]{0.4\linewidth}
                        \centering
                        \includegraphics[width=\linewidth]{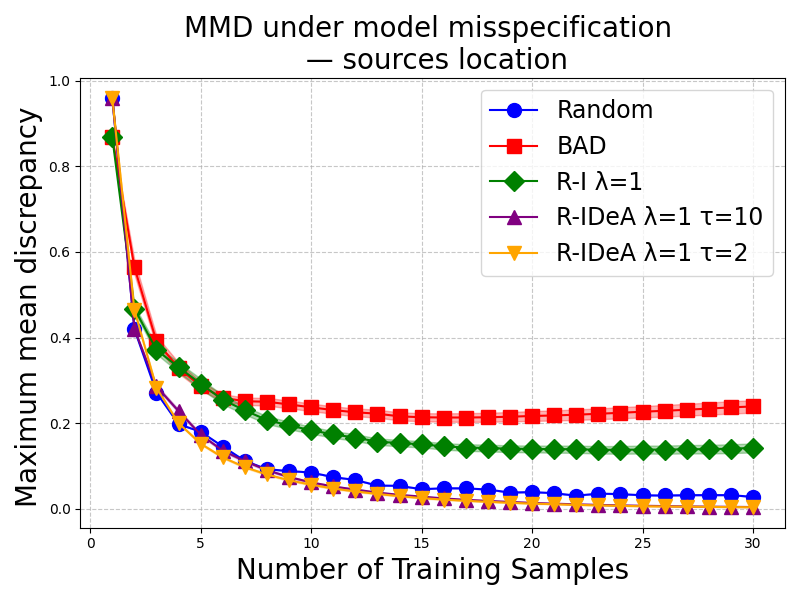}
                        \caption{}
                        \label{subfig:source_lambda1_tau_misresult_mmd}
                    \end{subfigure}
                    \caption{\textbf{Sources location experiments (effect of $\tau$)} Comparison of baseline methods (Random, BAD), our proposed \acqfri{} and our proposed \acqfria{} with varying $\tau$ in the source location experiments.
                    Rows correspond to variation in $\lambda$.
                    \textit{Left}: Generalization error across methods.
                    \textit{Right}: MMD distance across methods; higher values indicate a greater degree of covariate shift.}
                    \label{fig:source_tau_misresults}
            \end{figure}

        \paragraph{\acqfria{} - varying $\tau$}
            \Cref{fig:source_tau_misresults} illustrates that across different values of $\tau$, \acqfria{} consistently outperforms \acqfri{} in terms of generalization performance. These results support the conclusion we drew from the polynomial regression experiments that incorporating error de-amplification improves generalization. 
            For $\lambda = 0.25$ and $\lambda = 0.5$, R-IDeA can lead to designs that, depending on the value of $\tau$, exhibit more or less covariate shift than BAD (\Cref{subfig:source_lambda025_tau_misresult_mmd} and \Cref{subfig:source_lambda05_tau_misresult_mmd}).
            However, our experimental results suggest that even when less representative, these designs lead to higher generalization performance (\Cref{subfig:source_lambda025_tau_misresults_gerror} and \Cref{subfig:source_lambda05_tau_misresults_gerror}).
              When $\lambda = 1$, \acqfria{} performs worse than Random but still better than BAD (\Cref{subfig:source_lambda1_tau_misresults_gerror}), showing that both representativeness and de-amplification contribute to robustness.

    \subsection{Pharmacokinetic Model}\label{appendix:ex_pk}
        \subsubsection{Model Well-specification}
\begin{figure}  
                \centering 
                \begin{subfigure}[b]{0.4\linewidth}
                    \centering
                    \includegraphics[width=\linewidth]{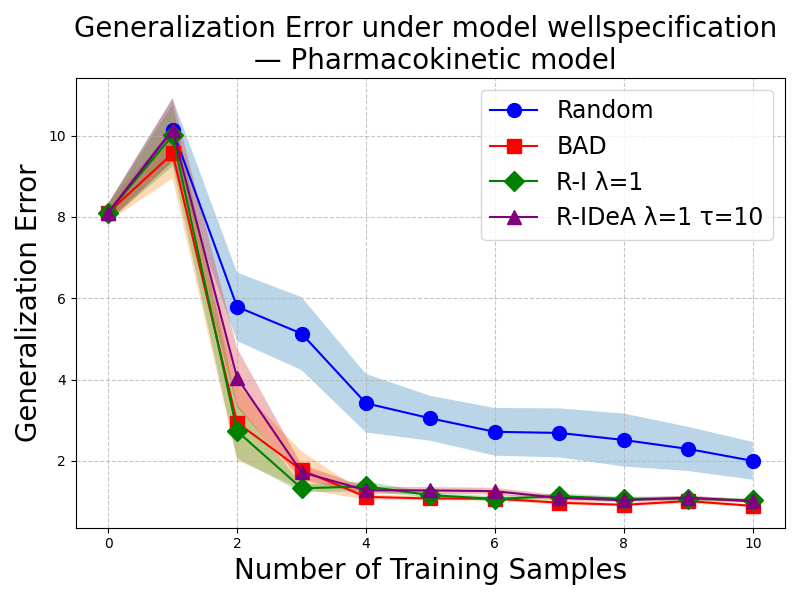}
                    \caption{}
                    \label{subfig:pk_base_results_gerror}
                \end{subfigure}
                \begin{subfigure}[b]{0.4\linewidth}
                    \centering
                    \includegraphics[width=\linewidth]{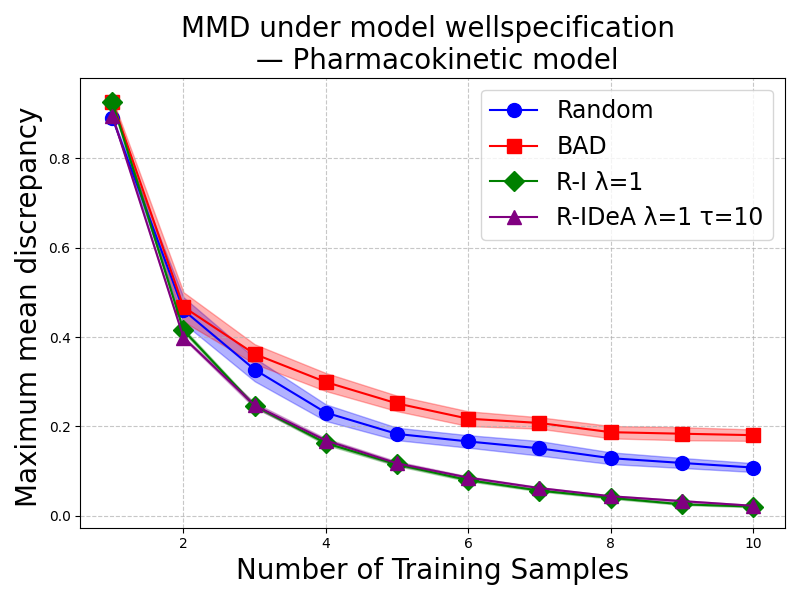}
                    \caption{}
                    \label{subfig:pk_base_results_mmd}
                \end{subfigure}
                \caption{\textbf{Pharmacokinetic model experiments (well-specified case).} Comparison of different design strategies (Random, BAD, proposed \acqfri{}, proposed \acqfria{} and \acqfria{}-oracle which uses the $\barf$ instead of the proxy $g$) under well-specified models in the Pharmacokinetic model experiment.
                \textit{Left}: Generalization error across methods.
                \textit{Right}: MMD distance across methods; higher values indicate a greater degree of covariate shift.
                }
                \label{fig:pk_results_well}
            \end{figure}

            Like in the polynomial regression experiments (\Cref{fig:toy_results_well}) and source localization experiments (\Cref{fig:source_results_well}), \Cref{fig:pk_results_well}
            illustrates that covariate shift does not impact generalization performance when the model is well-specified.
        
            BAD, \acqfri{} and \acqfria{} decrease error more quickly than random in the well-specified case, suggesting that the expected information gain (EIG) leads to informativeness designs. 
            
        \subsubsection{Model Misspecification} \label{sub:ex_pk_hyper}
                 
\begin{figure}  
                    \centering 
                    \begin{subfigure}[b]{0.4\linewidth}
                        \centering
                        \includegraphics[width=\linewidth]{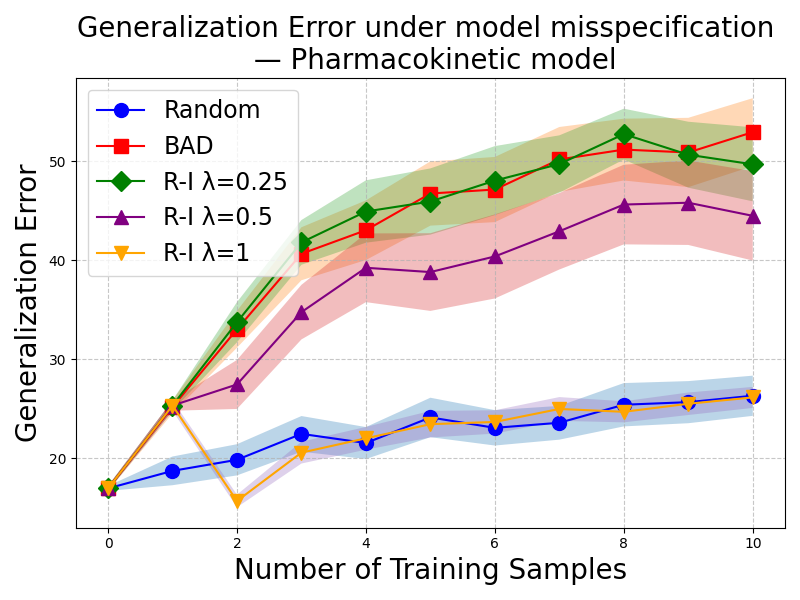}
                        \caption{}
                        \label{subfig:pk_lambda_misresults_gerror}
                    \end{subfigure}
                    \begin{subfigure}[b]{0.4\linewidth}
                        \centering
                        \includegraphics[width=\linewidth]{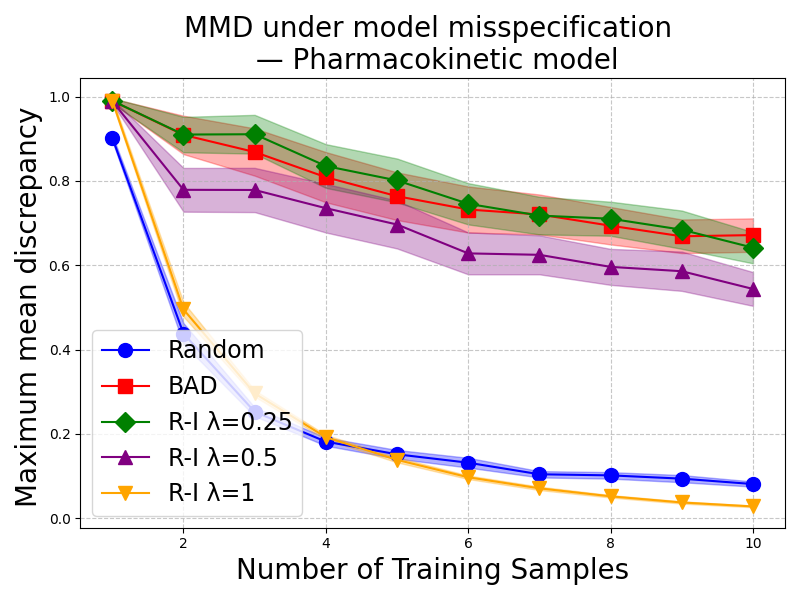}
                        \caption{}
                        \label{subfig:pk_lambda_misresult_mmd}
                    \end{subfigure}
                    \caption{\textbf{Pharmacokinetic model experiments (effect of $\lambda$).} Comparison of baseline methods (Random, BAD) and our proposed \acqfri{} with varying $\lambda$ in the Pharmacokinetic model experiments
                    \textit{Left}: Generalization error across methods.
                    \textit{Right}: MMD distance across methods; higher values indicate a greater degree of covariate shift.}
                    \label{fig:pk_lambda_misresults}
            \end{figure}

        \paragraph{\acqfri{} - varying $\lambda$}
            \Cref{fig:pk_lambda_misresults} presents the performance of our \acqfri{} acquisition function under different values of $\lambda$.
            As shown in \Cref{subfig:pk_lambda_misresult_mmd}, more representative designs lead to lower generalization error (\Cref{subfig:pk_base_results_gerror}), consistent with the theoretical prediction in \Cref{prop:gen_error_covshift}. 
            The effect of varying $\lambda$ follows the same trend as in the Polynomial Regression experiment in \Cref{fig:toy_lambda_misresults}, further demonstrating the robustness of the \acqfri{}.
            
\begin{figure}[H]  
                    \centering 
                    \begin{subfigure}[b]{0.4\linewidth}
                        \centering
                        \includegraphics[width=\linewidth]{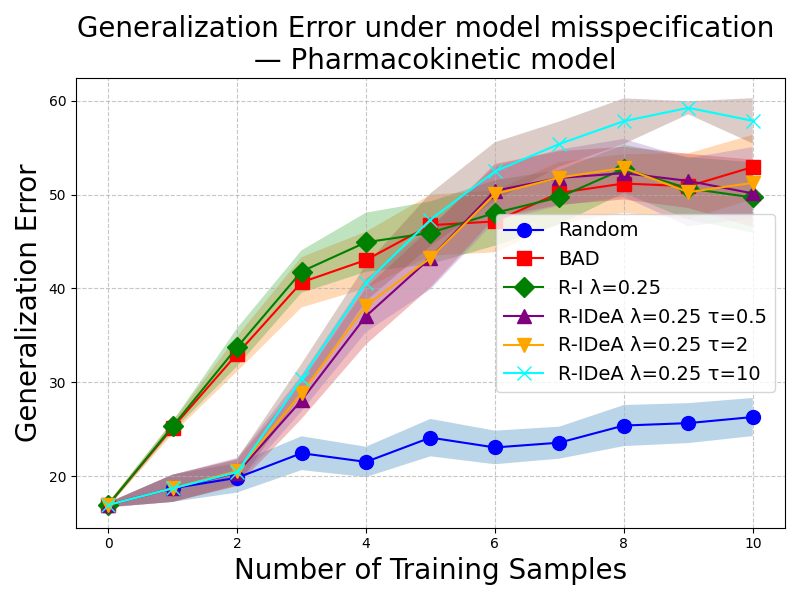}
                        \caption{}
                        \label{subfig:pk_lambda025_tau_misresults_gerror}
                    \end{subfigure}
                    \begin{subfigure}[b]{0.4\linewidth}
                        \centering
                        \includegraphics[width=\linewidth]{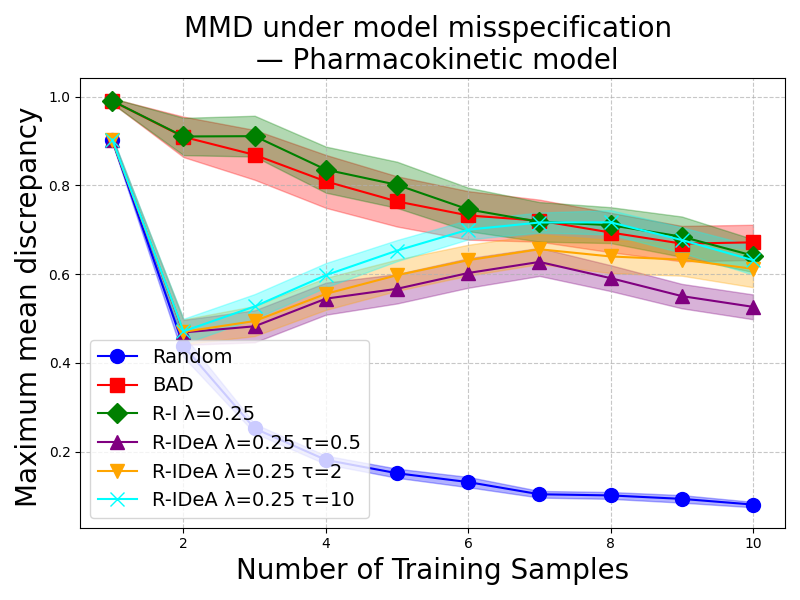}
                        \caption{}
                        \label{subfig:pk_lambda025_tau_misresult_mmd}
                    \end{subfigure}
                    
                    \begin{subfigure}[b]{0.4\linewidth}
                        \centering
                        \includegraphics[width=\linewidth]{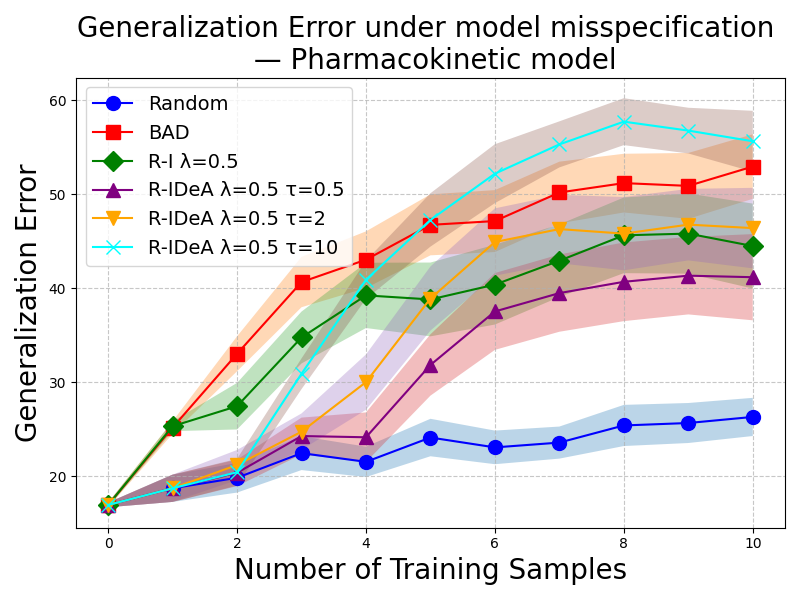}
                        \caption{}
                        \label{subfig:pk_lambda05_tau_misresults_gerror}
                    \end{subfigure}
                    \begin{subfigure}[b]{0.4\linewidth}
                        \centering
                        \includegraphics[width=\linewidth]{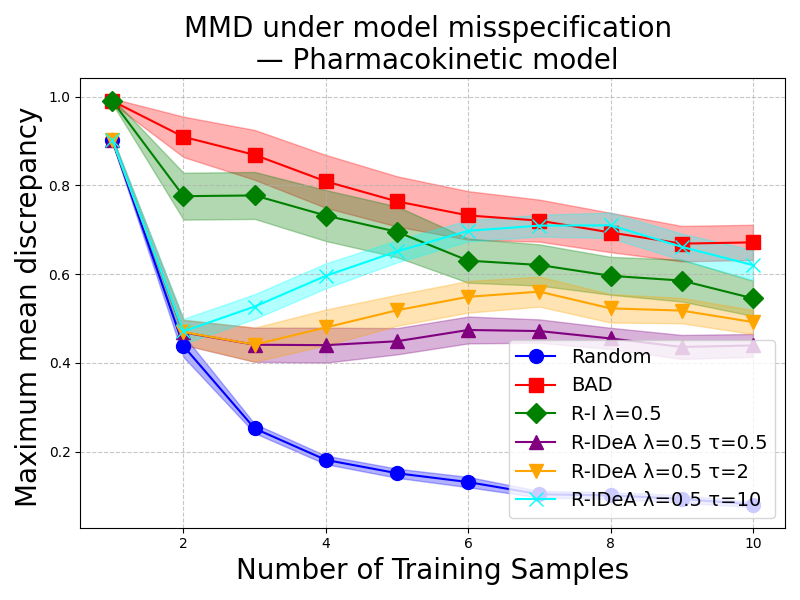}
                        \caption{}
                        \label{subfig:pk_lambda05_tau_misresult_mmd}
                    \end{subfigure}
                    
                    \begin{subfigure}[b]{0.4\linewidth}
                        \centering
                        \includegraphics[width=\linewidth]{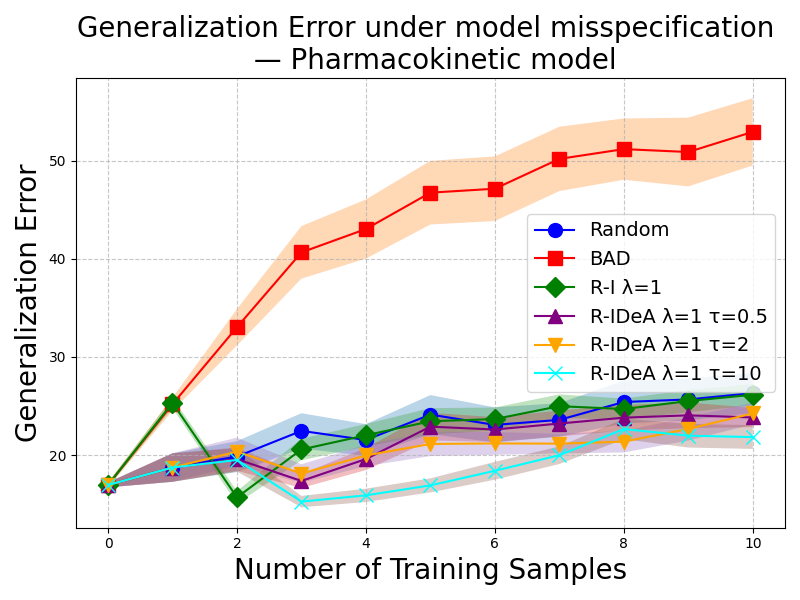}
                        \caption{}
                        \label{subfig:pk_lambda1_tau_misresults_gerror}
                    \end{subfigure}
                    \begin{subfigure}[b]{0.4\linewidth}
                        \centering
                        \includegraphics[width=\linewidth]{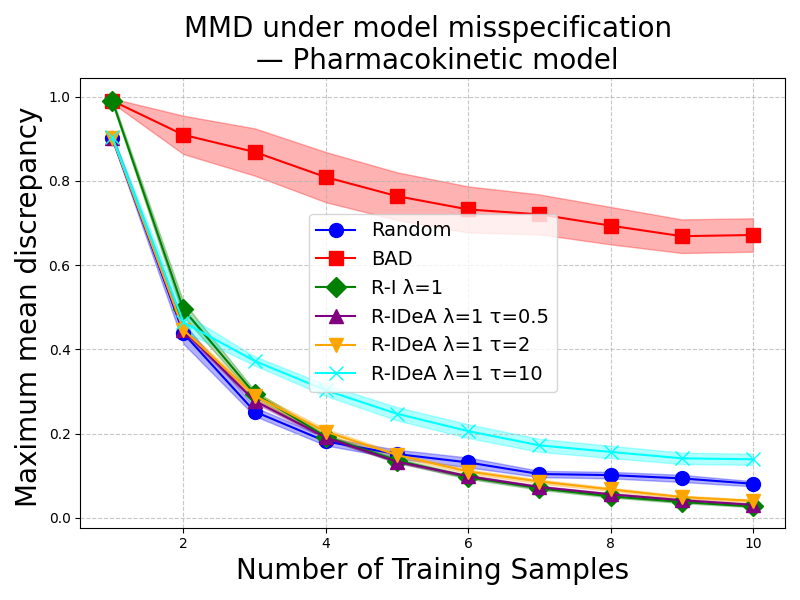}
                        \caption{}
                        \label{subfig:pk_lambda1_tau_misresult_mmd}
                    \end{subfigure}
                    \caption{\textbf{Pharmacokinetic model experiments (effect of $\tau$).} Comparison of baseline methods (Random, BAD), our proposed \acqfri{} and our proposed \acqfria{} with varying $\tau$ in the Pharmacokinetic model experiments.
                    Rows correspond to variation in $\lambda$.
                    \textit{Left}: Generalization error across methods.
                    \textit{Right}: MMD distance across methods; higher values indicate a greater degree of covariate shift.}
                    \label{fig:pk_tau_misresults}
            \end{figure}

        \paragraph{\acqfria{} - varying $\tau$}
             To explore how the hyperparameter $\tau$ affects our proposed acquisition function, \Cref{fig:pk_tau_misresults} shows the performance of our novel \acqfria{} acquisition function with different values of $\tau$.

             When $\lambda = 1$, \Cref{subfig:pk_lambda1_tau_misresults_gerror} illustrates that across different values of $\tau$, \acqfria{} consistently outperforms \acqfri{} and Random in terms of generalization performance, while some value of $\tau$ leads to a larger degree of covariate shift (\Cref{subfig:pk_lambda1_tau_misresult_mmd}).
             These findings support our earlier conclusion in \Cref{prop:gen_error_covshift} that incorporating error de-amplification improves generalization.
           
             Interestingly, for $\lambda = 0.25$ (\Cref{subfig:pk_lambda025_tau_misresults_gerror}), \acqfria{} performs worse than both Random and \acqfri{}. We speculate that this is due to an inappropriate choice of $\tau.$
             When $\lambda = 0.5$ (\Cref{subfig:pk_lambda05_tau_misresults_gerror}), \acqfria{} with $ \tau = 0.5$ outperforms \acqfri{}, whereas other values of $\tau$ lead to worse performance, highlighting the importance of properly choosing $\tau$.
             For $\lambda = 1 $ (\Cref{subfig:pk_lambda1_tau_misresults_gerror}), \acqfria{} outperforms both Random and \acqfri{}. Moreover, \acqfria{} with $\tau = 10$ achieves better results than with other values of $\tau$, illustrating that selecting an appropriate hyperparameter cannot be achieved by simply increasing or decreasing its value in a heuristic manner.  
             From \Cref{equ:Lg} and \Cref{equ:eig_de_amplify}, larger values of $\tau$ cause the de-amplifying term to dominate the acquisition function, enforcing a strict de-amplifying constraint, whereas smaller values loosen this constraint and yield more candidate designs. Therefore, $\tau$ should be carefully chosen to balance de-amplification with other properties such as informativeness and representativeness. How to choose the hyperparameters in our \acqfri{} and \acqfria{} is an avenue for future work.
           
% \onecolumn
% \appendix
\newpage

\end{document}